\documentclass{article}

\PassOptionsToPackage{round,authoryear}{natbib}

\usepackage[main,final]{neurips_2026}

\usepackage[utf8]{inputenc} 
\usepackage[T1]{fontenc}    
\usepackage{hyperref}       
\usepackage{url}            
\usepackage{booktabs}       
\usepackage{amsfonts}       
\usepackage{nicefrac}       
\usepackage{microtype}      
\usepackage{xcolor}         

\usepackage{amsmath}
\usepackage{amsthm}
\usepackage{amssymb}
\usepackage{wrapfig}
\usepackage{float}
\usepackage{enumitem}
\usepackage[capitalise]{cleveref}
\usepackage{multirow}
\usepackage{makecell}
\usepackage{graphicx}
\usepackage{tikz}
\usetikzlibrary{arrows.meta,positioning,calc,fit,shapes.geometric}
\usepackage{subcaption}
\graphicspath{{figs/}}

\definecolor{peach}{HTML}{F88379}
\definecolor{mango}{HTML}{FFD94D}

\newtheorem{theorem}{Theorem}[section]

\newtheorem{proposition}[theorem]{Proposition}

\newtheorem{assumption}[theorem]{Assumption}

\theoremstyle{remark}

\makeatletter
\newcommand*{\rom}[1]{\expandafter\@slowromancap\romannumeral #1@}
\makeatother

\newcommand{\M}{[\textup{\textbf{M}}]}

\title{
A Theoretical Analysis of Why Masked Diffusion Models Mitigate the Reversal Curse}

\author{
Moongyu Jeon\thanks{Equal contribution.} \quad\;
Sangwoo Shin\footnotemark[1] \quad\;
Bumjun Kim \quad\;
Kyelim Lee \quad\;
Albert No\thanks{Corresponding author: \texttt{albertno@yonsei.ac.kr}.} \\[0.7em]
Department of Artificial Intelligence, Yonsei University
}

\begin{document}

\maketitle

\begin{abstract}
Autoregressive language models (ARMs) suffer from the \textit{reversal curse}: after learning ``\(A\) is \(B\),'' they often fail on the reverse query ``\(B\) is \(A\).''
Masked diffusion language models (MDMs) exhibit this failure to a much lesser extent, but the underlying reason remains unclear.
This mitigation is commonly attributed to the random-masking objective, which trains MDMs to reconstruct masked tokens using bidirectional context.
However, learning to recover \(A\) from ``\([\textnormal{\textbf{M}}]\)~is~\(B\)'' does not by itself explain how this learning transfers to recovering \(A\) from ``\(B\)~is~\([\textnormal{\textbf{M}}]\).''
We provide a theoretical analysis showing how this transfer can arise in a one-layer MDM from a parameter-level coupling between forward and reverse positional conditionals:
shared Transformer parameters store token-pair evidence, while relative positional encodings route attention through queries and keys without changing the value-side evidence being retrieved.
Specifically, we prove that forward masked training strengthens evidence reusable in reverse queries, establish positive expected forward--reverse attention-score correlation under stated assumptions, and show that the shared-storage component of a gradient-descent step on the forward loss decreases the reverse loss to first order.
Controlled one-layer and large-scale LLaDA/Dream experiments exhibit the predicted signatures, supporting the proposed mechanism as an explanation for improved reverse prediction.
\end{abstract}

\section{Introduction}
\label{sec:introduction}

Autoregressive language models (ARMs) with causal-attention Transformers are the dominant paradigm for large-scale language modeling~\citep{vaswani2017attention,radford2018improving,radford2019language,brown2020language}.
Despite their empirical success, they exhibit the \emph{reversal curse}: after learning a relational fact in one order, a model may fail to answer the corresponding order-reversed query~\citep{berglundreversal}.
For example, a model trained on ``The capital of France is Paris'' may answer ``What is the capital of France?'' but fail to answer ``Which country has Paris as its capital?''
This failure exposes a gap between storing a fact in the learned weights and making it accessible under order reversal: when the fact is explicitly provided in context, strong models can often infer the reverse relation, but the same relation need not be retrievable from the weights alone~\citep{berglundreversal}.

Under the left-to-right next-token prediction objective used by ARMs, this failure mode arises naturally.
An ARM trained on ``\(A\) is \(B\)'' directly reinforces predicting \(B\) from a context containing \(A\), whereas the reverse query asks for \(A\) from a context containing \(B\), corresponding to a different conditional.
\citet{zhu2024towards} formalize this asymmetry through training dynamics in a simplified one-layer Transformer, showing that training on one direction does not strengthen the reverse direction.

Masked diffusion language models (MDMs) simultaneously predict token distributions at masked positions, supporting flexible any-order generation~\citep{austin2021structured,lou2024discrete,sahoo2024simple,shi2024simplified,ouyour,nie2025large}.
Recent work reports that MDMs exhibit a substantially weaker reversal curse than comparable ARMs~\citep{he2026differ}.
This advantage is often attributed to their random-masking training objective, which allows MDMs to recover masked tokens using context from both sides, in contrast to ARMs that predict tokens using only preceding context~\citep{kitouni2024factorizationcursetokenspredict,nie2025scaling,nie2025large}.

\begin{wrapfigure}{r}{0.46\textwidth}
\vspace{-0.13in}
\centering
\includegraphics[width=\linewidth]{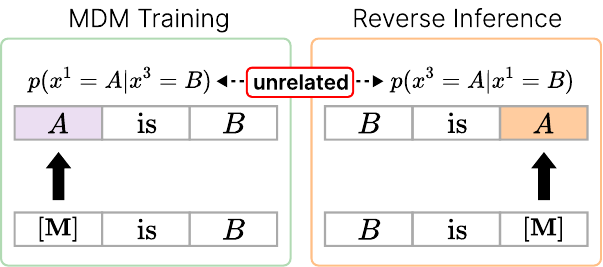}
\vspace{-0.13in}
\caption{
Masked training teaches one conditional, but the objective alone does not explain its link to the reverse conditional.
}
\label{fig:intro_masking_gap}
\vspace{-0.10in}
\end{wrapfigure}

However, we argue that this objective-only explanation is incomplete, as illustrated in Fig.~\ref{fig:intro_masking_gap}.
As a representative forward example, suppose ``\(A\) is \(B\)'' is trained, with \(A\) to the left of \(B\).
Random masking can create the forward masked example ``\([\textnormal{\textbf{M}}]\) is \(B\),'' where the masked position~\(i\) has target~\(A\) and the evidence token~\(B\) appears at position~\(j\), with \(i<j\).
This provides supervision for the forward conditional \(p_\theta(x^i\!=\!A|x^j\!=\!B)\).
The corresponding reverse query ``\(B\) is \([\textnormal{\textbf{M}}]\)'' asks for the same target~\(A\) from \(B\), but in the opposite positional arrangement, corresponding to the reverse conditional \(p_\theta(x^{i'}\!=\!A|x^{j'}\!=\!B)\), where \(j'<i'\).
Thus, random masking supervises recovering \(A\) from \(B\) in the forward conditional, but does not explain why this evidence transfers to the reverse conditional.

We establish that, in a one-layer MDM, forward and reverse positional conditionals are coupled through \textit{position-invariant evidence storage} and \textit{attention-based routing}.
With relative positional encodings such as RoPE~\citep{su2024roformer}, the contribution of token~\(B\) at position~\(j\) to the \(A\)-logit at masked position~\(i\) decomposes as \(\alpha_{i,j}Y_{A,B}\).
Here \(Y_{A,B}\) stores position-invariant \(B\)-to-\(A\) evidence because position affects query-key attention scores but not value vectors, while \(\alpha_{i,j}\) controls how strongly this evidence is retrieved.
We prove that forward masked training strengthens \(Y_{A,B}\), relative positional encodings preserve positive expected forward--reverse attention-score correlation under stated conditions, and the shared-storage component of a gradient-descent step on the forward loss decreases the reverse loss to first order.

Controlled one-layer experiments show that the ingredients identified by our theory are present in trained models: forward--reverse attention correlation, gradient alignment, and reverse-probability improvement under forward-only training.
Positional encoding interventions further support the analysis by changing the storage-routing components and producing the predicted changes in reverse transfer.
Experiments with the multi-layer masked diffusion LLMs LLaDA~\citep{nie2025large} and Dream~\citep{ye2025dream7bdiffusionlarge} exhibit signatures consistent with the predictions of the one-layer theory.

\section{Background and Related Work}
\label{sec:background}
\subsection{Autoregressive Models and the Reversal Curse}
\label{sec:prelim_arm}

Autoregressive language models (ARMs)~\citep{radford2018improving,radford2019language} generate tokens from left to right, factorizing the probability of a sequence \(\mathbf{x}=x^1x^2\ldots x^L\) as
\(p_\theta(\mathbf{x})=\prod_{i=1}^L p_\theta(x^i|\mathbf{x}^{<i})\),
where \(\mathbf{x}^{<i}\) denotes the tokens preceding position~\(i\).
In Transformer-based ARMs with causal self-attention, next-token prediction uses only this left context, making the training signal intrinsically directional.

This directionality helps explain the \emph{reversal curse}~\citep{berglundreversal}.
When an ARM is trained on ``\(A\) is \(B\),'' its objective directly reinforces predicting \(B\) from a context containing \(A\).
The reverse query instead asks the model to predict \(A\) from a context containing \(B\), a prediction not directly reinforced by the forward example.\footnote{Following~\citet{berglundreversal}, we use ``\(A\) is \(B\)'' and ``\(B\) is \(A\)'' as schematic notation. Our focus is whether a model can answer a semantically valid reverse query whose answer follows from the learned forward fact. See App.~\ref{app:reversal_curse_scope} for further discussion.}
Several recent works investigate reversal failures through reverse training, dataset design, relation-aware augmentation, and broader analyses of order-sensitive generalization~\citep{golovnevareverse,lin2024delving,lu2024rethinking,wang2025reversal}.
These works primarily characterize or mitigate reversal failures without providing a formal theoretical analysis of their underlying mechanisms.

In contrast, \citet{zhu2024towards} provide a theoretical explanation of the reversal curse in a simplified one-layer ARM.
When trained on ``\(A\) is \(B\),'' the model strengthens the input-target weight from \(A\) to \(B\), but not the reverse weight from \(B\) to \(A\), an effect they call \textit{weight asymmetry}.
Under zero initialization of the reparameterized weights, they further prove that the forward prediction probability approaches one, while the reverse probability does not exceed the random-guess level.
This result motivates a corresponding analysis of MDMs.
For a direct comparison, App.~\ref{app:arm_comparison} reviews their assumptions and establishes an analogous result for a one-layer ARM.

\subsection{Masked Diffusion Models and Reversal Curse Mitigation}
\label{sec:prelim_mdm}

Discrete diffusion models generate discrete data by learning to reverse a gradual noising process that transforms clean data into a simple noise distribution~\citep{hoogeboom2021argmax,austin2021structured,campbell2022continuous,lou2024discrete}.
Masked diffusion models (MDMs) instantiate this framework by using token masking as the noising process, generating text from fully masked sequences by iteratively recovering masked positions~\citep{sahoo2024simple,shi2024simplified,ouyour}.
During training, a clean sequence \(\mathbf{x}=x^1x^2\ldots x^L\) is corrupted into \(\mathbf{x}_t\) by independently replacing each token with \([\textnormal{\textbf{M}}]\) with probability \(t\in[0,1]\).
The training objective is a weighted expectation of the cross-entropy loss \(-\log p_\theta(x^i|\mathbf{x}_t)\) for reconstructing the original token~\(x^i\) at each masked position~\(i\).
Unlike ARMs, MDMs use full attention, so a masked position can attend to unmasked tokens on either side.
A useful point of comparison is BERT-style masked language modeling~\citep{devlin2019bert}: MDMs share its full-attention masked reconstruction but use it for iterative generation, without requiring explicit time embeddings in the Transformer backbone~\citep{sahoo2024simple,ouyour,nie2025large,ye2025dream7bdiffusionlarge} (see App.~\ref{app:mdm_framing}).

Recent work reports that MDMs exhibit a weaker reversal curse than ARMs.
\citet{kitouni2024factorizationcursetokenspredict} show that uniform-rate masked language models, which are closely related to MDMs, mitigate reversal failures in controlled settings.
\citet{nie2025scaling} and \citet{nie2025large} further report reversal advantages in masked diffusion LLMs.
These results motivate a natural explanation based on the random-masking objective: unlike left-to-right ARM training, MDMs learn to reconstruct masked tokens using context from both preceding and following positions.

However, this objective-only explanation is incomplete (Fig.~\ref{fig:intro_masking_gap}).
Concretely, a forward statement ``\(A\) is \(B\)'' can yield a masked training example that provides supervision for the schematic conditional \(p_\theta(x^i\!=\!A|x^j\!=\!B)\).
The reverse query asks for the same target~\(A\) from the same evidence token~\(B\), but in a different positional configuration, corresponding to \(p_\theta(x^{i'}\!=\!A|x^{j'}\!=\!B)\).
The masking objective alone does not explain why improving the first conditional should also improve the second.
We show that these conditionals are linked through a parameter-level coupling in a one-layer MDM.

\section{One-Layer Setup: The Storage-Routing Decomposition}
\label{sec:setup}

This section introduces a simplified one-layer MDM to isolate the storage-routing mechanism, distinguishing \emph{what evidence is stored} from \emph{how that evidence is routed} by attention.
We represent these roles by a position-invariant storage term \(Y_{A,B}\) and a position-dependent attention weight \(\alpha_{i,j}\), respectively.
This decomposition forms the basis for our theoretical results in Sec.~\ref{sec:theory}.
Our use of a single Transformer attention layer is motivated by prior theoretical studies that simplify Transformer architectures to isolate learning mechanisms~\citep{tian2023scan,li2023transformers,zhu2024towards,huang2025generalization}.
Although this setup does not capture the full complexity of deep MDMs, we test the mechanism's predicted signatures in trained one-layer MDMs and large-scale masked diffusion LLMs in Secs.~\ref{sec:one_layer_validation} and \ref{sec:large_scale_validation}.
See App.~\ref{app:one_layer_setup} for explicit dimensions and further discussion.

\paragraph{One-layer MDM with relative positional encoding.}
Consider a one-layer MDM applied to a partially masked sequence \(\mathbf{x}=x^1x^2\ldots x^L\).
Let \(i\) be a masked position with \(x^i=[\textnormal{\textbf{M}}]\), and let \(A\) be the target token to be recovered at this position.
Let \(j\) be a position containing the unmasked input token \(x^j=B\).
For a token \(x\), let \(\mathbf{e}_x\) denote its one-hot vector and \(\mathbf{h}_x=W_E\mathbf{e}_x\) its token embedding.
The query, key, and value vectors are
\(\mathbf{q}_x=W_Q\mathbf{h}_x\), \(\mathbf{k}_x=W_K\mathbf{h}_x\), \(\mathbf{v}_x=W_V\mathbf{h}_x\).

We consider relative positional encodings such as RoPE~\citep{su2024roformer} and ALiBi~\citep{press2022alibi}, where position affects attention scores but not value vectors. 
To cover these encodings, write the attention score and attention weight from masked position \(i\) to position \(k\) as
\[
S_{i,k}=S(\mathbf{q}_{x^i},\mathbf{k}_{x^k},k-i),
\quad
\alpha_{i,k}
=
\frac{\exp(S_{i,k}/\sqrt{D})}{\sum_{k'=1}^L\exp(S_{i,k'}/\sqrt{D})}.
\]
Thus \(\alpha_{i,k}>0\) and \(\sum_{k=1}^L\alpha_{i,k}=1\).
The masked-position logit vector and prediction distribution are
\[
\mathbf{z}_i
=
W_U\sum_{k=1}^L \alpha_{i,k}\mathbf{v}_{x^k},
\quad
\mathbf{p}_i=p_\theta(x^i=\cdot\,|\,\mathbf{x})=\operatorname{softmax}(\mathbf{z}_i),
\]
where \(W_U\) is the unembedding matrix mapping hidden states to vocabulary logits.

\begin{figure}[t]
\centering
\includegraphics[width=1\linewidth]{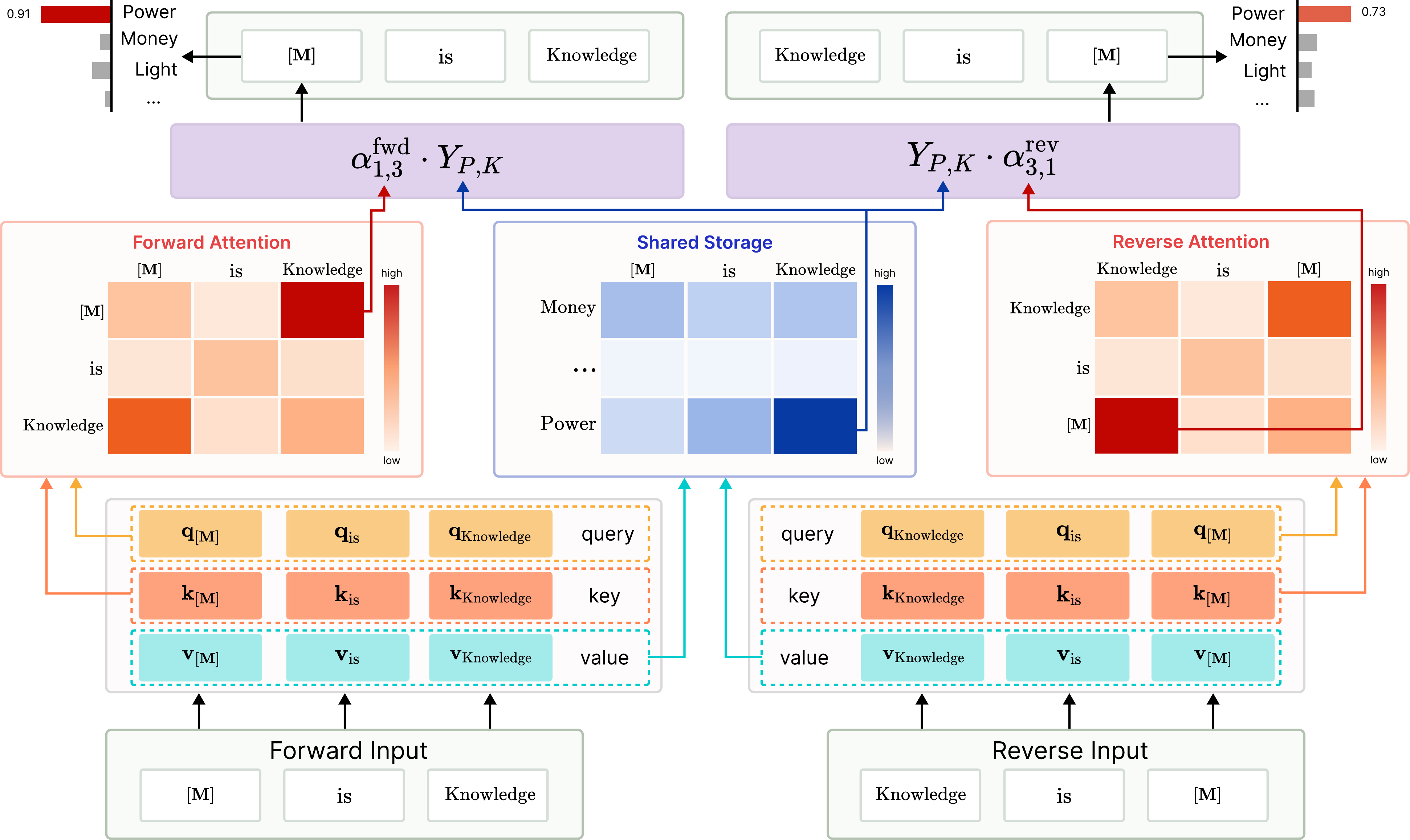}
\vspace{0.2em}
\caption{
\textbf{Storage-routing decomposition under positional reversal.}
For the illustrative statement ``Power is Knowledge,'' random masking can produce ``\(\M\) is Knowledge,'' where evidence token~Knowledge contributes to the target~Power logit through \(\alpha^{\mathrm{fwd}}_{1,3}Y_{P,K}\).
The reverse query ``Knowledge is \(\M\)'' has the corresponding contribution \(\alpha^{\mathrm{rev}}_{3,1}Y_{P,K}\).
Thus, both configurations share the position-invariant storage term \(Y_{P,K}\), while their attention weights determine how strongly this evidence is retrieved.
Under the stated assumptions, the corresponding forward and reverse attention scores are positively correlated in expectation, supporting reverse access to the same evidence.
Actual retrieval also depends on competing context (see App.~\ref{app:retrieval_margin}).
}
\vspace{-0.2em}
\label{fig:workflow}
\end{figure}

\paragraph{Effective reparameterization.}
Following prior analyses of Transformer training dynamics~\citep{tian2023scan,zhu2024towards}, we use the effective weight
\[
Y=W_UW_VW_E.
\]
By combining the input embedding, value projection, and unembedding into a single map, this effective weight isolates the token-to-token evidence stored by the model.
Crucially, relative positional encodings such as RoPE incorporate positional information through query-key attention scores rather than value vectors, making this storage position-invariant.
Specifically, \(Y_{A,B}=\mathbf{e}_A^\top Y\mathbf{e}_B\) captures how strongly input token~\(B\) supports target token~\(A\), independently of its position.
The masked-position logit vector and its \(A\)-coordinate can then be written as
\[
\mathbf{z}_i
=
\sum_{k=1}^L\alpha_{i,k}Y_{:,x^k},
\quad
(\mathbf{z}_i)_A
=
\sum_{k=1}^L\alpha_{i,k}Y_{A,x^k}.
\]
Thus, if the unmasked input token~\(B\) appears at position~\(j\), i.e., \(x^j=B\), the contribution of this token to the \(A\)-logit at masked position~\(i\) can be identified with the term
\[
\alpha_{i,j}Y_{A,B}.
\]

This decomposition separates the position-invariant storage term \(Y_{A,B}\) from the position-dependent routing weight \(\alpha_{i,j}\).
While \(Y_{A,B}\) represents shared evidence, \(\alpha_{i,j}\) determines how strongly the masked position~\(i\) retrieves that evidence from token~\(B\) at position~\(j\).
Thus, changing the positional configuration affects the strength of evidence retrieval without changing the stored evidence itself.

By contrast, with additive absolute positional embeddings~\citep{vaswani2017attention,devlin2019bert}, the representation of input token~\(B\) becomes \(W_E\mathbf{e}_B+\mathbf{p}_j\), making its value vector dependent on position~\(j\).
Consequently, the evidence that token~\(B\) provides for predicting target token~\(A\) generally depends on its position and cannot be captured by a single position-invariant storage term.
We focus on relative encodings, as they are widely adopted in modern LLM architectures, especially RoPE~\citep{su2024roformer}.
Absolute encodings are used for comparison in Tab.~\ref{tab:pe_intervention}.

\section{Theoretical Results: Storage, Routing, and Gradient Alignment}
\label{sec:theory}

In this section, we analyze how forward masked training can improve prediction under positional reversal in the simplified one-layer MDM introduced in Sec.~\ref{sec:setup}.
Building on the storage-routing decomposition \(\alpha_{i,j}Y_{A,B}\), we first show that forward training strengthens the shared evidence term \(Y_{A,B}\), then show how relative positional encodings preserve positively correlated attention scores to token~\(B\) across forward and reverse configurations.
Finally, we analyze the training dynamics at the loss level, showing that the shared-storage component of a forward gradient-descent update decreases the reverse loss to first order.
See Apps.~\ref{app:proofs}, \ref{app:details_attention_correlation}, and \ref{app:additional_theory} for proofs and additional analyses.

\subsection{Storage: Forward Training Reinforces a Reverse-Usable Relation}
\label{sec:shared_y_update}

We first examine how forward masked training updates the storage term \(Y_{A,B}\) shared with the corresponding reverse query.
Consider a forward masked instance ``\([\textnormal{\textbf{M}}]\) is \(B\),'' where token~\(B\) is at position~\(j\) and target~\(A\) is recovered at masked position~\(i\).
The contribution of \(B\) to the \(A\)-logit is represented by \(\alpha_{i,j}^{\mathrm{fwd}}Y_{A,B}\).
In the reverse query, \(B\) appears at position~\(j'\) and \(A\) is recovered at position~\(i'\), yielding the contribution \(\alpha_{i',j'}^{\mathrm{rev}}Y_{A,B}\).
Thus, the two configurations share the same storage term \(Y_{A,B}\), while their attention weights determine how strongly this evidence is retrieved.

For the cross-entropy loss of recovering \(A\) at masked position \(i\) from the partially unmasked input \(\mathbf{x}\),
\[
\mathcal{L}_i=-\log p_i(A)=-\log p_\theta(x^i=A|\mathbf{x}),
\]
we have the following direct gradient calculation.

\begin{proposition}[Forward Masked Training Strengthens the Shared \(B\)-to-\(A\) Relation]
\label{prop:shared_y_update}
Consider the one-layer MDM setup of Sec.~\ref{sec:setup}, with \(Y\) as the effective storage parameter.
If \(x^i=[\textnormal{\textbf{M}}]\) has target token~\(A\) and \(x^j=B\) is an unmasked input token with attention weight \(\alpha_{i,j}>0\), then
\[
\frac{\partial \mathcal{L}_i}{\partial Y_{A,B}}
\le
-\alpha_{i,j}(1-p_i(A))<0
\quad
\text{and}
\quad
\frac{\partial \mathcal{L}_i}{\partial Y_{C,B}}
\ge
\alpha_{i,j}p_i(C)>0
\quad \forall\, C\neq A.
\]
Thus, a gradient-descent step with respect to \(Y\) increases \(Y_{A,B}\) and decreases \(Y_{C,B}\) for all \(C\neq A\).
\end{proposition}

The proposition shows that forward masked training strengthens the shared \(B\)-to-\(A\) evidence that reverse queries can reuse.
Specifically, a gradient-descent step with respect to \(Y\) increases \(Y_{A,B}\) while decreasing \(Y_{C,B}\) for every competing target~\(C\neq A\).
The positive attention weight assigned to token~\(B\) controls the magnitude of these updates but not their signs.

\subsection{Routing: Attention Correlation under Relative Positional Encodings}
\label{sec:attention_correlation}

We now investigate whether the reverse query can still retrieve the shared evidence \(Y_{A,B}\) strengthened by forward masked training.
This depends on the routing term \(\alpha_{i,j}\): even if \(Y_{A,B}\) is shared across positions, reverse prediction benefits only when the masked position assigns sufficient attention to the evidence token~\(B\).
Because the attention weight is obtained by a softmax over all tokens, we study the attention score assigned specifically to \(B\) as the local quantity governing this route (see App.~\ref{app:retrieval_margin} for a precise retrieval condition).
We analyze the Pearson correlation between forward and reverse attention scores under RoPE and ALiBi to assess how well the two configurations preserve their routes to the same evidence token.

\paragraph{RoPE.}
We first analyze RoPE~\citep{su2024roformer}, the dominant relative positional encoding in modern LLMs and the encoding used by the masked diffusion LLMs~\citep{nie2025large,ye2025dream7bdiffusionlarge} studied in this paper.
RoPE injects relative position into attention scores through block-wise rotations in two-dimensional frequency planes.
Consider a forward prompt where \(B\) is \(\Delta_1\) positions to the right of \(\M\), and a reverse prompt where \(B\) is \(\Delta_2\) positions to the left of \(\M\).
We analyze the pre-softmax attention score underlying \(\alpha_{i,j}\), using the masked-token query \(\mathbf{q}_{\M}\) and the evidence-token key \(\mathbf{k}_B\).
The corresponding forward and reverse scores are
\[
S_{\mathrm{fwd}}
=
\mathbf{q}_{\M}^{\top}R(\Delta_1)\mathbf{k}_B,
\quad
S_{\mathrm{rev}}
=
\mathbf{q}_{\M}^{\top}R(-\Delta_2)\mathbf{k}_B,
\]
where \(R(\Delta)\) is the RoPE rotation matrix.
Reversal changes the relative rotation, raising the question of whether the forward and reverse attention scores remain positively correlated.

To analyze their correlation, we introduce a covariance assumption capturing RoPE's rotation geometry and potential query-key dependence induced by training.

\begin{assumption}[RoPE-Compatible Block Covariance]
\label{assump:rope_conditional_covariance}
Let \(D\) be even, and partition \(\mathbf{q}_{\M}\) and \(\mathbf{k}_B\) into \(D/2\) two-dimensional RoPE blocks \(\mathbf{q}_{\M}^{(s)}\) and \(\mathbf{k}_B^{(s)}\).
Assume \(\mathbf{q}_{\M}\sim\mathcal{N}(0,\sigma^2 I_D)\).
Conditioned on \(\mathbf{q}_{\M}\), the key blocks are independent with equal total conditional variance.
Within each block, the query direction and its orthogonal direction are covariance eigenvectors, with at least as much conditional variance along the query direction.
\end{assumption}

Intuitively, this assumption rules out key variation dominated by directions orthogonal to the query.
It is not a property guaranteed by RoPE or established for practical models, but a statistical idealization introduced to derive a tractable closed-form correlation bound.
See App.~\ref{app:relaxed_fixed_query_covariance} for conditions ensuring positive covariance under relaxed assumptions.

\begin{theorem}[RoPE Attention-Score Correlation]
\label{thm:rope_attention_correlation}
If Assumption~\ref{assump:rope_conditional_covariance} holds and \(\Delta_1=\Delta_2=\Delta\), then
\[
\mathbb{E}_{\mathbf{q}_{\M}}
\left[
\operatorname{Corr}
\left(
S_{\mathrm{fwd}},
S_{\mathrm{rev}}
\,|\,\mathbf{q}_{\M}
\right)
\right]
\ge
\frac{1}{D}\operatorname{Tr}(R(2\Delta)).
\]
\end{theorem}

The theorem establishes positive expected forward--reverse attention-score correlation under the stated assumptions.
For standard RoPE frequencies, the lower bound is positive for \(0\le\Delta\le50\) and \(D\in\{64,128,256\}\), with a minimum of \(0.337\) (see App.~\ref{app:rope_trace_bound}).
The equal-distance condition \(\Delta_1=\Delta_2\) is used to obtain a closed-form bound; for unequal distances, numerical estimates of a corresponding bound remain positive over the tested range (see App.~\ref{app:asymmetric_rope_check}).

RoPE does not make forward and reverse attention scores identical, but preserves positive correlation between them.
Although the resulting attention weights also depend on competing tokens, this correlation helps explain how the reverse query can retain access to the same stored \(B\)-to-\(A\) evidence.

\paragraph{ALiBi.}
To show that the routing analysis is not specific to RoPE, we also consider ALiBi~\citep{press2022alibi}.
With ALiBi, position enters as an additive distance bias, so the attention scores between the masked-token query \(\mathbf{q}_{\M}\) and the evidence-token key \(\mathbf{k}_B\) take the form
\[
S_{\mathrm{fwd}}
=
\mathbf{q}_{\M}^{\top}\mathbf{k}_B+b\Delta_1,
\quad
S_{\mathrm{rev}}
=
\mathbf{q}_{\M}^{\top}\mathbf{k}_B-b\Delta_2.
\]
Thus, the two scores share the same content term \(\mathbf{q}_{\M}^{\top}\mathbf{k}_B\) and differ only by constants.

\begin{proposition}[ALiBi Attention-Score Correlation]
\label{prop:alibi_attention_correlation}
For ALiBi, suppose the shared content score \(\mathbf{q}_{\M}^{\top}\mathbf{k}_B\) has nonzero variance.
Then, for any joint distribution of \((\mathbf{q}_{\M},\mathbf{k}_B)\),
\[
\operatorname{Corr}\left(S_{\mathrm{fwd}},S_{\mathrm{rev}}\right)=1.
\]
\end{proposition}

The result is immediate but useful: ALiBi preserves perfect forward--reverse attention-score correlation because positional biases do not affect the shared content term.
Together with Thm.~\ref{thm:rope_attention_correlation}, this identifies an attention-routing mechanism through which relative positional encodings can preserve access to forward-learned evidence, helping explain how MDMs improve reverse prediction.

\subsection{Dynamics: Forward-to-Reverse Gradient Alignment}
\label{sec:gradient_alignment}

We now analyze whether a forward gradient-descent step increases the reverse target probability without training on the reverse prompt.
For intuition, consider the paired prompts ``\(\M\) is \(B\)'' and ``\(B\) is \(\M\),'' both with target~\(A\).
Define
\[
p_{\mathrm{fwd}}(A)=p_\theta\!\left(\M=A\,\middle|\,\M\text{ is }B\right),
\quad
p_{\mathrm{rev}}(A)=p_\theta\!\left(\M=A\,\middle|\,B\text{ is }\M\right).
\]
More generally, let \(\mathbf{x}_{\mathrm{fwd}}\) and \(\mathbf{x}_{\mathrm{rev}}\) denote paired masked prompts with arbitrary additional context tokens, placing \(\M\) before and after token~\(B\), respectively.
Let \(\mathbf{p}_{\mathrm{fwd}}\) and \(\mathbf{p}_{\mathrm{rev}}\) denote their masked-position prediction distributions, both with target~\(A\).
We define the corresponding losses as
\[
\mathcal{L}_{\mathrm{fwd}}=-\log p_{\mathrm{fwd}}(A),
\quad
\mathcal{L}_{\mathrm{rev}}=-\log p_{\mathrm{rev}}(A).
\]
To analyze how a forward gradient-descent step affects the reverse loss, consider the update \(\theta^+=\theta-\eta\nabla_\theta\mathcal{L}_{\mathrm{fwd}}\).
A first-order expansion yields
\[
\mathcal{L}_{\mathrm{rev}}(\theta^+)-\mathcal{L}_{\mathrm{rev}}(\theta)
=
-\eta\left\langle
\nabla_\theta\mathcal{L}_{\mathrm{rev}},\,
\nabla_\theta\mathcal{L}_{\mathrm{fwd}}
\right\rangle+O(\eta^2).
\]
Thus, a positive gradient inner product implies that the forward step decreases the reverse loss, equivalently increasing the reverse probability of target~\(A\), to first order.

We now show that a forward gradient-descent update of the shared storage column \(Y_{:,B}\) decreases the reverse loss to first order.
This extends the storage-strengthening result of Sec.~\ref{sec:shared_y_update} to reverse prediction.
Let \(\alpha_B^{\mathrm{fwd}}\) and \(\alpha_B^{\mathrm{rev}}\) denote the total attention mass assigned to all occurrences of token~\(B\) in the forward and reverse prompts \(\mathbf{x}_{\mathrm{fwd}}\) and \(\mathbf{x}_{\mathrm{rev}}\), respectively.

\begin{theorem}[Shared Storage Column Gradient Alignment]
\label{thm:b_column_gradient_alignment}
Consider the one-layer MDM setup of Sec.~\ref{sec:setup}.
Suppose the unmasked token~\(B\) appears in both forward and reverse prompts, with the same target token~\(A\) at their masked positions.
Then the forward and reverse loss gradients with respect to the shared storage column \(Y_{:,B}\) satisfy
\[
\left\langle
\nabla_{Y_{:,B}}\mathcal{L}_{\mathrm{fwd}},\,
\nabla_{Y_{:,B}}\mathcal{L}_{\mathrm{rev}}
\right\rangle
=
\alpha_B^{\mathrm{fwd}}\alpha_B^{\mathrm{rev}}
\left\langle
\mathbf{p}_{\mathrm{fwd}}-\mathbf{e}_A,\,
\mathbf{p}_{\mathrm{rev}}-\mathbf{e}_A
\right\rangle
>0.
\]
Consequently, a forward gradient-descent update of \(Y_{:,B}\) decreases the reverse loss to first order.
\end{theorem}

This theorem analyzes forward-to-reverse transfer along the shared storage column \(Y_{:,B}\), but does not characterize the full-parameter update, which also involves query-key terms.
We test full-parameter gradient alignment in Secs.~\ref{sec:one_layer_validation} and \ref{sec:large_scale_validation}, finding positive forward--reverse gradient cosine similarity in trained models.
See App.~\ref{app:full_grad_update} for query-key analysis in effective coordinates, App.~\ref{app:original_parameter_alignment} for gradient alignment in the original Transformer parameterization, and App.~\ref{app:probability_transfer} for a probability-level convergence result inspired by \citet{zhu2024towards}.

\section{One-Layer Validation}
\label{sec:one_layer_validation}

We test the storage-routing account in a trained one-layer MDM through attention-score correlation, gradient alignment, and reverse-probability improvement.
We also vary the positional encoding to probe the roles of storage and routing.
See App.~\ref{app:one_layer_exp_details} for details.

\paragraph{Setup.}
We instantiate a one-layer MDM using RADD~\citep{ouyour}, with RoPE as the default positional encoding.
We construct length-\(L\) sequences containing one lowercase--uppercase token pair and fill the remaining \(L-2\) positions with a placeholder \(\texttt{0}\).
The lowercase token is the target, and the uppercase token is the unmasked evidence token.
Training uses only the forward order, where the lowercase target precedes its paired uppercase token; all reversed-order sequences are excluded.
For example, with \(L=3\) and pair \((\texttt{d},\texttt{D})\), the training set contains \(\texttt{dD0}\), \(\texttt{d0D}\), and \(\texttt{0dD}\), but never \(\texttt{Dd0}\), \(\texttt{D0d}\), or \(\texttt{0Dd}\).
The placeholders introduce positional variation without adding semantic cues.

At each checkpoint, we evaluate paired forward and reverse masked queries from the same lowercase--uppercase pair.
Both recover the lowercase target from the same unmasked uppercase evidence token, but in reversed positional configurations.
We measure attention-score correlation, gradient cosine similarity over all trainable parameters, and target-token probabilities under both queries.

\begin{figure}[t]
\centering
\resizebox{\linewidth}{!}{%
    \includegraphics[height=4.8cm,keepaspectratio]{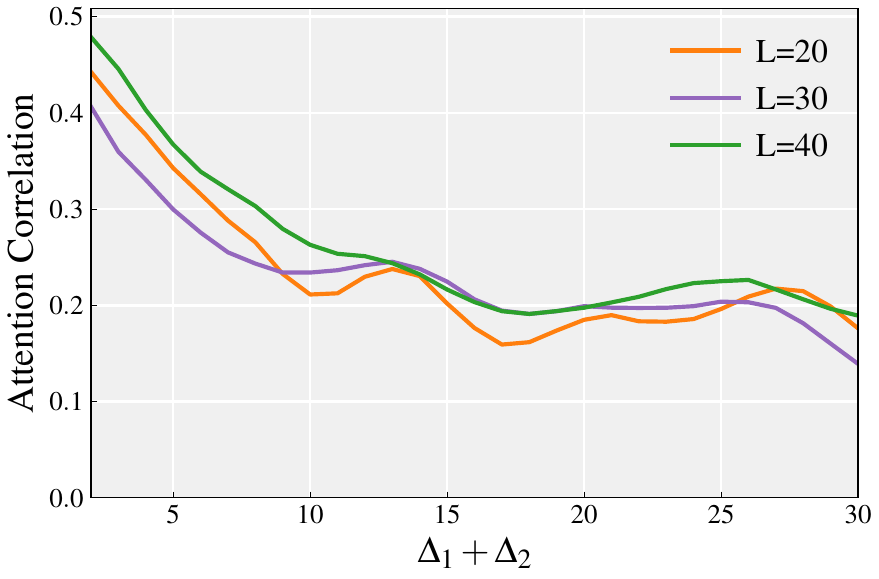}
    \hspace{0.01\linewidth}
    \includegraphics[height=4.8cm,keepaspectratio]{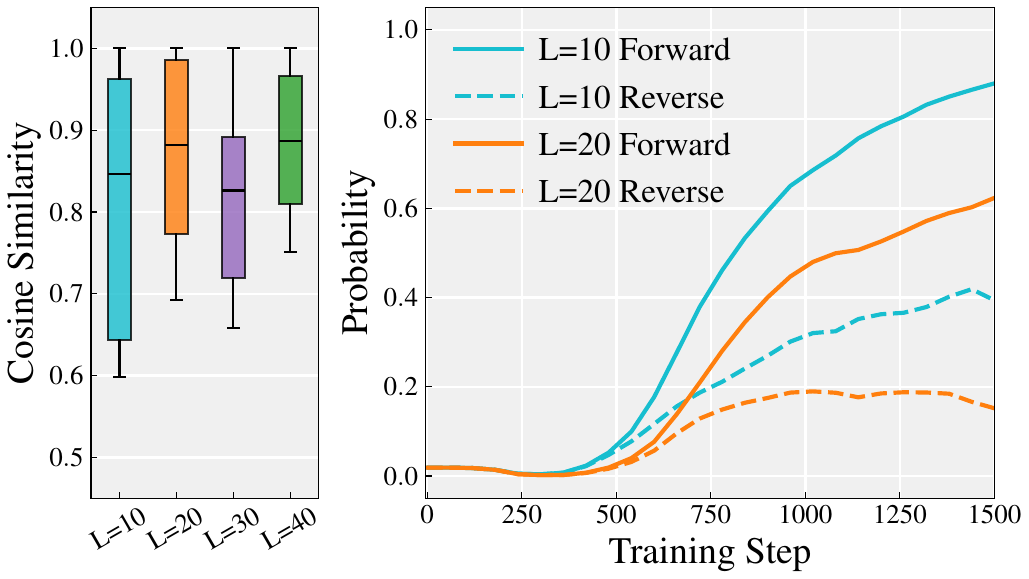}
}
\caption{
\textbf{One-layer validation of storage-routing signatures.}
\textbf{Left:} Pearson correlation between forward and reverse attention scores for the evidence token across total relative distance \(\Delta_1+\Delta_2\).
\textbf{Middle:} Forward--reverse gradient cosine similarity over all trainable parameters during training.
\textbf{Right:} Target-token probabilities under forward and reverse masked queries during training.
}
\vspace{-1em}
\label{fig:toy_validation}
\end{figure}

\begin{wraptable}{r}{0.4\textwidth}
\centering
\caption{
In the one-layer setting, the MDM retains substantial reverse accuracy, whereas the ARM collapses.
}
\vspace{-0.6em}
\renewcommand{\arraystretch}{1.12}
\setlength{\tabcolsep}{10pt} 
\resizebox{\linewidth}{!}
{%
\begin{tabular}{lcccc}
\toprule
& \multicolumn{2}{c}{\textbf{MDM}}
& \multicolumn{2}{c}{\textbf{ARM}} \\
\cmidrule(lr){2-3} \cmidrule(lr){4-5}
\(L\) & Fwd. & Rev. & Fwd. & Rev. \\
\midrule
10 & 99.31 & \textbf{43.10} & 99.83 & 0.00 \\
20 & 97.36 & \textbf{55.70} & 99.80 & 0.00 \\
30 & 96.91 & \textbf{33.89} & 99.93 & 0.00 \\
40 & 97.27 & \textbf{38.37} & 99.93 & 0.00 \\
\bottomrule
\end{tabular}
}
\vspace{-2em}
\label{tab:toy_accuracy}
\end{wraptable}

\paragraph{Results.}
We first compute the attention score between the masked-token query and the uppercase evidence-token key.
Fig.~\ref{fig:toy_validation} (left) shows that the forward and reverse scores remain strongly positively correlated across positional configurations, consistent with Thm.~\ref{thm:rope_attention_correlation}.
This supports the proposed routing mechanism, through which reverse queries can access the same forward-learned evidence.

We next compute the forward--reverse gradient cosine similarity over all trainable parameters, including query-key projections and MLP parameters.
Fig.~\ref{fig:toy_validation} (middle) shows consistently positive values across sequence lengths.
Thus, the gradient alignment identified along the shared storage column in Sec.~\ref{sec:gradient_alignment} is also reflected in the trained model's full-parameter gradients.

Finally, we examine whether these signals accompany improved reverse prediction.
Fig.~\ref{fig:toy_validation} (right) shows that the correct target probability increases on reverse queries under forward-only training.
Together, the attention-score correlation, gradient alignment, and probability improvement support the proposed storage-routing account.
A one-layer GPT-2-style ARM baseline~\citep{radford2019language} learns the forward direction but fails to transfer, as shown in Tab.~\ref{tab:toy_accuracy}.

\begin{wraptable}{r}{0.4\textwidth}
\vspace{-1.25em}
\caption{
Relative encodings transfer better than learned absolute embeddings in this experiment, with ALiBi giving the strongest transfer.
This ordering is consistent with the storage-routing theory.
}
\vspace{-0.3em}
\label{tab:pe_intervention}
\centering
\renewcommand{\arraystretch}{1.12}
\resizebox{\linewidth}{!}{%
\begin{tabular}{llccc}
\toprule
\(L\) & Dir. & Absolute & RoPE & ALiBi \\
\midrule
\multirow{2}{*}{\(10\)}
& Fwd. & 98.3 & 99.3 & 99.7 \\
& Rev. & 11.5 & 43.1 & 99.8 \\
\midrule
\multirow{2}{*}{\(20\)}
& Fwd. & 97.8 & 97.4 & 99.5 \\
& Rev. & 27.8 & 55.7 & 99.5 \\
\bottomrule
\end{tabular}
}
\vspace{-0.5em}
\end{wraptable}

\paragraph{Positional encoding intervention.}
We change only the positional encoding to test the storage and routing components of the theory.
Relative encodings preserve position-invariant value representations, whereas learned absolute embeddings~\citep{gehring2017convolutional,devlin2019bert} introduce position-dependent value contributions, preventing the full \(B\)-to-\(A\) evidence from being represented by a single position-invariant storage term.
Among relative encodings, RoPE and ALiBi differ in routing: RoPE preserves positive expected attention-score correlation under the analyzed conditions, whereas ALiBi preserves perfect score correlation through a shared content term with additive distance biases, as shown in Sec.~\ref{sec:attention_correlation}.
Tab.~\ref{tab:pe_intervention} matches this pattern: learned absolute embeddings fit the forward direction but transfer poorly, RoPE transfers nontrivially but imperfectly, and ALiBi transfers almost perfectly.

Together, these results show that the storage-routing signatures predicted by the analysis under effective reparameterization in Sec.~\ref{sec:theory} are also observed in trained one-layer MDMs.

\section{Large-Scale Validation}
\label{sec:large_scale_validation}

We now test whether the storage-routing signatures observed in the one-layer experiments also appear in multi-layer masked diffusion LLMs.
We present the main results for LLaDA~\citep{nie2025large}; see App.~\ref{app:large_scale_details} for experimental details and additional Dream~\citep{ye2025dream7bdiffusionlarge} results.

\begin{figure}[t]
\centering
\resizebox{\linewidth}{!}{%
    \includegraphics[height=4.2cm,keepaspectratio]{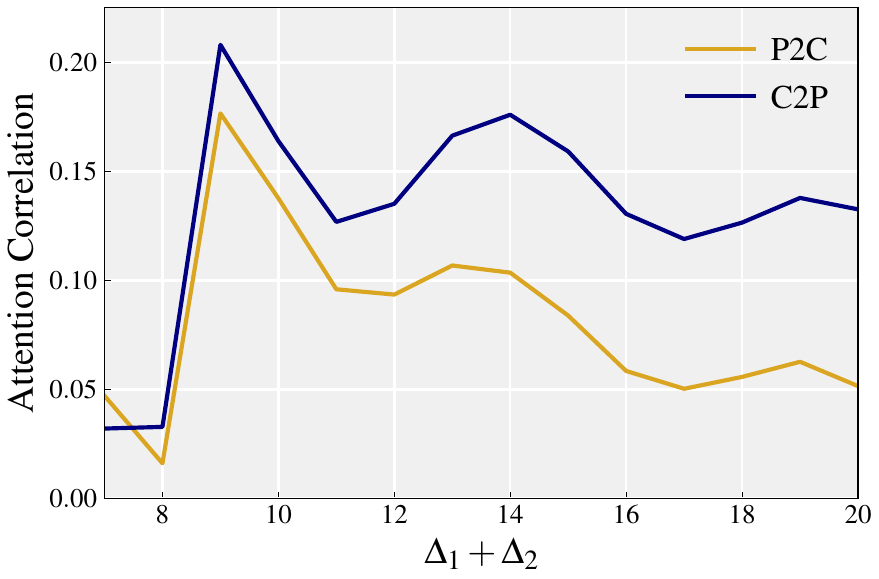}
    \hspace{0.03\linewidth}
    \includegraphics[height=4.2cm,keepaspectratio]{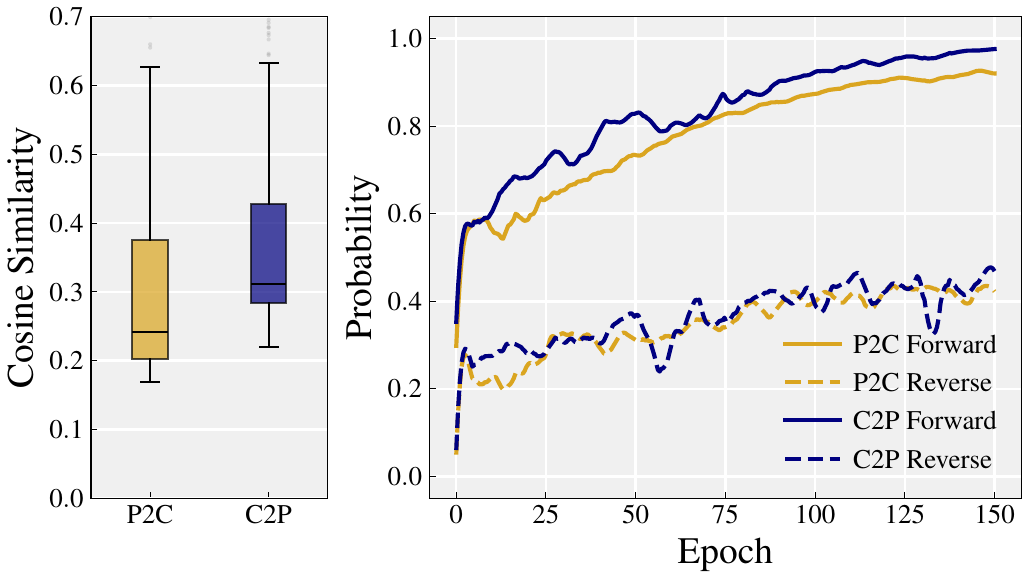}
}
\caption{
\textbf{Large-scale validation of storage-routing signatures in LLaDA.}
\textbf{Left:} Pearson correlation between forward and reverse attention scores from the mask to the evidence entity across relative-distance placements.
\textbf{Middle:} Forward--reverse gradient cosine similarity over trainable parameters during forward-only fine-tuning.
\textbf{Right:} Correct target-entity probabilities under forward and reverse masked queries throughout forward-only fine-tuning on the Parent--Child dataset.
}
\label{fig:large_mechanism}
\end{figure}

\paragraph{Setup.}
We perform forward-only fine-tuning on the Parent--Child dataset~\citep{berglundreversal}.
Tab.~\ref{tab:large_scale_reversal} reports the broader reversal-evaluation results, and we use the Parent--Child task to test the signatures predicted by our theory.
For each relation, we construct paired forward and reverse masked-prediction queries.
For illustration, consider the P2C (Parent-to-Child) statement ``Alice's child is Bob.''
The forward masked query is ``\(\M\)'s child is Bob,'' while the reverse query is ``Bob's parent is \(\M\).''
Both queries recover the same target~Alice from the same evidence token~Bob, but in reversed positional configurations.

\paragraph{Results.}
Fig.~\ref{fig:large_mechanism} (left) shows that forward and reverse attention scores assigned to the evidence entity remain positively correlated across relative-distance placements.
Fig.~\ref{fig:large_mechanism} (middle) shows that forward--reverse gradient cosine similarity over all trainable parameters remains positive.
Finally, Fig.~\ref{fig:large_mechanism} (right) shows that the correct reverse-entity probability steadily increases during forward-only fine-tuning, while remaining below the forward probability.
Together, these results show that the attention-routing and gradient-alignment signatures predicted by our one-layer theory persist in a practical multi-layer MDM, despite additional query-key, MLP, and inter-layer interactions, and accompany improved reverse prediction.

\section{Limitations}
\label{sec:discussion}

Our analysis explains how evidence learned from a forward fact can be reused to answer semantically valid reverse queries, but does not address how inverse-relation semantics are learned or when such reuse leads to harmful over-reversal.
The theoretical guarantees are also limited to a simplified one-layer MDM, while the LLaDA and Dream experiments test the predicted signatures in practical multi-layer models.
See App.~\ref{app:discussion} for further discussion of the modeling framework, task formulation, and scope of our claims.

\section{Conclusion}
\label{sec:conclusion}
We analyzed how MDMs mitigate the reversal curse through shared evidence storage, attention routing, and gradient alignment in a one-layer model, and tested the predicted signatures in controlled and large-scale experiments.
Our findings show that bidirectional masked supervision alone does not explain why evidence learned in one positional configuration remains usable after reversal.
Understanding this transfer requires looking beyond the training objective to how information is stored in model parameters during training and retrieved under different inputs.
In particular, the separation between position-invariant evidence storage and position-dependent attention routing under relative positional encodings offers a useful perspective on how Transformers store and retrieve relational knowledge.
Beyond the reversal curse, this perspective may help explain broader patterns of knowledge retrieval and generalization in MDMs and related architectures.

\begin{ack}
This work was supported by the National Research Foundation of Korea (NRF) grant funded by the Korean government (MSIT) in 2026 (Grant No. RS-2026-25612350, Project Title: ``Development of an End-to-End Design and Validation Framework for Mathematically Principled Next-Generation AI Architectures and Learning Algorithms.''), National Research Foundation of Korea (NRF) grant funded by the Korea government (MSIT) (No. RS-2024-00410005), and the Pioneer Research Center Program through the National Research Foundation of Korea funded by the Ministry of Science, ICT \& Future Planning (RS-2022-NR067569)

The authors declare that, during the 36 months preceding submission, they had no financial relationships outside the submitted work that could reasonably be perceived as influencing this research.
\end{ack}


\bibliographystyle{plainnat}
\bibliography{neurips_2026_references}

@inproceedings{berglundreversal,
  title={The Reversal Curse: LLMs trained on “A is B” fail to learn “B is A”},
  author={Berglund, Lukas and Tong, Meg and Kaufmann, Maximilian and Balesni, Mikita and Stickland, Asa Cooper and Korbak, Tomasz and Evans, Owain},
  booktitle={ICLR},
  year = {2024}
}

@inproceedings{golovnevareverse,
  title={Reverse Training to Nurse the Reversal Curse},
  author={Golovneva, Olga and Allen-Zhu, Zeyuan and Weston, Jason E and Sukhbaatar, Sainbayar},
  booktitle={COLM},
  year={2024}
}

@inproceedings{lin2024delving,
  title={Delving into the Reversal Curse: How Far Can Large Language Models Generalize?},
  author={Lin, Zhengkai and Fu, Zhihang and Liu, Kai and Xie, Liang and Lin, Binbin and Wang, Wenxiao and Cai, Deng and Wu, Yue and Ye, Jieping},
  booktitle={NeurIPS},
  year={2024}
}

@inproceedings{wang2025reversal,
  title={Is the reversal curse a binding problem? uncovering limitations of transformers from a basic generalization failure},
  author={Wang, Boshi and Sun, Huan},
  booktitle={ICLR},
  year={2026}
}

@inproceedings{lu2024rethinking,
  title={Rethinking the reversal curse of LLMs: a prescription from human knowledge reversal},
  author={Lu, Zhicong and Jin, Li and Li, Peiguang and Tian, Yu and Zhang, Linhao and Wang, Sirui and Xu, Guangluan and Tian, Changyuan and Cai, Xunliang},
  booktitle={EMNLP},
  year={2024}
}

@inproceedings{zhu2024towards,
  title={Towards a Theoretical Understanding of the 'Reversal Curse' via Training Dynamics},
  author={Zhu, Hanlin and Huang, Baihe and Zhang, Shaolun and Jordan, Michael and Jiao, Jiantao and Tian, Yuandong and Russell, Stuart J},
  booktitle={NeurIPS},
  year={2024}
}

@article{radford2019language,
  title={Language Models are Unsupervised Multitask Learners},
  author={Radford, Alec and Wu, Jeff and Child, Rewon and Luan, David and Amodei, Dario and Sutskever, Ilya},
  journal={OpenAI Blog},
  year={2019}
}

@inproceedings{brown2020language,
  title={Language models are few-shot learners},
  author={Brown, Tom and Mann, Benjamin and Ryder, Nick and Subbiah, Melanie and Kaplan, Jared and Dhariwal, Prafulla and others},
  booktitle={NeurIPS},
  year={2020}
}

@article{raffel2020exploring,
  title={Exploring the limits of transfer learning with a unified text-to-text transformer},
  author={Raffel, Colin and Shazeer, Noam and Roberts, Adam and Lee, Katherine and Narang, Sharan and Matena, Michael and Zhou, Yanqi and Li, Wei and Liu, Peter J},
  journal={JMLR},
  year={2020}
}

@inproceedings{devlin2019bert,
  title={BERT: Pre-training of deep bidirectional transformers for language understanding},
  author={Devlin, Jacob and Chang, Ming-Wei and Lee, Kenton and Toutanova, Kristina},
  booktitle={NAACL},
  year={2019}
}

@inproceedings{lou2024discrete,
  title={Discrete diffusion modeling by estimating the ratios of the data distribution},
  author={Lou, Aaron and Meng, Chenlin and Ermon, Stefano},
  booktitle={ICML},
  year={2024}
}

@inproceedings{ouyour,
  title={Your Absorbing Discrete Diffusion Secretly Models the Conditional Distributions of Clean Data},
  author={Ou, Jingyang and Nie, Shen and Xue, Kaiwen and Zhu, Fengqi and Sun, Jiacheng and Li, Zhenguo and Li, Chongxuan},
  booktitle={ICLR},
  year = {2025}
}

@inproceedings{
nie2025large,
title={Large Language Diffusion Models},
author={Shen Nie and Fengqi Zhu and Zebin You and Xiaolu Zhang and Jingyang Ou and Jun Hu and JUN ZHOU and Yankai Lin and Ji-Rong Wen and Chongxuan Li},
booktitle={NeurIPS},
year={2025},
}

@inproceedings{shi2024simplified,
  title={Simplified and generalized masked diffusion for discrete data},
  author={Shi, Jiaxin and Han, Kehang and Wang, Zhe and Doucet, Arnaud and Titsias, Michalis},
  booktitle={NeurIPS},
  year={2024}
}

@inproceedings{austin2021structured,
  title={Structured denoising diffusion models in discrete state-spaces},
  author={Austin, Jacob and Johnson, Daniel D and Ho, Jonathan and Tarlow, Daniel and Van Den Berg, Rianne},
  booktitle={NeurIPS},
  year={2021}
}

@inproceedings{sahoo2024simple,
  title={Simple and effective masked diffusion language models},
  author={Sahoo, Subham and Arriola, Marianne and Schiff, Yair and Gokaslan, Aaron and Marroquin, Edgar and Chiu, Justin and Rush, Alexander and Kuleshov, Volodymyr},
  booktitle={NeurIPS},
  year={2024}
}

@article{su2024roformer,
  title={Roformer: Enhanced transformer with rotary position embedding},
  author={Su, Jianlin and Ahmed, Murtadha and Lu, Yu and Pan, Shengfeng and Bo, Wen and Liu, Yunfeng},
  journal={Neurocomputing},
  year={2024}
}

@article{ye2025dream7bdiffusionlarge,
  title={Dream 7B: Diffusion Large Language Models}, 
  author={Jiacheng Ye and Zhihui Xie and Lin Zheng and Jiahui Gao and Zirui Wu and Xin Jiang and Zhenguo Li and Lingpeng Kong},
  journal={arXiv preprint arXiv:2508.15487},
  year={2025}
}

@inproceedings{nie2025scaling,
    title={Scaling up Masked Diffusion Models on Text},
    author={Shen Nie and Fengqi Zhu and Chao Du and Tianyu Pang and Qian Liu and Guangtao Zeng and Min Lin and Chongxuan Li},
    booktitle={ICLR},
    year={2025}
}

@inproceedings{kitouni2024factorizationcursetokenspredict,
      title={The Factorization Curse: Which Tokens You Predict Underlie the Reversal Curse and More}, 
      author={Ouail Kitouni and Niklas Nolte and Diane Bouchacourt and Adina Williams and Mike Rabbat and Mark Ibrahim},
      booktitle={NeurIPS},
      year={2024}
}

@article{radford2018improving,
    title = {Improving language understanding by generative pre-training},
    author = {Radford, Alec and Narasimhan, Karthik and Salimans, Tim and Sutskever, Ilya},
    journal = {OpenAI Blog},
    year = {2018}
}

@inproceedings{vaswani2017attention,
  title={Attention is all you need},
  author={Vaswani, Ashish and Shazeer, Noam and Parmar, Niki and Uszkoreit, Jakob and Jones, Llion and Gomez, Aidan N and Kaiser, {\L}ukasz and Polosukhin, Illia},
  booktitle={NeurIPS},
  year={2017}
}

@inproceedings{hoogeboom2021argmax,
  title={Argmax flows and multinomial diffusion: Learning categorical distributions},
  author={Hoogeboom, Emiel and Nielsen, Didrik and Jaini, Priyank and Forr{\'e}, Patrick and Welling, Max},
  booktitle={NeurIPS},
  year={2021}
}

@inproceedings{campbell2022continuous,
  title={A continuous time framework for discrete denoising models},
  author={Campbell, Andrew and Benton, Joe and De Bortoli, Valentin and Rainforth, Thomas and Deligiannidis, George and Doucet, Arnaud},
  booktitle={NeurIPS},
  year={2022}
}

@inproceedings{
huang2025generalization,
title={Generalization or Hallucination? Understanding Out-of-Context Reasoning in Transformers},
author={Yixiao Huang and Hanlin Zhu and Tianyu Guo and Jiantao Jiao and Somayeh Sojoudi and Michael I. Jordan and Stuart Russell and Song Mei},
booktitle={NeurIPS},
year={2025}
}

@inproceedings{tian2023scan,
  title={Scan and snap: Understanding training dynamics and token composition in 1-layer transformer},
  author={Tian, Yuandong and Wang, Yiping and Chen, Beidi and Du, Simon S},
  booktitle={NeurIPS},
  year={2023}
}

@inproceedings{
press2022alibi,
title={Train Short, Test Long: Attention with Linear Biases Enables Input Length Extrapolation},
author={Ofir Press and Noah Smith and Mike Lewis},
booktitle={ICLR},
year={2022}
}

@inproceedings{gehring2017convolutional,
  title={Convolutional sequence to sequence learning},
  author={Gehring, Jonas and Auli, Michael and Grangier, David and Yarats, Denis and Dauphin, Yann N},
  booktitle={ICML},
  year={2017}
}

@article{dinkelbach1967nonlinear,
 author = {Werner Dinkelbach},
 journal = {Management Science},
 title = {On Nonlinear Fractional Programming},
 year = {1967}
}

@inproceedings{he2021deberta,
  title={DEBERTA: DECODING-ENHANCED BERT WITH DISENTANGLED ATTENTION},
  author={He, Pengcheng and Liu, Xiaodong and Gao, Jianfeng and Chen, Weizhu},
  booktitle={ICLR},
  year={2021}
}

@inproceedings{dai2019transformer,
  title={Transformer-xl: Attentive language models beyond a fixed-length context},
  author={Dai, Zihang and Yang, Zhilin and Yang, Yiming and Carbonell, Jaime G and Le, Quoc and Salakhutdinov, Ruslan},
  booktitle={ACL},
  year={2019}
}

@inproceedings{li2023transformers,
  title={How do transformers learn topic structure: Towards a mechanistic understanding},
  author={Li, Yuchen and Li, Yuanzhi and Risteski, Andrej},
  booktitle={ICML},
  year={2023}
}

@article{li2024one,
  title={One-layer transformer provably learns one-nearest neighbor in context},
  author={Li, Zihao and Cao, Yuan and Gao, Cheng and He, Yihan and Liu, Han and Klusowski, Jason M and Fan, Jianqing and Wang, Mengdi},
  journal={NeurIPS},
  year={2024}
}

@inproceedings{huangtransformers,
  title={How Transformers Learn Regular Language Recognition: A Theoretical Study on Training Dynamics and Implicit Bias},
  author={Huang, Ruiquan and Liang, Yingbin and Yang, Jing},
  booktitle={ICML},
  year={2025}
}

@inproceedings{he2026differ,
  title={Differ: Diffusion entity-relation modeling for reversal curse in diffusion large language models},
  author={He, Shaokai and Wei, Kaiwen and Zeng, Xinyi and Chen, Xiang and Yang, Xue and Li, Zhenyang and Zhong, Jiang and Tian, Yu},
  booktitle={Findings of ACL},
  year={2026}
}

\clearpage
\appendix

\begingroup
\renewcommand{\contentsname}{Contents}
\setcounter{tocdepth}{2}
\tableofcontents
\endgroup

\section{Proofs for the Main Theoretical Results}
\label{app:proofs}

\subsection{One-Layer MDM Setup and Effective Reparameterization}
\label{app:one_layer_setup}

We restate the one-layer setup from Sec.~\ref{sec:setup}, making the dimensions explicit.
Let \(\mathcal{V}\) be the vocabulary of clean tokens, and let
\(\mathbf{x}=x^1x^2\ldots x^L\) be a partially masked sequence of length \(L\), where each token satisfies
\(x^k\in\mathcal{V}\cup\{[\textnormal{\textbf{M}}]\}\).
Let \(i\) be a masked position with \(x^i=[\textnormal{\textbf{M}}]\), and let \(A\in\mathcal{V}\) be the target token to be recovered at this position.

For any token \(x\in\mathcal{V}\cup\{[\textnormal{\textbf{M}}]\}\), let
\(\mathbf{e}_x\in\mathbb{R}^{|\mathcal{V}|+1}\) be its input one-hot vector, including the mask token.
When indexing output logits or subtracting a target one-hot vector from an output distribution, the same notation denotes the corresponding vector in \(\mathbb{R}^{|\mathcal{V}|}\); its dimension is determined by context.
Let
\[
W_E\in\mathbb{R}^{D\times (|\mathcal{V}|+1)}
\]
be the token embedding matrix, and define the token embedding
\[
\mathbf{h}_x=W_E\mathbf{e}_x\in\mathbb{R}^D.
\]
Given query, key, and value matrices
\[
W_Q,W_K,W_V\in\mathbb{R}^{D\times D},
\]
we define
\[
\mathbf{q}_x=W_Q\mathbf{h}_x,\quad
\mathbf{k}_x=W_K\mathbf{h}_x,\quad
\mathbf{v}_x=W_V\mathbf{h}_x.
\]

For relative positional encodings such as RoPE or ALiBi, position affects the query-key attention score but not the value vector.
We write the attention score from masked position \(i\) to position \(k\) as
\[
S_{i,k}
=
S(\mathbf{q}_{x^i},\mathbf{k}_{x^k},k-i).
\]
The corresponding attention weight is
\[
\alpha_{i,k}
=
\frac{\exp(S_{i,k}/\sqrt{D})}
{\sum_{k'=1}^L\exp(S_{i,k'}/\sqrt{D})}.
\]
For finite scores, \(\alpha_{i,k}>0\) and \(\sum_{k=1}^L\alpha_{i,k}=1\).

Let
\[
W_U\in\mathbb{R}^{|\mathcal{V}|\times D}
\]
be the unembedding matrix mapping hidden states to logits over clean vocabulary tokens.
The logit vector and prediction distribution at the masked position are
\[
\mathbf{z}_i
=
W_U\sum_{k=1}^L\alpha_{i,k}\mathbf{v}_{x^k},
\quad
\mathbf{p}_i
=
p_\theta(x^i=\cdot|\mathbf{x})
=
\operatorname{softmax}(\mathbf{z}_i).
\]
Thus \(\mathbf{z}_i,\mathbf{p}_i\in\mathbb{R}^{|\mathcal{V}|}\).

\paragraph{Effective reparameterization.}
Following prior analyses of Transformer training dynamics~\citep{tian2023scan,zhu2024towards}, we introduce the effective weights
\[
Y=W_UW_VW_E\in\mathbb{R}^{|\mathcal{V}|\times (|\mathcal{V}|+1)},
\quad
Q=W_QW_E\in\mathbb{R}^{D\times (|\mathcal{V}|+1)},
\quad
K=W_KW_E\in\mathbb{R}^{D\times (|\mathcal{V}|+1)}.
\]
Then
\[
\mathbf{q}_x=Q\mathbf{e}_x,
\quad
\mathbf{k}_x=K\mathbf{e}_x,
\]
and the masked-position logits can be written as
\[
\mathbf{z}_i
=
\sum_{k=1}^L\alpha_{i,k}Y\mathbf{e}_{x^k},
\quad
(\mathbf{z}_i)_A
=
\sum_{k=1}^L\alpha_{i,k}Y_{A,x^k}.
\]
Although the main text uses only \(Y\), we also introduce \(Q\) and \(K\) here to make the query-key gradient analysis in App.~\ref{app:full_grad_update} explicit.
Therefore, if \(x^j=B\), the contribution of the unmasked input token \(B\) to the \(A\)-logit at masked position \(i\) is
\[
\alpha_{i,j}Y_{A,B}.
\]
This is the storage-routing decomposition from Sec.~\ref{sec:setup}.

\paragraph{Role of the effective reparameterization.}
The effective reparameterization is an analytic coordinate system rather than the original training parameterization.
For effective-coordinate derivatives and gradient dynamics, we treat \(Y,Q,K\) as independent parameters.
Following \citet{tian2023scan,zhu2024towards}, we collect the input-to-value-to-logit path into \(Y=W_UW_VW_E\), so that the stored contribution of input token \(B\) to target logit \(A\) appears directly as \(Y_{A,B}\).
We also collect the input-to-query and input-to-key maps into \(Q=W_QW_E\) and \(K=W_KW_E\), which determine attention routing.
Thus, the effective coordinates make the storage-routing decomposition explicit: \(Y\) represents shared storage, while \((Q,K)\) determine how attention routes the stored evidence.

We do not claim that gradient descent in these effective parameters exactly preserves the training dynamics of the factorized original parameters.
Rather, the effective coordinates expose the mechanism in its cleanest form.
App.~\ref{app:original_parameter_alignment} provides a complementary gradient decomposition in the factorized parameterization for an explicit reduced model, separating the sign-determined contribution from terms whose signs depend on the learned parameters.

\subsection{Proof of Proposition~\ref{prop:shared_y_update}}

Recall that
\[
\mathbf{z}_i=\sum_{k=1}^L\alpha_{i,k}Y\mathbf{e}_{x^k},
\]
where \(Y\in\mathbb{R}^{|\mathcal{V}|\times(|\mathcal{V}|+1)}\) is the effective storage parameter, and \(\mathbf{p}_i=\operatorname{softmax}(\mathbf{z}_i)\).
For the loss
\[
\mathcal{L}_i
=
-\log p_i(A)
=
-(\mathbf{z}_i)_A+\log\sum_{C\in\mathcal{V}}\exp((\mathbf{z}_i)_C),
\]
we have
\[
\frac{\partial \mathcal{L}_i}{\partial (\mathbf{z}_i)_C}
=
p_i(C)-\delta_{C,A}.
\]
Moreover,
\[
(\mathbf{z}_i)_C
=
\sum_{k=1}^L\alpha_{i,k}Y_{C,x^k},
\quad
\frac{\partial (\mathbf{z}_i)_C}{\partial Y_{C,B}}
=
\sum_{k:x^k=B}\alpha_{i,k}.
\]
Let
\[
\beta_{i,B}:=\sum_{k:x^k=B}\alpha_{i,k}.
\]
If \(x^j=B\), then \(\beta_{i,B}\ge \alpha_{i,j}\). This is the step that turns the exact derivative into the inequalities stated in the proposition.

Combining the two derivatives gives
\[
\frac{\partial \mathcal{L}_i}{\partial Y_{C,B}}
=
\left(p_i(C)-\delta_{C,A}\right)\beta_{i,B}.
\]
Hence
\[
\frac{\partial \mathcal{L}_i}{\partial Y_{A,B}}
=
-\beta_{i,B}(1-p_i(A))
\le
-\alpha_{i,j}(1-p_i(A))
<0,
\]
and for \(C\neq A\),
\[
\frac{\partial \mathcal{L}_i}{\partial Y_{C,B}}
=
\beta_{i,B}p_i(C)
\ge
\alpha_{i,j}p_i(C)
>0.
\]
Therefore, a gradient step increases \(Y_{A,B}\) and decreases \(Y_{C,B}\) for every \(C\neq A\).
\qed

\subsection{Proof of Theorem~\ref{thm:rope_attention_correlation}}
\label{app:proof_rope_attention_correlation}

\paragraph{Formal covariance assumption.}
We use the following formal version of Assumption~\ref{assump:rope_conditional_covariance}.
Let \(D\) be even, and decompose the RoPE representation into \(D/2\) two-dimensional planes:
\[
\mathbf{q}_{\M}
=
\left(
(\mathbf{q}_{\M}^{(1)})^\top,\ldots,
(\mathbf{q}_{\M}^{(D/2)})^\top
\right)^\top,
\quad
\mathbf{k}_{B}
=
\left(
(\mathbf{k}_{B}^{(1)})^\top,\ldots,
(\mathbf{k}_{B}^{(D/2)})^\top
\right)^\top.
\]
Assume \(\mathbf{q}_{\M}\sim\mathcal{N}(0,\sigma^2 I_D)\) with \(\sigma>0\).
Conditioned on \(\mathbf{q}_{\M}\), the blocks \(\mathbf{k}_{B}^{(s)}\) are independent across RoPE planes, and
\[
\operatorname{Cov}(\mathbf{k}_B|\mathbf{q}_{\M})
=
\operatorname{diag}(\Sigma_1,\ldots,\Sigma_{D/2}).
\]
Assume a common scale \(\tau^2>0\) and, for each plane~\(s\), a coefficient
\(\rho_s=\rho_s(\mathbf{q}_{\M})\in[0,1]\) such that
\[
\Sigma_s
=
\tau^2
\left(
\frac{1-\rho_s}{2}I_2
+
\rho_s
\frac{
\mathbf{q}_{\M}^{(s)}(\mathbf{q}_{\M}^{(s)})^\top
}{
\|\mathbf{q}_{\M}^{(s)}\|^2
}
\right),
\]
with the convention that the last term is defined arbitrarily on the measure-zero event
\(\mathbf{q}_{\M}^{(s)}=\mathbf{0}\).
All correlation statements concern nonzero conditional score variances.

This assumption matches the two-dimensional plane structure of RoPE.
The block-diagonal condition removes cross-plane covariance terms, and the equal-trace condition
\(\operatorname{Tr}(\Sigma_s)=\tau^2\) prevents the bound from being dominated by a few frequency planes.
Within each plane, \(\rho_s\) allows the covariance to range from isotropic \((\rho_s=0)\) to query-aligned \((\rho_s=1)\).
Thus the assumption does not require identical covariance blocks or a fixed alignment strength across planes; it assumes covariance eigenvectors aligned with the query and its orthogonal direction, and excludes larger variance in the orthogonal direction.

\paragraph{Proof.}
Let \(\mathbf{q}=\mathbf{q}_{\M}\) and \(\mathbf{k}=\mathbf{k}_B\).
For RoPE~\citep{su2024roformer}, define
\[
R_s(\Delta)
=
\begin{pmatrix}
\cos(\theta_s\Delta) & -\sin(\theta_s\Delta)\\
\sin(\theta_s\Delta) & \cos(\theta_s\Delta)
\end{pmatrix},
\quad
R(\Delta)
=
\operatorname{diag}\left(R_1(\Delta),\ldots,R_{D/2}(\Delta)\right),
\]
where the standard RoPE frequencies are
\[
\theta_s=\theta_{\mathrm{base}}^{-2(s-1)/D},
\quad
\theta_{\mathrm{base}}=10000.
\]
In the equal-distance case \(\Delta_1=\Delta_2=\Delta\), the attention scores assigned to evidence token \(B\) are
\[
S_{\mathrm{fwd}}
=
\sum_{s=1}^{D/2}
(\mathbf{q}^{(s)})^\top R_s(\Delta)\mathbf{k}^{(s)},
\quad
S_{\mathrm{rev}}
=
\sum_{s=1}^{D/2}
(\mathbf{q}^{(s)})^\top R_s(-\Delta)\mathbf{k}^{(s)}.
\]

Condition on \(\mathbf{q}\).
By the block independence assumption, cross-plane covariance terms vanish. Therefore,
\[
\operatorname{Cov}(S_{\mathrm{fwd}},S_{\mathrm{rev}}|\mathbf{q})
=
\sum_{s=1}^{D/2}
(\mathbf{q}^{(s)})^\top
R_s(\Delta)\Sigma_s R_s(-\Delta)^\top
\mathbf{q}^{(s)}.
\]
Since \(R_s(-\Delta)^\top=R_s(\Delta)\), the covariance becomes
\[
\operatorname{Cov}(S_{\mathrm{fwd}},S_{\mathrm{rev}}|\mathbf{q})
=
\frac{\tau^2}{2}
\sum_{s=1}^{D/2}
\|\mathbf{q}^{(s)}\|^2
\left[
\cos(2\theta_s\Delta)+\rho_s
\right].
\]
The calculation uses
\[
(\mathbf{q}^{(s)})^\top R_s(2\Delta)\mathbf{q}^{(s)}
=
\|\mathbf{q}^{(s)}\|^2\cos(2\theta_s\Delta),
\quad
\frac{
\left((\mathbf{q}^{(s)})^\top R_s(\Delta)\mathbf{q}^{(s)}\right)^2
}{
\|\mathbf{q}^{(s)}\|^2
}
=
\|\mathbf{q}^{(s)}\|^2\cos^2(\theta_s\Delta).
\]

Similarly,
\[
\operatorname{Var}(S_{\mathrm{fwd}}|\mathbf{q})
=
\sum_{s=1}^{D/2}
(\mathbf{q}^{(s)})^\top
R_s(\Delta)\Sigma_s R_s(\Delta)^\top
\mathbf{q}^{(s)},
\]
so
\[
\operatorname{Var}(S_{\mathrm{fwd}}|\mathbf{q})
=
\frac{\tau^2}{2}
\sum_{s=1}^{D/2}
\|\mathbf{q}^{(s)}\|^2
\left[
1+\rho_s\cos(2\theta_s\Delta)
\right].
\]
The same expression holds for \(\operatorname{Var}(S_{\mathrm{rev}}|\mathbf{q})\), since cosine is even.

Define
\[
w_s
=
\frac{\|\mathbf{q}^{(s)}\|^2}{\|\mathbf{q}\|^2},
\quad
\gamma_s
=
\cos(2\theta_s\Delta).
\]
Then
\[
\operatorname{Corr}(S_{\mathrm{fwd}},S_{\mathrm{rev}}|\mathbf{q})
=
\frac{
\sum_{s=1}^{D/2}w_s(\gamma_s+\rho_s)
}{
1+\sum_{s=1}^{D/2}w_s\rho_s\gamma_s
}.
\]
The denominator is positive whenever the conditional variances are nonzero.

We lower bound this expression. Since
\[
\left|\sum_{s=1}^{D/2}w_s\gamma_s\right|\le 1,
\quad
\left|\sum_{s=1}^{D/2}w_s\rho_s\gamma_s\right|
\le
\sum_{s=1}^{D/2}w_s\rho_s,
\]
we have
\[
\left(
\sum_{s=1}^{D/2}w_s\gamma_s
\right)
\left(
\sum_{s=1}^{D/2}w_s\rho_s\gamma_s
\right)
\le
\sum_{s=1}^{D/2}w_s\rho_s.
\]
Hence
\[
\operatorname{Corr}(S_{\mathrm{fwd}},S_{\mathrm{rev}}|\mathbf{q})
\ge
\sum_{s=1}^{D/2}w_s\gamma_s
=
\sum_{s=1}^{D/2}
w_s\cos(2\theta_s\Delta).
\]

Finally, since \(\mathbf{q}\sim\mathcal{N}(0,\sigma^2I_D)\), the quantities
\(\|\mathbf{q}^{(s)}\|^2\) are i.i.d. across RoPE planes.
Thus the weights \(w_s\) are exchangeable and satisfy
\[
\mathbb{E}_{\mathbf{q}}[w_s]
=
\frac{2}{D}.
\]
Taking expectations gives
\[
\mathbb{E}_{\mathbf{q}}
\left[
\operatorname{Corr}(S_{\mathrm{fwd}},S_{\mathrm{rev}}|\mathbf{q})
\right]
\ge
\frac{2}{D}
\sum_{s=1}^{D/2}
\cos(2\theta_s\Delta).
\]
Since each two-dimensional block satisfies
\[
\operatorname{Tr}(R_s(2\Delta))
=
2\cos(2\theta_s\Delta),
\]
the right-hand side equals
\[
\frac{1}{D}\operatorname{Tr}(R(2\Delta)).
\]
This proves the theorem.
\qed

\subsection{Proof of Proposition~\ref{prop:alibi_attention_correlation}}

For ALiBi, the informative-token scores can be written as
\[
S_{\mathrm{fwd}}
=
\mathbf{q}_{\M}^{\top}\mathbf{k}_B+b\Delta_1,
\quad
S_{\mathrm{rev}}
=
\mathbf{q}_{\M}^{\top}\mathbf{k}_B-b\Delta_2,
\]
where \(b,\Delta_1,\Delta_2\) are fixed.
The same proof applies to any fixed additive positional biases, including symmetric distance penalties.
Thus the two scores differ from the same random content score
\(\mathbf{q}_{\M}^{\top}\mathbf{k}_B\) only by additive constants.
Hence
\[
\operatorname{Var}(S_{\mathrm{fwd}})
=
\operatorname{Var}(S_{\mathrm{rev}})
=
\operatorname{Var}(\mathbf{q}_{\M}^{\top}\mathbf{k}_B),
\quad
\operatorname{Cov}(S_{\mathrm{fwd}},S_{\mathrm{rev}})
=
\operatorname{Var}(\mathbf{q}_{\M}^{\top}\mathbf{k}_B).
\]
When \(\operatorname{Var}(\mathbf{q}_{\M}^{\top}\mathbf{k}_B)>0\), these identities immediately give
\[
\operatorname{Corr}(S_{\mathrm{fwd}},S_{\mathrm{rev}})=1.
\]
\qed

\subsection{Proof of Theorem~\ref{thm:b_column_gradient_alignment}}

Let \(\mathbf{x}_{\mathrm{fwd}}\) and \(\mathbf{x}_{\mathrm{rev}}\) be paired masked prompts with the same target token \(A\) and the same unmasked evidence token \(B\).
Let \(\mathbf{p}_{\mathrm{fwd}}\) and \(\mathbf{p}_{\mathrm{rev}}\) be the prediction distributions at the corresponding masked positions, and define
\[
\mathcal{L}_{\mathrm{fwd}}=-\log p_{\mathrm{fwd}}(A),
\quad
\mathcal{L}_{\mathrm{rev}}=-\log p_{\mathrm{rev}}(A).
\]

For any prompt with prediction distribution \(\mathbf{p}\), target \(A\), and total attention mass \(\alpha_B\) assigned to all occurrences of token \(B\), the \(B\)-column gradient is
\[
\nabla_{Y_{:,B}}\mathcal{L}
=
\alpha_B(\mathbf{p}-\mathbf{e}_A).
\]
Therefore,
\[
\nabla_{Y_{:,B}}\mathcal{L}_{\mathrm{fwd}}
=
\alpha_B^{\mathrm{fwd}}(\mathbf{p}_{\mathrm{fwd}}-\mathbf{e}_A),
\quad
\nabla_{Y_{:,B}}\mathcal{L}_{\mathrm{rev}}
=
\alpha_B^{\mathrm{rev}}(\mathbf{p}_{\mathrm{rev}}-\mathbf{e}_A).
\]
Taking the inner product gives
\[
\left\langle
\nabla_{Y_{:,B}}\mathcal{L}_{\mathrm{fwd}},
\nabla_{Y_{:,B}}\mathcal{L}_{\mathrm{rev}}
\right\rangle
=
\alpha_B^{\mathrm{fwd}}\alpha_B^{\mathrm{rev}}
\left\langle
\mathbf{p}_{\mathrm{fwd}}-\mathbf{e}_A,
\mathbf{p}_{\mathrm{rev}}-\mathbf{e}_A
\right\rangle.
\]
It remains to check that the last inner product is positive. Expanding over the vocabulary,
\[
\left\langle
\mathbf{p}_{\mathrm{fwd}}-\mathbf{e}_A,
\mathbf{p}_{\mathrm{rev}}-\mathbf{e}_A
\right\rangle
=
(1-p_{\mathrm{fwd}}(A))(1-p_{\mathrm{rev}}(A))
+
\sum_{C\neq A}
p_{\mathrm{fwd}}(C)p_{\mathrm{rev}}(C)
>0.
\]
Since \(B\) appears in both prompts and the attention scores are finite,
\(\alpha_B^{\mathrm{fwd}},\alpha_B^{\mathrm{rev}}>0\).
Hence
\[
\left\langle
\nabla_{Y_{:,B}}\mathcal{L}_{\mathrm{fwd}},
\nabla_{Y_{:,B}}\mathcal{L}_{\mathrm{rev}}
\right\rangle
>0.
\]
\qed

Consequently, a forward gradient step restricted to the shared column \(Y_{:,B}\) decreases the reverse loss to first order:
\[
\mathcal{L}_{\mathrm{rev}}
\left(
Y_{:,B}-\eta\nabla_{Y_{:,B}}\mathcal{L}_{\mathrm{fwd}}
\right)
-
\mathcal{L}_{\mathrm{rev}}(Y_{:,B})
=
-\eta
\left\langle
\nabla_{Y_{:,B}}\mathcal{L}_{\mathrm{rev}},
\nabla_{Y_{:,B}}\mathcal{L}_{\mathrm{fwd}}
\right\rangle
+
O(\eta^2)
<0
\]
for sufficiently small \(\eta>0\).

\section{Details on Attention Correlation Bounds}
\label{app:details_attention_correlation}
\subsection{Numerical Positivity of the RoPE Trace Bound}
\label{app:rope_trace_bound}

Thm.~\ref{thm:rope_attention_correlation} lower bounds the expected
forward--reverse attention-score correlation by
\[
\frac{1}{D}\operatorname{Tr}(R(2\Delta)).
\]
Thus, to show that the theorem gives a nontrivial positive correlation in the
distance range used in our experiments, it remains to check that this trace
term is positive for standard RoPE frequencies.

\begin{table}[H]
\caption{
Minimum of \(\operatorname{Tr}(R(2\Delta))/D\) over integer
\(0\le \Delta\le 50\).
}
\vspace{0.5em}
\label{tab:rope_trace_bound}
\centering
\small

\setlength{\tabcolsep}{5.5pt}
\begin{tabular}{lccc}
\toprule
\(D\) & 64 & 128 & 256 \\
\midrule
Minimum bound & 0.337 & 0.384 & 0.410 \\
\bottomrule
\end{tabular}
\end{table}

For RoPE with base frequency \(\theta_{\mathrm{base}}=10000\), the normalized
trace is
\[
\frac{1}{D}\operatorname{Tr}(R(2\Delta))
=
\frac{2}{D}
\sum_{s=1}^{D/2}
\cos\left(
2\Delta\cdot 10000^{-2(s-1)/D}
\right).
\]
This quantity is exact and can be evaluated directly for any finite \(D\).
Tab.~\ref{tab:rope_trace_bound} reports the minimum over integer
\(0\le \Delta\le 50\) for \(D\in\{64,128,256\}\).
The bound is positive throughout this range for all tested dimensions.

The large-\(D\) behavior is also consistent with this numerical pattern.
The normalized trace is a Riemann sum:
\[
\frac{2}{D}
\sum_{s=1}^{D/2}
\cos\left(
2\Delta\cdot 10000^{-2(s-1)/D}
\right)
=
\int_0^1
\cos\left(
\frac{2\Delta}{10000^x}
\right)
dx
+
O\left(\frac{1}{D}\right).
\]
Using the cosine integral
\[
\operatorname{Ci}(x)
=
-\int_x^\infty \frac{\cos t}{t}\,dt,
\]
for \(\Delta>0\), this limiting integral can be written as
\[
\int_0^1
\cos\left(
\frac{2\Delta}{10000^x}
\right)
dx
=
\frac{
\operatorname{Ci}(2\Delta)
-
\operatorname{Ci}(2\Delta/10000)
}{
\log 10000
}.
\]
At \(\Delta=0\), the integral equals one by continuity.
Numerically, this limiting approximation is positive over the same range; at
\(\Delta=50\), it is approximately \(0.437\).
Consequently, Thm.~\ref{thm:rope_attention_correlation} gives a positive
lower bound in the distance regime considered here.

\subsection{A High-Probability Bound for RoPE Correlation}
\label{app:rope_cantelli_bound}

Thm.~\ref{thm:rope_attention_correlation} gives a positive lower bound on the
expected forward--reverse attention-score correlation in the distance range of
interest.
Here we show that this positivity is not only an average-case statement: the
conditional correlation is also bounded away from zero with high probability.

Using the notation from the proof of Thm.~\ref{thm:rope_attention_correlation},
the conditional correlation satisfies
\[
\operatorname{Corr}(S_{\mathrm{fwd}},S_{\mathrm{rev}}|\mathbf{q})
\ge
\sum_{s=1}^m w_s\cos(2\Delta\theta_s),
\quad
m=\frac{D}{2},
\]
where
\[
w_s
=
\frac{\|\mathbf{q}^{(s)}\|^2}{\|\mathbf{q}\|^2}.
\]
Since \(\mathbf{q}\sim\mathcal{N}(0,\sigma^2I_D)\), the weights
\((w_1,\ldots,w_m)\) follow \(\operatorname{Dirichlet}(1,\ldots,1)\).
Thus
\[
\mathbb{E}[w_s]=\frac{1}{m},
\quad
\mathbb{E}[w_sw_t]
=
\frac{1+\mathbf{1}_{s=t}}{m(m+1)}.
\]

Define
\[
E_\Delta
=
\frac{1}{m}
\sum_{s=1}^m
\cos(2\Delta\theta_s),
\]
and
\[
V_\Delta
=
\frac{1}{m+1}
\left[
\frac{1}{m}
\sum_{s=1}^m
\cos^2(2\Delta\theta_s)
-
\left(
\frac{1}{m}
\sum_{s=1}^m
\cos(2\Delta\theta_s)
\right)^2
\right].
\]
Then
\[
\mathbb{E}
\left[
\sum_{s=1}^m w_s\cos(2\Delta\theta_s)
\right]
=
E_\Delta,
\]
and
\[
\operatorname{Var}
\left[
\sum_{s=1}^m w_s\cos(2\Delta\theta_s)
\right]
=
V_\Delta.
\]

For any threshold \(0<c<E_\Delta\), Cantelli's inequality gives
\[
\Pr\left[
\sum_{s=1}^m w_s\cos(2\Delta\theta_s)\ge c
\right]
\ge
\frac{(E_\Delta-c)^2}{V_\Delta+(E_\Delta-c)^2}.
\]
Since the conditional correlation is lower bounded by this weighted sum,
\[
\Pr\left[
\operatorname{Corr}(S_{\mathrm{fwd}},S_{\mathrm{rev}}|\mathbf{q})\ge c
\right]
\ge
\frac{(E_\Delta-c)^2}{V_\Delta+(E_\Delta-c)^2}.
\]

\begin{table}[H]
\centering
\small
\setlength{\tabcolsep}{4pt}

\caption{
Cantelli lower bounds for
\(\Pr[\operatorname{Corr}(S_{\mathrm{fwd}},S_{\mathrm{rev}}|\mathbf{q})\ge c]\)
under standard RoPE frequencies.
}
\vspace{0.5em}
\label{tab:cantelli_corr_bounds}
\begin{tabular}{c|ccc|ccc}
\toprule
& \multicolumn{3}{c|}{\(D/2=64\)}
& \multicolumn{3}{c}{\(D/2=128\)} \\
\cmidrule(lr){2-4} \cmidrule(l){5-7}
\(\Delta\)
& \(c=0.10\) & \(c=0.20\) & \(c=0.30\)
& \(c=0.10\) & \(c=0.20\) & \(c=0.30\) \\
\midrule
10 & 0.978 & 0.966 & 0.941 & 0.989 & 0.983 & 0.970 \\
20 & 0.959 & 0.930 & 0.859 & 0.982 & 0.969 & 0.938 \\
30 & 0.947 & 0.905 & 0.791 & 0.977 & 0.958 & 0.906 \\
40 & 0.942 & 0.893 & 0.753 & 0.974 & 0.952 & 0.885 \\
50 & 0.956 & 0.922 & 0.828 & 0.973 & 0.949 & 0.874 \\
\bottomrule
\end{tabular}
\end{table}

Tab.~\ref{tab:cantelli_corr_bounds} shows that the lower bound remains high
for the distances used in our experiments.
For example, when \(\Delta=20\), \(c=0.3\), and \(m=128\), we obtain
\[
\Pr\left[
\operatorname{Corr}(S_{\mathrm{fwd}},S_{\mathrm{rev}}|\mathbf{q})\ge 0.3
\right]
\ge
0.938.
\]

\subsection{Numerical Check for Asymmetric RoPE Distances}
\label{app:asymmetric_rope_check}

This subsection provides the numerical check referenced in
the discussion in Sec.~\ref{sec:attention_correlation}.
Thm.~\ref{thm:rope_attention_correlation} gives a closed-form lower bound in
the equal-distance case \(\Delta_1=\Delta_2\).
We now check that the same positive-correlation behavior persists when the
forward and reverse relative distances are asymmetric, i.e.,
\(\Delta_1\neq\Delta_2\).

Let \(m=D/2\) be the number of RoPE planes.
For each plane \(s\), define
\[
a_s=\cos((\Delta_1+\Delta_2)\theta_s),
\quad
b_s=\cos((\Delta_1-\Delta_2)\theta_s),
\]
and
\[
c_s=\cos(2\Delta_1\theta_s),
\quad
d_s=\cos(2\Delta_2\theta_s).
\]
Using the same conditional covariance structure as in
Assumption~\ref{assump:rope_conditional_covariance}, the same two-dimensional
calculation as in App.~\ref{app:proof_rope_attention_correlation} gives
\[
\operatorname{Corr}(S_{\mathrm{fwd}},S_{\mathrm{rev}}|\mathbf{q})
=
\frac{
\sum_{s=1}^m w_s(a_s+\rho_s b_s)
}{
\sqrt{
\left(1+\sum_{s=1}^m w_s\rho_s c_s\right)
\left(1+\sum_{s=1}^m w_s\rho_s d_s\right)
}
},
\]
where
\[
w_s=\frac{\|\mathbf{q}^{(s)}\|^2}{\|\mathbf{q}\|^2},
\quad
\rho_s\in[0,1].
\]
Here \(\rho_s\) is the within-plane query-alignment strength from
Assumption~\ref{assump:rope_conditional_covariance}.

To obtain a conservative numerical check, we apply the arithmetic-geometric
mean inequality to the denominator.
Since
\[
c_s+d_s
=
2a_sb_s,
\]
we define the surrogate
\[
\underline{\kappa}(\mathbf{w},\boldsymbol{\rho})
=
\frac{
2\sum_{s=1}^m w_s(a_s+\rho_s b_s)
}{
2+\sum_{s=1}^m w_s\rho_s c_s+\sum_{s=1}^m w_s\rho_s d_s
}
=
\frac{
\sum_{s=1}^m w_s(a_s+\rho_s b_s)
}{
\sum_{s=1}^m w_s(1+\rho_s a_sb_s)
}.
\]
When its numerator is nonnegative, this surrogate is a lower bound on the exact
conditional correlation, because the exact denominator is no larger than the
arithmetic-geometric mean denominator.
For a negative numerator, the inequality reverses, so the surrogate cannot be averaged as an unconditional lower bound.
 
For a candidate threshold \(\ell\in(-1,1)\), the condition
\[
\underline{\kappa}(\mathbf{w},\boldsymbol{\rho})\ge \ell
\]
is equivalent to
\[
\sum_{s=1}^m
w_s
\left[
a_s+\rho_s b_s-\ell(1+\rho_s a_sb_s)
\right]
\ge 0.
\]
For fixed \(\ell\), the inner term is linear in \(\rho_s\):
\[
a_s-\ell+\rho_s b_s(1-\ell a_s).
\]
Since \(1-\ell a_s>0\) for \(\ell\in(-1,1)\) and \(a_s\in[-1,1]\), the
worst-case choice of \(\rho_s\) is
\[
\rho_s^\star
=
\begin{cases}
1, & b_s<0,\\
0, & b_s\ge 0.
\end{cases}
\]
Since \(\underline{\kappa}\in[-1,1]\), this endpoint choice attains its minimum over the admissible \(\rho_s\)'s, including when the minimum is negative.
To obtain a pointwise-valid bound, define
\[
r_*(\mathbf{w})=\inf_{\boldsymbol{\rho}\in[0,1]^m}\underline{\kappa}(\mathbf{w},\boldsymbol{\rho}),
\qquad
h(u)=
\begin{cases}
u, & u\ge 0,\\
-1, & u<0.
\end{cases}
\]
If \(r_*\ge0\), the surrogate bounds the correlation for every admissible covariance; otherwise, the universal correlation bound \(-1\) applies.
Consequently,
\[
\operatorname{Corr}(S_{\mathrm{fwd}},S_{\mathrm{rev}}\mid\mathbf{q})
\ge h(r_*(\mathbf{w})).
\]

We estimate the resulting expected lower bound
\[
\mathbb{E}_{\mathbf{w}}[h(r_*(\mathbf{w}))],
\qquad
\mathbf{w}\sim\operatorname{Dirichlet}(1,\ldots,1),
\]
on the integer grid \(0\le\Delta_1,\Delta_2\le30\).
The inner linear-fractional minimization~\citep{dinkelbach1967nonlinear} uses the endpoint choice above.
For each \(D\in\{64,128,256\}\), we draw \(50{,}000\) independent samples using NumPy's PCG64 generator with seed \(20261011+D\), reusing those samples across the distance grid.
Tab.~\ref{tab:asymmetric_rope_check} reports the minimum sample mean over the grid.
Since \(h(r_*)\in[-1,1]\), Hoeffding's inequality and a union bound over the \(3\cdot31^2\) dimension--distance configurations give a simultaneous \(99\%\) lower confidence bound by subtracting
\[
\varepsilon=\sqrt{\frac{2\log(3\cdot31^2/0.01)}{50{,}000}}\approx0.022425
\]
from each sample mean.
The bounds remain positive throughout the tested grid, supporting extension beyond equal distances.

\begin{table}[H]
\centering
\small
\caption{Minimum estimated expected correlation lower bound over integer \(0\le\Delta_1,\Delta_2\le30\), with a sign-corrected surrogate. The simultaneous lower confidence bounds are rounded downward.}
\vspace{0.5em}
\label{tab:asymmetric_rope_check}
\begin{tabular}{lccc}
\toprule
\(D\) & 64 & 128 & 256 \\
\midrule
Estimated minimum & 0.220 & 0.363 & 0.397 \\
\(99\%\) simultaneous lower bound & 0.197 & 0.340 & 0.374 \\
\bottomrule
\end{tabular}
\end{table}

\subsection{A Relaxed Fixed-Query Covariance Result}
\label{app:relaxed_fixed_query_covariance}

Theorem~\ref{thm:rope_attention_correlation} uses Assumption~\ref{assump:rope_conditional_covariance} to obtain a representation-independent lower bound.
For a fixed learned query, most of its statistical assumptions are unnecessary.

\paragraph{Assumption.}
Fix an arbitrary masked-token query \(\mathbf{q}=\mathbf{q}_{\M}\), and write \(\mathbf{q}_s,\mathbf{k}_s\in\mathbb{R}^2\) for the query and evidence-key vectors in RoPE plane~\(s\).
We assume only finite conditional second moments and
\[
\operatorname{Cov}(\mathbf{k}_B|\mathbf{q})
=
\operatorname{diag}(\Sigma_1,\ldots,\Sigma_{D/2}),
\quad
\Sigma_s\succeq 0.
\]
For \(\mathbf{q}_s\neq\mathbf{0}\) and \(\tau_s^2>0\), let \(\mathbf{q}_s^\perp\) denote its \(90^\circ\) rotation and define
\[
\tau_s^2=\operatorname{Tr}(\Sigma_s),
\quad
\rho_s
=
\frac{
\mathbf{q}_s^\top\Sigma_s\mathbf{q}_s
-
(\mathbf{q}_s^\perp)^\top\Sigma_s\mathbf{q}_s^\perp
}{
\tau_s^2\|\mathbf{q}_s\|^2
}
\in[-1,1].
\]
A plane with \(\mathbf{q}_s=\mathbf{0}\) or \(\tau_s^2=0\) contributes zero.
Compared with Assumption~\ref{assump:rope_conditional_covariance}, the query need not be random or Gaussian, the plane scales \(\tau_s^2\) may differ, \(\Sigma_s\) need not be aligned with \(\mathbf{q}_s\), and \(\rho_s\) may have either sign.
The only structural restriction retained is zero cross-plane conditional covariance; independence across planes is not required.

\begin{proposition}[Fixed-Query RoPE Score Covariance]
\label{prop:fixed_query_rope_covariance}
For \(\Delta_1=\Delta_2=\Delta\),
\[
\operatorname{Cov}(S_{\mathrm{fwd}},S_{\mathrm{rev}}|\mathbf{q})
=
\frac{1}{2}\sum_s
\tau_s^2\|\mathbf{q}_s\|^2
\left[\cos(2\theta_s\Delta)+\rho_s\right].
\]
Hence the covariance is positive if and only if
\[
\sum_s
\tau_s^2\|\mathbf{q}_s\|^2
\left[\cos(2\theta_s\Delta)+\rho_s\right]>0.
\]
Whenever both score variances are nonzero, the conditional correlation has the same sign.
\end{proposition}

\begin{proof}
Condition on \(\mathbf{q}\) and center \(\mathbf{k}_B\).
Block diagonality gives
\[
\operatorname{Cov}(S_{\mathrm{fwd}},S_{\mathrm{rev}}|\mathbf{q})
=
\sum_s\mathbf{q}_s^\top R_s(\Delta)\Sigma_s R_s(-\Delta)^\top\mathbf{q}_s
=
\sum_s\mathbf{q}_s^\top R_s(\Delta)\Sigma_s R_s(\Delta)\mathbf{q}_s.
\]
For each plane, a direct two-dimensional expansion in the basis
\(\{\mathbf{q}_s/\|\mathbf{q}_s\|,\mathbf{q}_s^\perp/\|\mathbf{q}_s\|\}\) gives
\[
\mathbf{q}_s^\top R_s(\Delta)\Sigma_s R_s(\Delta)\mathbf{q}_s
=
\frac{1}{2}\tau_s^2\|\mathbf{q}_s\|^2
\left[\cos(2\theta_s\Delta)+\rho_s\right].
\]
Summing over \(s\) proves the result.
\end{proof}

The sign is therefore determined by a learned weighted aggregate of the RoPE cosine terms and signed anisotropies; the anisotropies need not be nonnegative plane by plane.
The original theorem imposes stronger symmetry conditions to obtain a parameter-free lower bound.
The remaining block-diagonal assumption excludes cross-plane covariance terms, whose signs are unrestricted in general.

\paragraph{Cross-plane covariance.}
The block-diagonal restriction can be removed by retaining its omitted contribution explicitly.
For a fixed query~\(\mathbf{q}=\mathbf{q}_{\M}\), let
\(\Sigma_{s,t}=\operatorname{Cov}(\mathbf{k}_B^{(s)},\mathbf{k}_B^{(t)}\mid\mathbf{q})\),
so that \(\Sigma_{s,s}=\Sigma_s\).
Write \(R_s(\Delta)\) for the rotation in plane~\(s\), and define
\[
\mathbf{a}_s=R_s(\Delta)^\top\mathbf{q}_s,
\quad
\mathbf{b}_s=R_s(-\Delta)^\top\mathbf{q}_s,
\quad
\mathcal{R}_{\mathrm{cross}}(\Delta)
=\sum_{s\ne t}\mathbf{a}_s^\top\Sigma_{s,t}\mathbf{b}_t.
\]
Expanding the conditional covariance over all pairs of planes gives the exact identity
\[
\operatorname{Cov}(S_{\mathrm{fwd}},S_{\mathrm{rev}}\mid\mathbf{q})
=\frac{1}{2}\sum_s\tau_s^2\|\mathbf{q}_s\|^2
\bigl[\cos(2\theta_s\Delta)+\rho_s\bigr]
+\mathcal{R}_{\mathrm{cross}}(\Delta).
\]
Thus, the fixed-query formula in Proposition~\ref{prop:fixed_query_rope_covariance}
is the block-diagonal case \(\mathcal{R}_{\mathrm{cross}}(\Delta)=0\).
Without this restriction, the sign depends on the sum of the within-plane weighted contribution and the cross-plane residual.
In particular, a positive within-plane contribution larger than
\(\lvert\mathcal{R}_{\mathrm{cross}}(\Delta)\rvert\) is sufficient for positive covariance.
Whenever both score variances are nonzero, positive covariance is equivalent to positive conditional correlation.

\paragraph{Empirical diagnostic.}
We estimate \(\rho_s\) in the \(128\) RoPE planes of nine trained one-layer RADD checkpoints across three sequence lengths and three random seeds.
Tab.~\ref{tab:rope_anisotropy} shows mixed signs and small median anisotropy.
In particular, the strong condition \(\rho_s\ge 0\) does not hold plane by plane.
This motivates Proposition~\ref{prop:fixed_query_rope_covariance}, whose positivity condition depends on the learned weighted aggregate rather than a per-plane sign constraint.

\begin{table}[H]
\centering
\caption{Per-plane RoPE anisotropy in nine trained one-layer RADD checkpoints.}
\vspace{0.5em}
\label{tab:rope_anisotropy}
\begin{tabular}{ccc}
\toprule
\(L\) & \(\rho_s\ge 0\) & Median \(\rho_s\) \\
\midrule
10 & 56.8\% & 0.0448 \\
20 & 55.5\% & 0.0258 \\
30 & 57.0\% & 0.0333 \\
\bottomrule
\end{tabular}
\end{table}

The sign proportions and medians in Tab.~\ref{tab:rope_anisotropy} do not determine the theorem-relevant weighted covariance:
the latter also depends on \(\tau_s^2\|\mathbf{q}_s\|^2\), the relative distance, and the cross-plane contribution.
These measurements therefore motivate the relaxed condition but do not validate it, or every component of Assumption~\ref{assump:rope_conditional_covariance}, in trained models.
The observed score correlations in Secs.~\ref{sec:one_layer_validation} and \ref{sec:large_scale_validation} provide evidence for the predicted qualitative outcome, rather than direct validation of the covariance assumptions.

\section{Additional Theoretical Analyses on Training Dynamics}
\label{app:additional_theory}
This section collects additional analyses that complement the main training-dynamics results.
App.~\ref{app:retrieval_margin} characterizes the exact reverse target-versus-competitor margin with competing context.
Apps.~\ref{app:full_grad_update} and \ref{app:original_parameter_alignment} decompose gradient contributions in effective coordinates and in a reduced model with factorized parameters, respectively.
App.~\ref{app:probability_transfer} establishes an idealized MDM probability-convergence result, followed by a one-layer ARM comparison in App.~\ref{app:arm_comparison}.

\subsection{Exact Retrieval Margin with Competing Context}
\label{app:retrieval_margin}

The routing theorem concerns the raw attention score on the evidence token, whereas successful retrieval depends on its softmax weight and the value-side evidence from all context tokens.
The following identity makes this distinction explicit in the one-layer model.

\begin{proposition}[Reverse Target-versus-Competitor Margin]
\label{prop:reverse_retrieval_margin}
Let \(i\) be the masked position in a reverse prompt, let \(x^j=B\) be the designated evidence token, let \(A\) be the target token, and let \(C\neq A\) be any competitor.
Then
\[
(\mathbf{z}_i^{\mathrm{rev}})_A-(\mathbf{z}_i^{\mathrm{rev}})_C
=
\alpha_{i,j}^{\mathrm{rev}}(Y_{A,B}-Y_{C,B})
+
\sum_{k\neq j}\alpha_{i,k}^{\mathrm{rev}}(Y_{A,x^k}-Y_{C,x^k}).
\]
The first term captures the contribution of token~\(B\), while the second accounts for the remaining context.
The target~\(A\) is the strict top-1 prediction if and only if this margin is positive for every \(C\neq A\).
\end{proposition}

\begin{proof}
The one-layer logit representation gives
\[
(\mathbf{z}_i^{\mathrm{rev}})_A
=
\sum_k\alpha_{i,k}^{\mathrm{rev}}Y_{A,x^k},
\quad
(\mathbf{z}_i^{\mathrm{rev}})_C
=
\sum_k\alpha_{i,k}^{\mathrm{rev}}Y_{C,x^k}.
\]
Subtracting and separating the term \(k=j\) gives the stated identity.
Strict top-1 prediction is equivalent to the target logit exceeding every competitor logit.
\end{proof}

The decomposition exposes two interference routes.
Softmax competition determines \(\alpha_{i,j}^{\mathrm{rev}}\), while the remaining context contributes competing value-side evidence through the second term.
Proposition~\ref{prop:shared_y_update} strengthens the \(B\)-to-\(A\) storage margin, and Sec.~\ref{sec:attention_correlation} couples the raw evidence-token scores, but neither alone guarantees retrieval: the total margin above must be positive for every competitor.
In particular, even a large \(\alpha_{i,j}^{\mathrm{rev}}\) does not ensure retrieval if the remaining-context term is sufficiently negative.

\paragraph{Reverse attention-weight diagnostic.}
To examine softmax competition directly, we pool reverse-query placements from the eight successful one-layer checkpoints used in Sec.~\ref{sec:one_layer_validation}, excluding one \(L=20\) run with zero forward accuracy.
Tab.~\ref{tab:reverse_attention_weights} reports substantial reverse attention on the evidence token.
At every reported length, \(B\) receives the highest attention weight among all context positions in \(100\%\) of the evaluated reverse queries.

\begin{table}[H]
\centering
\caption{Reverse attention weight on the evidence token in successful one-layer RADD checkpoints.}
\vspace{0.5em}
\label{tab:reverse_attention_weights}
\begin{tabular}{ccc}
\toprule
\(L\) & Median \(\alpha_{i,j}^{\mathrm{rev}}\) on \(B\) & Uniform \(1/L\) \\
\midrule
10 & 0.589 & 0.100 \\
20 & 0.833 & 0.050 \\
30 & 0.567 & 0.033 \\
\bottomrule
\end{tabular}
\end{table}

\subsection{Full-Gradient Alignment Analysis}
\label{app:full_grad_update}

Thm.~\ref{thm:b_column_gradient_alignment} proves a sign-determined alignment
for the shared storage column \(Y_{:,B}\): a forward masked update through the
common evidence token \(B\) decreases the reverse loss to first order.
To understand why the full-gradient cosine measured in
Secs.~\ref{sec:one_layer_validation} and \ref{sec:large_scale_validation}
remains positive, we revisit the storage-side \(Y\) gradient and then analyze
the attention-routing parameters
\[
Q=W_QW_E,
\quad
K=W_KW_E.
\]
This is the first point where the effective query-key coordinates \(Q\) and
\(K\) are used in the gradient analysis.

We use cosine similarity because it separates direction from scale and yields more interpretable decompositions.
The \(Y\)-gradient cosine factors into a probability-error cosine and an attention-mass cosine.
For the query-key terms, the corresponding cosine decomposes into a loss-sensitivity sign and a geometric score-gradient cosine.
The analysis separates the guaranteed storage contribution from routing terms whose signs depend on score sensitivities and geometry.

\paragraph{Storage-side cosine decomposition.}
For a prompt with target token \(A\), prediction distribution \(\mathbf{p}\),
and attention weights \(\alpha_k\), define the token-type attention vector
\[
\mathbf{a}
=
\sum_{k=1}^L
\alpha_k \mathbf{e}_{x^k}
\in\mathbb{R}^{|\mathcal{V}|+1}.
\]
The \(x\)-coordinate of \(\mathbf{a}\) is the total attention mass assigned to
all occurrences of token \(x\).
Since
\[
\mathbf{z}
=
\sum_{k=1}^L \alpha_k Y\mathbf{e}_{x^k}
=
Y\mathbf{a},
\]
the effective storage gradient is the outer product
\[
\nabla_Y\mathcal{L}
=
(\mathbf{p}-\mathbf{e}_A)\mathbf{a}^{\top}.
\]
Therefore,
\[
\left\langle
\nabla_Y\mathcal{L}_{\mathrm{fwd}},
\nabla_Y\mathcal{L}_{\mathrm{rev}}
\right\rangle_F
=
\left\langle
\mathbf{p}_{\mathrm{fwd}}-\mathbf{e}_A,
\mathbf{p}_{\mathrm{rev}}-\mathbf{e}_A
\right\rangle
\left\langle
\mathbf{a}_{\mathrm{fwd}},
\mathbf{a}_{\mathrm{rev}}
\right\rangle.
\]
Since the Frobenius norm of an outer product factorizes,
\[
\|\nabla_Y\mathcal{L}\|_F
=
\|\mathbf{p}-\mathbf{e}_A\|\,
\|\mathbf{a}\|,
\]
we obtain the exact cosine decomposition
\[
\cos_F
\left(
\nabla_Y\mathcal{L}_{\mathrm{fwd}},
\nabla_Y\mathcal{L}_{\mathrm{rev}}
\right)
=
\cos
\left(
\mathbf{p}_{\mathrm{fwd}}-\mathbf{e}_A,
\mathbf{p}_{\mathrm{rev}}-\mathbf{e}_A
\right)
\cos
\left(
\mathbf{a}_{\mathrm{fwd}},
\mathbf{a}_{\mathrm{rev}}
\right).
\]
The probability-error cosine is positive because
\[
\left\langle
\mathbf{p}_{\mathrm{fwd}}-\mathbf{e}_A,
\mathbf{p}_{\mathrm{rev}}-\mathbf{e}_A
\right\rangle
=
(1-p_{\mathrm{fwd}}(A))(1-p_{\mathrm{rev}}(A))
+
\sum_{C\neq A}
p_{\mathrm{fwd}}(C)p_{\mathrm{rev}}(C)
>0.
\]
The attention-mass cosine is nonnegative because
\(\mathbf{a}_{\mathrm{fwd}}\) and \(\mathbf{a}_{\mathrm{rev}}\) have
nonnegative entries, and it is strictly positive whenever the two prompts share
at least one token receiving positive attention in both prompts.
In particular, the shared evidence token \(B\) makes this cosine positive.
Thus the full \(Y\)-gradient cosine is positive, strengthening the
\(B\)-column inner-product result of Thm.~\ref{thm:b_column_gradient_alignment}.

\paragraph{Query-key cosine decomposition.}
We next isolate the component of the \(Q\) and \(K\) gradients that flows
through the attention score assigned to the evidence token \(B\).
For a single masked position, write
\[
\mathbf{z}
=
\sum_{\ell=1}^L\alpha_\ell Y\mathbf{e}_{x^\ell},
\quad
\mathbf{p}=\operatorname{softmax}(\mathbf{z}),
\quad
\mathcal{L}=-\log p(A).
\]
Define the token-wise storage error
\[
\mu_\ell
=
(\mathbf{p}-\mathbf{e}_A)^\top
Y\mathbf{e}_{x^\ell}.
\]
Since \(\partial \mathcal{L}/\partial \alpha_\ell=\mu_\ell\) and
\(\alpha_\ell=\operatorname{softmax}(S_\ell/\sqrt{D})\), the loss sensitivity
of score \(S_k\) is
\[
\frac{\partial \mathcal{L}}{\partial S_k}
=
\frac{\alpha_k}{\sqrt{D}}
\left(
\mu_k-\sum_{\ell=1}^L\alpha_\ell\mu_\ell
\right).
\]

To isolate the clean transfer term, we restrict the query-key calculation to the backpropagation path through the attention score assigned to the evidence token \(B\).
Thus the derivation below keeps the \(B\)-associated score-gradient contribution and leaves the other token contributions to the final interpretation paragraph.

In this routing calculation, \(S_B\) and \(\alpha_B\) refer to one designated occurrence of token~\(B\), rather than the total attention mass over repeated occurrences.
Define
\[
\xi_B
:=
\frac{\partial \mathcal{L}}{\partial S_B}
=
\frac{\alpha_B}{\sqrt{D}}
\left(
\mu_B-\sum_{\ell=1}^L\alpha_\ell\mu_\ell
\right).
\]
The sign of \(\xi_B\) is determined by
\[
\mu_B-\sum_{\ell=1}^L\alpha_\ell\mu_\ell.
\]
This contrast measures whether attending more to \(B\) would reduce the loss compared with the current attention mixture.
For relative positional encodings, the value carried by token \(B\) does not depend on its position, so the forward and reverse prompts retrieve the same \(B\)-side evidence even though they route attention through different attention scores.
If increasing attention to \(B\) reduces the loss in both prompts, then both score sensitivities are negative and hence
\[
\operatorname{sgn}(\xi_B^{\mathrm{fwd}}\xi_B^{\mathrm{rev}})=1.
\]

For a score \(S_B=S(\mathbf{q}_{\M},\mathbf{k}_B,\Delta)\), the \(B\)-route
gradient components are
\[
\nabla_Q\mathcal{L}^{(B)}
=
\xi_B
(\nabla_{\mathbf{q}_{\M}}S_B)
\mathbf{e}_{\M}^{\top},
\quad
\nabla_K\mathcal{L}^{(B)}
=
\xi_B
(\nabla_{\mathbf{k}_B}S_B)
\mathbf{e}_{B}^{\top}.
\]
Hence, whenever the relevant norms are nonzero,
\[
\cos_F
\left(
\nabla_Q\mathcal{L}_{\mathrm{fwd}}^{(B)},
\nabla_Q\mathcal{L}_{\mathrm{rev}}^{(B)}
\right)
=
\operatorname{sgn}
\left(
\xi_B^{\mathrm{fwd}}\xi_B^{\mathrm{rev}}
\right)
\cos
\left(
\nabla_{\mathbf{q}}S_B^{\mathrm{fwd}},
\nabla_{\mathbf{q}}S_B^{\mathrm{rev}}
\right),
\]
and
\[
\cos_F
\left(
\nabla_K\mathcal{L}_{\mathrm{fwd}}^{(B)},
\nabla_K\mathcal{L}_{\mathrm{rev}}^{(B)}
\right)
=
\operatorname{sgn}
\left(
\xi_B^{\mathrm{fwd}}\xi_B^{\mathrm{rev}}
\right)
\cos
\left(
\nabla_{\mathbf{k}}S_B^{\mathrm{fwd}},
\nabla_{\mathbf{k}}S_B^{\mathrm{rev}}
\right).
\]
Thus, once the sensitivity signs agree, the \(B\)-route query-key cosine is
controlled by the geometry of the attention-score gradients.

\paragraph{RoPE.}
For RoPE,
\[
S_B^{\mathrm{fwd}}
=
\mathbf{q}_{\M}^{\top}R(\Delta_1)\mathbf{k}_B,
\quad
S_B^{\mathrm{rev}}
=
\mathbf{q}_{\M}^{\top}R(-\Delta_2)\mathbf{k}_B.
\]
Therefore,
\[
\nabla_{\mathbf{q}}S_B^{\mathrm{fwd}}
=
R(\Delta_1)\mathbf{k}_B,
\quad
\nabla_{\mathbf{q}}S_B^{\mathrm{rev}}
=
R(-\Delta_2)\mathbf{k}_B,
\]
and
\[
\nabla_{\mathbf{k}}S_B^{\mathrm{fwd}}
=
R(-\Delta_1)\mathbf{q}_{\M},
\quad
\nabla_{\mathbf{k}}S_B^{\mathrm{rev}}
=
R(\Delta_2)\mathbf{q}_{\M}.
\]
Thus the geometric cosines are
\[
\cos
\left(
\nabla_{\mathbf{q}}S_B^{\mathrm{fwd}},
\nabla_{\mathbf{q}}S_B^{\mathrm{rev}}
\right)
=
\frac{
\mathbf{k}_B^\top R(\Delta_1+\Delta_2)\mathbf{k}_B
}{
\|\mathbf{k}_B\|^2
},
\]
and
\[
\cos
\left(
\nabla_{\mathbf{k}}S_B^{\mathrm{fwd}},
\nabla_{\mathbf{k}}S_B^{\mathrm{rev}}
\right)
=
\frac{
\mathbf{q}_{\M}^\top R(\Delta_1+\Delta_2)\mathbf{q}_{\M}
}{
\|\mathbf{q}_{\M}\|^2
}.
\]
These quantities are not pointwise positive for every vector, because RoPE
rotates different frequency planes by different angles.
For isotropic Gaussian reference distributions, their expectations equal normalized trace terms.
For example, if \(\mathbf{k}_B\sim\mathcal{N}(0,\sigma^2 I_D)\), then
\[
\mathbb{E}
\left[
\frac{
\mathbf{k}_B^\top R(\Delta_1+\Delta_2)\mathbf{k}_B
}{
\|\mathbf{k}_B\|^2
}
\right]
=
\frac{1}{D}\operatorname{Tr}(R(\Delta_1+\Delta_2)).
\]
Thus, in this isotropic reference model, the geometric query-key cosine has positive expectation in trace-positive distance regimes.
Its sign for a particular learned representation is not fixed by the trace alone.

\paragraph{ALiBi.}
For ALiBi,
\[
S_B
=
\mathbf{q}_{\M}^{\top}\mathbf{k}_B
+
b\Delta,
\]
The bias \(b\Delta\) affects the attention score and hence the routing weight, but it
vanishes when differentiating with respect to \(\mathbf{q}_{\M}\) or
\(\mathbf{k}_B\):
\[
\nabla_{\mathbf{q}}S_B=\mathbf{k}_B,
\quad
\nabla_{\mathbf{k}}S_B=\mathbf{q}_{\M}.
\]
Consequently,
\[
\cos
\left(
\nabla_{\mathbf{q}}S_B^{\mathrm{fwd}},
\nabla_{\mathbf{q}}S_B^{\mathrm{rev}}
\right)
=
1,
\quad
\cos
\left(
\nabla_{\mathbf{k}}S_B^{\mathrm{fwd}},
\nabla_{\mathbf{k}}S_B^{\mathrm{rev}}
\right)
=
1.
\]
Thus, when the sensitivity signs agree, ALiBi gives exact \(B\)-route
query-key cosine alignment.
This sharply contrasts with RoPE: ALiBi changes the scalar attention scores by
an additive distance bias, but leaves the query-key gradient directions
identical across forward and reverse configurations.

\paragraph{Full-gradient interpretation.}
The \(Y\)-gradient cosine gives the clean storage-side signal, while the
\(B\)-route \(Q,K\) analysis shows that the attention-routing terms do not
inherently oppose it.
The remaining \(Q,K\) terms involve other context tokens and cross-token interactions through the attention softmax, so their signs are not determined by the \(B\)-route calculation.
Their aggregate effect is assessed empirically through the trainable-parameter gradient diagnostics in Secs.~\ref{sec:one_layer_validation} and \ref{sec:large_scale_validation}.

This analysis complements Thm.~\ref{thm:b_column_gradient_alignment} by separating storage and routing contributions.
Under the common-sign condition, ALiBi gives exact \(B\)-route cosine alignment, while RoPE's geometric factors admit the reference-distribution analysis above.
The full trainable-parameter gradient cosine is measured in Secs.~\ref{sec:one_layer_validation} and \ref{sec:large_scale_validation}.

\subsection{Gradient Alignment in the Factorized Parameterization}
\label{app:original_parameter_alignment}

The main text analyzes the transfer mechanism through the shared storage
parameter \(Y\), and the preceding appendix section uses the effective
query-key parameters \(Q\) and \(K\) to analyze routing.
We now relate this effective-parameter picture to the original
parameterization, where \(W_E,W_Q,W_K,W_V\), and \(W_U\) are updated jointly.
We derive the gradient contributions in these factor matrices for an explicit reduced routing model.

\paragraph{Setup.}
To isolate the dynamics, we consider a single evidence token setting with arbitrary relative distances.
Even though there is only one informative token, we keep the relative distances explicit so that the effect of positional encoding remains visible in the attention score.
The forward prompt contains a masked target position and the evidence token \(B\), with \(B\) lying \(\Delta_1\) positions to the right of the mask.
The reverse prompt contains the same evidence token \(B\), with \(B\) lying \(\Delta_2\) positions to the left of the mask.
Both prompts ask the model to recover the same target token \(A\).

Let
\[
\mathbf{h}_{\M}=W_E\mathbf{e}_{\M},
\quad
\mathbf{h}_B=W_E\mathbf{e}_B,
\]
and
\[
\mathbf{q}_{\M}=W_Q\mathbf{h}_{\M},
\quad
\mathbf{k}_B=W_K\mathbf{h}_B,
\quad
\mathbf{v}_B=W_V\mathbf{h}_B.
\]
Let
\[
S_B=S(\mathbf{q}_{\M},\mathbf{k}_B,\Delta)
\]
denote the attention score assigned to the evidence token \(B\).
We use an auxiliary two-route model, in which the evidence score competes with a fixed reference score of zero and the reference route has a fixed zero value.
The attention weight on \(B\) is then
\[
\alpha_B=\sigma(S_B/\sqrt{D}),
\]
where \(\sigma\) is the sigmoid function.
The reference route is independent of all trainable parameters; it is not the learned mask-token route of the full Transformer.

The logits of this reduced model retain the evidence-token contribution:
\[
\mathbf{z}=W_U(\alpha_B\mathbf{v}_B),
\quad
\mathbf{p}=\operatorname{softmax}(\mathbf{z}),
\quad
\mathbf{u}=W_U^\top(\mathbf{p}-\mathbf{e}_A).
\]

\begin{proposition}[Original-Parameter Gradient Decomposition]
\label{prop:original_param_alignment}
Under the reduced single-evidence-token setup above, define
\[
I_W
=
\left\langle
\nabla_W\mathcal{L}_{\mathrm{fwd}},
\nabla_W\mathcal{L}_{\mathrm{rev}}
\right\rangle_F
\]
for each original parameter matrix \(W\in\{W_U,W_V,W_Q,W_K,W_E\}\).
Then the unembedding, value, query, and key projection contributions satisfy
\[
I_{W_U}
=
\alpha_B^{\mathrm{fwd}}\alpha_B^{\mathrm{rev}}
\left\langle
\mathbf{p}_{\mathrm{fwd}}-\mathbf{e}_A,
\mathbf{p}_{\mathrm{rev}}-\mathbf{e}_A
\right\rangle
\|\mathbf{v}_B\|^2,
\]
\[
I_{W_V}
=
\alpha_B^{\mathrm{fwd}}\alpha_B^{\mathrm{rev}}
\left\langle
\mathbf{u}_{\mathrm{fwd}},
\mathbf{u}_{\mathrm{rev}}
\right\rangle
\|\mathbf{h}_B\|^2,
\]
\[
I_{W_Q}
=
\frac{
\alpha_B^{\mathrm{fwd}}(1-\alpha_B^{\mathrm{fwd}})
\alpha_B^{\mathrm{rev}}(1-\alpha_B^{\mathrm{rev}})
}{D}
(\mathbf{u}_{\mathrm{fwd}}^\top\mathbf{v}_B)
(\mathbf{u}_{\mathrm{rev}}^\top\mathbf{v}_B)
\|\mathbf{h}_{\M}\|^2
\left\langle
\nabla_{\mathbf{q}}S_B^{\mathrm{fwd}},
\nabla_{\mathbf{q}}S_B^{\mathrm{rev}}
\right\rangle,
\]
and
\[
I_{W_K}
=
\frac{
\alpha_B^{\mathrm{fwd}}(1-\alpha_B^{\mathrm{fwd}})
\alpha_B^{\mathrm{rev}}(1-\alpha_B^{\mathrm{rev}})
}{D}
(\mathbf{u}_{\mathrm{fwd}}^\top\mathbf{v}_B)
(\mathbf{u}_{\mathrm{rev}}^\top\mathbf{v}_B)
\|\mathbf{h}_B\|^2
\left\langle
\nabla_{\mathbf{k}}S_B^{\mathrm{fwd}},
\nabla_{\mathbf{k}}S_B^{\mathrm{rev}}
\right\rangle.
\]
For the embedding matrix, the inner product decomposes as
\[
I_{W_E}
=
I_{W_E}^{(Q)}
+
I_{W_E}^{(K)}
+
I_{W_E}^{(V)}
+
I_{W_E}^{(KV)},
\]
where
\[
I_{W_E}^{(Q)}
=
\frac{
\alpha_B^{\mathrm{fwd}}(1-\alpha_B^{\mathrm{fwd}})
\alpha_B^{\mathrm{rev}}(1-\alpha_B^{\mathrm{rev}})
}{D}
(\mathbf{u}_{\mathrm{fwd}}^\top\mathbf{v}_B)
(\mathbf{u}_{\mathrm{rev}}^\top\mathbf{v}_B)
\left\langle
W_Q^\top\nabla_{\mathbf{q}}S_B^{\mathrm{fwd}},
W_Q^\top\nabla_{\mathbf{q}}S_B^{\mathrm{rev}}
\right\rangle,
\]
\[
I_{W_E}^{(K)}
=
\frac{
\alpha_B^{\mathrm{fwd}}(1-\alpha_B^{\mathrm{fwd}})
\alpha_B^{\mathrm{rev}}(1-\alpha_B^{\mathrm{rev}})
}{D}
(\mathbf{u}_{\mathrm{fwd}}^\top\mathbf{v}_B)
(\mathbf{u}_{\mathrm{rev}}^\top\mathbf{v}_B)
\left\langle
W_K^\top\nabla_{\mathbf{k}}S_B^{\mathrm{fwd}},
W_K^\top\nabla_{\mathbf{k}}S_B^{\mathrm{rev}}
\right\rangle,
\]
\[
I_{W_E}^{(V)}
=
\alpha_B^{\mathrm{fwd}}\alpha_B^{\mathrm{rev}}
\left\langle
W_V^\top\mathbf{u}_{\mathrm{fwd}},
W_V^\top\mathbf{u}_{\mathrm{rev}}
\right\rangle,
\]
and \(I_{W_E}^{(KV)}\) collects the cross-terms between the key and value paths
inside the \(B\)-embedding column.
\end{proposition}

\paragraph{Proof of Proposition~\ref{prop:original_param_alignment}.}
For \(\mathcal{L}=-\log p(A)\),
\[
\frac{\partial \mathcal{L}}{\partial \mathbf{z}}
=
\mathbf{p}-\mathbf{e}_A.
\]
Since \(\mathbf{z}=W_U(\alpha_B\mathbf{v}_B)\), the unembedding and value
projection gradients are
\[
\nabla_{W_U}\mathcal{L}
=
\alpha_B(\mathbf{p}-\mathbf{e}_A)\mathbf{v}_B^\top,
\]
and
\[
\nabla_{W_V}\mathcal{L}
=
\alpha_B W_U^\top(\mathbf{p}-\mathbf{e}_A)\mathbf{h}_B^\top
=
\alpha_B\mathbf{u}\mathbf{h}_B^\top.
\]
Taking Frobenius inner products of the corresponding forward and reverse
gradients gives the stated formulas for \(I_{W_U}\) and \(I_{W_V}\).

For the query and key projections,
\[
\frac{\partial \alpha_B}{\partial S_B}
=
\frac{\alpha_B(1-\alpha_B)}{\sqrt{D}}.
\]
Also,
\[
\frac{\partial \mathcal{L}}{\partial \alpha_B}
=
(\mathbf{p}-\mathbf{e}_A)^\top W_U\mathbf{v}_B
=
\mathbf{u}^\top\mathbf{v}_B.
\]
Hence
\[
\frac{\partial \mathcal{L}}{\partial S_B}
=
\frac{\alpha_B(1-\alpha_B)}{\sqrt{D}}
\mathbf{u}^\top\mathbf{v}_B.
\]
By the chain rule,
\[
\nabla_{W_Q}\mathcal{L}
=
\left[
\frac{\alpha_B(1-\alpha_B)}{\sqrt{D}}
\mathbf{u}^\top\mathbf{v}_B
\right]
(\nabla_{\mathbf{q}_{\M}}S_B)\mathbf{h}_{\M}^\top,
\]
and
\[
\nabla_{W_K}\mathcal{L}
=
\left[
\frac{\alpha_B(1-\alpha_B)}{\sqrt{D}}
\mathbf{u}^\top\mathbf{v}_B
\right]
(\nabla_{\mathbf{k}_B}S_B)\mathbf{h}_B^\top.
\]
Their Frobenius inner products yield the stated formulas for \(I_{W_Q}\) and
\(I_{W_K}\).

It remains to track the embedding matrix.
The masked-token embedding receives the query-path signal
\[
\nabla_{\mathbf{h}_{\M}}\mathcal{L}
=
\left[
\frac{\alpha_B(1-\alpha_B)}{\sqrt{D}}
\mathbf{u}^\top\mathbf{v}_B
\right]
W_Q^\top\nabla_{\mathbf{q}_{\M}}S_B.
\]
The evidence-token embedding receives both key-path and value-path signals:
\[
\nabla_{\mathbf{h}_B}\mathcal{L}
=
\left[
\frac{\alpha_B(1-\alpha_B)}{\sqrt{D}}
\mathbf{u}^\top\mathbf{v}_B
\right]
W_K^\top\nabla_{\mathbf{k}_B}S_B
+
\alpha_B W_V^\top\mathbf{u}.
\]
Thus
\[
\nabla_{W_E}\mathcal{L}
=
(\nabla_{\mathbf{h}_{\M}}\mathcal{L})\mathbf{e}_{\M}^\top
+
(\nabla_{\mathbf{h}_B}\mathcal{L})\mathbf{e}_B^\top.
\]
The masked-token column and the \(B\)-token column are orthogonal because
\(\mathbf{e}_{\M}^\top\mathbf{e}_B=0\), so the query-path contribution
separates from the \(B\)-column contribution.
Inside the \(B\)-column, the key and value paths contribute the \(K\)-path and
\(V\)-path terms above, along with their cross-terms \(I_{W_E}^{(KV)}\).
This gives the stated decomposition.
\qed

\paragraph{Storage-side interpretation.}
The decomposition above shows how the effective storage-side alignment appears
inside the original parameterization.
The \(W_U\) term is sign-determined:
\[
\left\langle
\mathbf{p}_{\mathrm{fwd}}-\mathbf{e}_A,
\mathbf{p}_{\mathrm{rev}}-\mathbf{e}_A
\right\rangle
=
(1-p_{\mathrm{fwd}}(A))(1-p_{\mathrm{rev}}(A))
+
\sum_{C\neq A}p_{\mathrm{fwd}}(C)p_{\mathrm{rev}}(C)
>0.
\]
Thus \(I_{W_U}>0\) whenever \(\mathbf{v}_B\neq\mathbf{0}\) and \(B\) receives positive attention in both prompts.

The \(W_V\) and \(W_E^{(V)}\) terms depend on the backpropagated error signals
\[
\left\langle
\mathbf{u}_{\mathrm{fwd}},
\mathbf{u}_{\mathrm{rev}}
\right\rangle
=
\left\langle
W_U^\top(\mathbf{p}_{\mathrm{fwd}}-\mathbf{e}_A),
W_U^\top(\mathbf{p}_{\mathrm{rev}}-\mathbf{e}_A)
\right\rangle.
\]
These terms contribute positively when the unembedding and value maps preserve the positive alignment of the two probability-error vectors.
Unlike the \(W_U\) term, their signs are not determined for arbitrary factor matrices.

\paragraph{Query-key routing interpretation.}
The \(W_Q,W_K\), and query/key paths of \(W_E\) share the scalar coefficient
\[
(\mathbf{u}_{\mathrm{fwd}}^\top\mathbf{v}_B)
(\mathbf{u}_{\mathrm{rev}}^\top\mathbf{v}_B).
\]
The scalar \(\mathbf{u}^\top\mathbf{v}_B\) measures whether increasing attention
to \(B\) lowers the loss.
If increasing attention to \(B\) reduces the loss in both prompts, the two scalar factors are negative, so their product is positive.
The remaining factor is geometric and depends on the positional encoding.

For RoPE,
\[
S_B^{\mathrm{fwd}}
=
\mathbf{q}_{\M}^{\top}R(\Delta_1)\mathbf{k}_B,
\quad
S_B^{\mathrm{rev}}
=
\mathbf{q}_{\M}^{\top}R(-\Delta_2)\mathbf{k}_B.
\]
Therefore,
\[
\left\langle
\nabla_{\mathbf{q}}S_B^{\mathrm{fwd}},
\nabla_{\mathbf{q}}S_B^{\mathrm{rev}}
\right\rangle
=
\mathbf{k}_B^\top R(\Delta_1+\Delta_2)\mathbf{k}_B,
\]
and
\[
\left\langle
\nabla_{\mathbf{k}}S_B^{\mathrm{fwd}},
\nabla_{\mathbf{k}}S_B^{\mathrm{rev}}
\right\rangle
=
\mathbf{q}_{\M}^\top R(\Delta_1+\Delta_2)\mathbf{q}_{\M}.
\]
These quadratic forms are not positive for every vector.
Under isotropic Gaussian reference distributions, their expectations are proportional to the RoPE trace, which is positive in the distance regimes analyzed above.
This reference calculation does not determine their signs for arbitrary learned parameters.

For ALiBi,
\[
S_B
=
\mathbf{q}_{\M}^{\top}\mathbf{k}_B+b_{\mathrm{pos}}(\Delta),
\]
so the additive positional bias vanishes under differentiation:
\[
\nabla_{\mathbf{q}}S_B=\mathbf{k}_B,
\quad
\nabla_{\mathbf{k}}S_B=\mathbf{q}_{\M}.
\]
The geometric factors are therefore \(\|\mathbf{k}_B\|^2\) and
\(\|\mathbf{q}_{\M}\|^2\), both nonnegative and positive for nonzero vectors.
This explains why ALiBi preserves query-key routing alignment more strongly
than RoPE.

\paragraph{Summary.}
For this reduced model, the decomposition isolates a sign-determined unembedding contribution from value, routing, and embedding cross-terms whose signs depend on the learned parameters.
It provides a factorized-parameter perspective on the shared-storage result without asserting positivity of the full gradient.

\subsection{Probability-level Transfer in a Single Evidence Token Setting}
\label{app:probability_transfer}

The gradient-alignment results above show that a forward masked update moves the shared storage column \(Y_{:,B}\) in a direction that decreases the reverse loss to first order.
Here we give a minimal probability-level training-dynamics counterpart to the analysis of \citet{zhu2024towards}, adapted to the MDM setting.
The result is intentionally idealized: it isolates a single-evidence-token mechanism and should not be read as a claim that reverse probabilities generally saturate in practical multi-token or multi-layer MDMs.
The distinction is that, in an MDM, storing the \(B\)-to-\(A\) evidence is not by itself enough.
Forward training can make the shared column \(Y_{:,B}\) strongly favor \(A\), but the reverse prompt benefits from this storage only if its masked position actually attends to \(B\).
This routing issue is absent in the single-prefix ARM comparison of App.~\ref{app:arm_comparison}, where the relevant attention weight is fixed at one.
Our result therefore makes the two steps explicit: forward training separates \(Y_{A,B}\) from \(Y_{C,B}\) for \(C\neq A\), and reverse transfer follows when the reverse route to \(B\) remains active.

\paragraph{Setup.}
We work with the effective parameters \(Y,Q,K\) in the auxiliary two-route model of App.~\ref{app:original_parameter_alignment}.
The reference route has fixed zero score and value, so only the evidence route contributes to the logits.
Both the forward and reverse prompts contain one masked position and one informative evidence token \(B\).
Although this is a single evidence token setting, we keep explicit relative distances in order to track the effect of positional encoding on the attention score.
In the forward prompt, \(B\) lies \(\Delta_1\) positions to the right of the mask.
In the reverse prompt, \(B\) lies \(\Delta_2\) positions to the left of the mask.
Both prompts ask the model to recover the same target token \(A\).

We isolate the evidence-token contribution to the masked-position logits:
\[
\mathbf{z}_{\mathrm{fwd}}(t)
=
\alpha_B^{\mathrm{fwd}}(t)Y_{:,B}(t),
\quad
\mathbf{z}_{\mathrm{rev}}(t)
=
\alpha_B^{\mathrm{rev}}(t)Y_{:,B}(t).
\]
Here
\[
\alpha_B(t)
=
\sigma(S_B(t)/\sqrt{D}),
\]
where \(\sigma\) is the sigmoid function.
This is a softmax between the evidence score and the fixed reference score, not the full two-token model with a trainable mask route.
The prediction distributions are
\[
\mathbf{p}_{\mathrm{fwd}}(t)
=
\operatorname{softmax}(\mathbf{z}_{\mathrm{fwd}}(t)),
\quad
\mathbf{p}_{\mathrm{rev}}(t)
=
\operatorname{softmax}(\mathbf{z}_{\mathrm{rev}}(t)).
\]
We use gradient flow on the forward loss
\[
\mathcal{L}_{\mathrm{fwd}}(t)
=
-\log p_{\mathrm{fwd}}(A;t)
\]
with respect to the shared storage column \(Y_{:,B}\) and the routing variables specified below.
Here \(t\) denotes gradient-flow time, not the masking probability used in the diffusion objective.
For a clean statement, we initialize the shared column uniformly and write this as
\[
Y_{:,B}(0)=\mathbf{0}.
\]

\paragraph{RoPE setup.}
For RoPE, the query and key play symmetric roles in the bilinear attention score.
To isolate the routing geometry in the simplest form, we fix the masked-token query \(\mathbf{q}_{\M}\) and update only the evidence-token key \(\mathbf{k}_B(t)\) by gradient flow on the forward loss.
The forward and reverse scores are
\[
S_{\mathrm{fwd}}(t)
=
\mathbf{q}_{\M}^{\top}R(\Delta_1)\mathbf{k}_B(t),
\quad
S_{\mathrm{rev}}(t)
=
\mathbf{q}_{\M}^{\top}R(-\Delta_2)\mathbf{k}_B(t).
\]
We assume the RoPE geometric alignment condition
\[
\mathbf{q}_{\M}^{\top}R(\Delta_1+\Delta_2)\mathbf{q}_{\M}>0.
\]
This is the same trace-controlled geometry that appears in the RoPE attention-correlation analysis.
For example, if \(\mathbf{q}_{\M}\sim\mathcal{N}(0,\sigma^2 I_D)\), then
\[
\mathbb{E}\left[
\mathbf{q}_{\M}^{\top}R(\Delta_1+\Delta_2)\mathbf{q}_{\M}
\right]
=
\sigma^2\operatorname{Tr}(R(\Delta_1+\Delta_2)).
\]
Thus, the corresponding quadratic form has positive expectation whenever the RoPE trace is positive.
This is the same trace quantity with argument \(r=\Delta_1+\Delta_2\); App.~\ref{app:rope_trace_bound} verifies positivity over the relevant range.

\begin{proposition}[RoPE Probability Transfer with a Fixed Query]
\label{prop:probability_transfer_rope_fixed_q}
Assume the RoPE setup above.
Then
\[
\lim_{t\to\infty}p_{\mathrm{fwd}}(A;t)=1,
\quad
\lim_{t\to\infty}p_{\mathrm{rev}}(A;t)=1.
\]
\end{proposition}

\paragraph{ALiBi setup.}
For ALiBi, write the forward and reverse scores as
\[
S_{\mathrm{fwd}}(t)
=
\mathbf{q}_{\M}(t)^\top\mathbf{k}_B(t)+b\Delta_1,
\quad
S_{\mathrm{rev}}(t)
=
\mathbf{q}_{\M}(t)^\top\mathbf{k}_B(t)-b\Delta_2.
\]
Here \(b\) is the fixed ALiBi bias slope, and \(\Delta_1,\Delta_2\) are the forward and reverse distances defined above.
We allow the routing variables \(\mathbf{q}_{\M}(t)\) and \(\mathbf{k}_B(t)\) to be either fixed or updated by gradient flow on the forward loss.

\begin{proposition}[ALiBi Probability Transfer]
\label{prop:probability_transfer_alibi}
Assume the ALiBi setup above.
Then
\[
\lim_{t\to\infty}p_{\mathrm{fwd}}(A;t)=1,
\quad
\lim_{t\to\infty}p_{\mathrm{rev}}(A;t)=1.
\]
\end{proposition}

\paragraph{Proof of Propositions~\ref{prop:probability_transfer_rope_fixed_q} and~\ref{prop:probability_transfer_alibi}.}
We first prove the shared storage part, which is common to both propositions.
For the forward prompt,
\[
(\mathbf{z}_{\mathrm{fwd}}(t))_C
=
\alpha_B^{\mathrm{fwd}}(t)Y_{C,B}(t),
\]
and therefore
\[
p_{\mathrm{fwd}}(C;t)
=
\frac{
\exp(\alpha_B^{\mathrm{fwd}}(t)Y_{C,B}(t))
}{
\sum_{C'\in\mathcal{V}}
\exp(\alpha_B^{\mathrm{fwd}}(t)Y_{C',B}(t))
}.
\]
The gradient of the forward loss with respect to \(Y_{C,B}\) is
\[
\nabla_{Y_{C,B}}\mathcal{L}_{\mathrm{fwd}}
=
\alpha_B^{\mathrm{fwd}}(t)
\left(
p_{\mathrm{fwd}}(C;t)-\delta_{C,A}
\right).
\]
Thus, under gradient flow with rate \(\eta>0\),
\[
\frac{d}{dt}Y_{A,B}(t)
=
\eta\alpha_B^{\mathrm{fwd}}(t)
\left(
1-p_{\mathrm{fwd}}(A;t)
\right),
\]
and, for \(C\neq A\),
\[
\frac{d}{dt}Y_{C,B}(t)
=
-\eta\alpha_B^{\mathrm{fwd}}(t)p_{\mathrm{fwd}}(C;t).
\]
Hence
\[
\frac{d}{dt}
\left(
Y_{A,B}(t)-Y_{C,B}(t)
\right)
=
\eta\alpha_B^{\mathrm{fwd}}(t)
\left(
1-p_{\mathrm{fwd}}(A;t)+p_{\mathrm{fwd}}(C;t)
\right)
>0.
\]
Since \(Y_{:,B}(0)=\mathbf{0}\), we have
\[
Y_{A,B}(t)-Y_{C,B}(t)\ge 0
\quad
(C\neq A)
\]
for all \(t\).

We next show that the forward route to \(B\) does not collapse.
First,
\[
\frac{\partial \mathcal{L}_{\mathrm{fwd}}}{\partial \alpha_B^{\mathrm{fwd}}}
=
(\mathbf{p}_{\mathrm{fwd}}(t)-\mathbf{e}_A)^\top Y_{:,B}(t).
\]
Using
\[
1-p_{\mathrm{fwd}}(A;t)
=
\sum_{C\neq A}p_{\mathrm{fwd}}(C;t),
\]
we obtain
\[
\begin{aligned}
(\mathbf{p}_{\mathrm{fwd}}(t)-\mathbf{e}_A)^\top Y_{:,B}(t)
&=
\sum_{C\neq A}p_{\mathrm{fwd}}(C;t)Y_{C,B}(t)
-
\left(1-p_{\mathrm{fwd}}(A;t)\right)Y_{A,B}(t)\\
&=
\sum_{C\neq A}p_{\mathrm{fwd}}(C;t)
\left(
Y_{C,B}(t)-Y_{A,B}(t)
\right)\\
&=
-\sum_{C\neq A}p_{\mathrm{fwd}}(C;t)
\left(
Y_{A,B}(t)-Y_{C,B}(t)
\right)
\le 0.
\end{aligned}
\]
Therefore,
\[
\frac{\partial \mathcal{L}_{\mathrm{fwd}}}{\partial S_{\mathrm{fwd}}}
=
\frac{
\alpha_B^{\mathrm{fwd}}(t)
\left(1-\alpha_B^{\mathrm{fwd}}(t)\right)
}{\sqrt{D}}
\frac{\partial \mathcal{L}_{\mathrm{fwd}}}{\partial \alpha_B^{\mathrm{fwd}}}
\le 0.
\]

For RoPE, gradient flow on \(\mathbf{k}_B(t)\) gives
\[
\frac{d}{dt}\mathbf{k}_B(t)
=
-\eta
\frac{\partial \mathcal{L}_{\mathrm{fwd}}}{\partial S_{\mathrm{fwd}}}
R(-\Delta_1)\mathbf{q}_{\M}.
\]
Hence
\[
\begin{aligned}
\frac{d}{dt}S_{\mathrm{fwd}}(t)
&=
\mathbf{q}_{\M}^{\top}R(\Delta_1)
\frac{d}{dt}\mathbf{k}_B(t)\\
&=
-\eta
\frac{\partial \mathcal{L}_{\mathrm{fwd}}}{\partial S_{\mathrm{fwd}}}
\mathbf{q}_{\M}^{\top}R(\Delta_1)R(-\Delta_1)\mathbf{q}_{\M}\\
&=
-\eta
\frac{\partial \mathcal{L}_{\mathrm{fwd}}}{\partial S_{\mathrm{fwd}}}
\left\|\mathbf{q}_{\M}\right\|^2
\ge 0.
\end{aligned}
\]
Thus
\[
S_{\mathrm{fwd}}(t)\ge S_{\mathrm{fwd}}(0),
\]
and
\[
\alpha_B^{\mathrm{fwd}}(t)
=
\sigma(S_{\mathrm{fwd}}(t)/\sqrt{D})
\ge
\sigma(S_{\mathrm{fwd}}(0)/\sqrt{D})
>0.
\]

For ALiBi, if the routing variables are fixed, then
\[
\mathbf{q}_{\M}(t)^\top\mathbf{k}_B(t)
=
\mathbf{q}_{\M}(0)^\top\mathbf{k}_B(0).
\]
If they are updated by gradient flow on the forward loss, then
\[
\frac{d}{dt}\mathbf{q}_{\M}(t)
=
-\eta
\frac{\partial \mathcal{L}_{\mathrm{fwd}}}{\partial S_{\mathrm{fwd}}}
\mathbf{k}_B(t),
\]
and
\[
\frac{d}{dt}\mathbf{k}_B(t)
=
-\eta
\frac{\partial \mathcal{L}_{\mathrm{fwd}}}{\partial S_{\mathrm{fwd}}}
\mathbf{q}_{\M}(t).
\]
Therefore,
\[
\begin{aligned}
\frac{d}{dt}
\left(
\mathbf{q}_{\M}(t)^\top\mathbf{k}_B(t)
\right)
&=
\left(\frac{d}{dt}\mathbf{q}_{\M}(t)\right)^\top\mathbf{k}_B(t)
+
\mathbf{q}_{\M}(t)^\top
\left(\frac{d}{dt}\mathbf{k}_B(t)\right)\\
&=
-\eta
\frac{\partial \mathcal{L}_{\mathrm{fwd}}}{\partial S_{\mathrm{fwd}}}
\left(
\left\|\mathbf{k}_B(t)\right\|^2
+
\left\|\mathbf{q}_{\M}(t)\right\|^2
\right)\\
&\ge 0.
\end{aligned}
\]
Thus, in either ALiBi case, \(S_{\mathrm{fwd}}(t)\) is bounded below, and hence
\[
\alpha_B^{\mathrm{fwd}}(t)
=
\sigma(S_{\mathrm{fwd}}(t)/\sqrt{D})
\]
is bounded away from zero.

We now prove that all storage margins diverge.
Fix \(C\neq A\).
Suppose for contradiction that
\[
Y_{A,B}(t)-Y_{C,B}(t)
\]
is bounded above.
Since it is nondecreasing, there exists \(M<\infty\) such that
\[
0
\le
Y_{A,B}(t)-Y_{C,B}(t)
\le M
\]
for all \(t\).
Also, for every \(C'\neq A\),
\[
Y_{A,B}(t)-Y_{C',B}(t)\ge 0.
\]
Therefore,
\[
\begin{aligned}
p_{\mathrm{fwd}}(C;t)
&=
\frac{
\exp\left(
\alpha_B^{\mathrm{fwd}}(t)Y_{C,B}(t)
\right)
}{
\exp\left(
\alpha_B^{\mathrm{fwd}}(t)Y_{A,B}(t)
\right)
+
\sum_{C'\neq A}
\exp\left(
\alpha_B^{\mathrm{fwd}}(t)Y_{C',B}(t)
\right)
}\\
&=
\frac{
\exp\left(
-\alpha_B^{\mathrm{fwd}}(t)
\left(
Y_{A,B}(t)-Y_{C,B}(t)
\right)
\right)
}{
1+
\sum_{C'\neq A}
\exp\left(
-\alpha_B^{\mathrm{fwd}}(t)
\left(
Y_{A,B}(t)-Y_{C',B}(t)
\right)
\right)
}.
\end{aligned}
\]
Since \(0<\alpha_B^{\mathrm{fwd}}(t)\le 1\),
\[
\exp\left(
-\alpha_B^{\mathrm{fwd}}(t)
\left(
Y_{A,B}(t)-Y_{C,B}(t)
\right)
\right)
\ge
e^{-M}.
\]
Moreover,
\[
\exp\left(
-\alpha_B^{\mathrm{fwd}}(t)
\left(
Y_{A,B}(t)-Y_{C',B}(t)
\right)
\right)
\le 1
\quad
(C'\neq A).
\]
Hence
\[
p_{\mathrm{fwd}}(C;t)
\ge
\frac{e^{-M}}{|\mathcal{V}|}
>0.
\]
Using the positive lower bound on \(\alpha_B^{\mathrm{fwd}}(t)\), we get
\[
\begin{aligned}
\frac{d}{dt}
\left(
Y_{A,B}(t)-Y_{C,B}(t)
\right)
&=
\eta\alpha_B^{\mathrm{fwd}}(t)
\left(
1-p_{\mathrm{fwd}}(A;t)+p_{\mathrm{fwd}}(C;t)
\right)\\
&\ge
\eta\alpha_B^{\mathrm{fwd}}(t)p_{\mathrm{fwd}}(C;t)\\
&\ge
\eta
\left(
\inf_{s\ge 0}\alpha_B^{\mathrm{fwd}}(s)
\right)
\frac{e^{-M}}{|\mathcal{V}|}
>0.
\end{aligned}
\]
This contradicts the boundedness of
\[
Y_{A,B}(t)-Y_{C,B}(t).
\]
Therefore,
\[
Y_{A,B}(t)-Y_{C,B}(t)\to+\infty
\quad
(C\neq A).
\]

It follows that
\[
\begin{aligned}
p_{\mathrm{fwd}}(A;t)
&=
\frac{
\exp\left(
\alpha_B^{\mathrm{fwd}}(t)Y_{A,B}(t)
\right)
}{
\sum_{C'\in\mathcal{V}}
\exp\left(
\alpha_B^{\mathrm{fwd}}(t)Y_{C',B}(t)
\right)
}\\
&=
\frac{1}{
1+
\sum_{C\neq A}
\exp\left(
-\alpha_B^{\mathrm{fwd}}(t)
\left(
Y_{A,B}(t)-Y_{C,B}(t)
\right)
\right)
}.
\end{aligned}
\]
Since \(\alpha_B^{\mathrm{fwd}}(t)\) is bounded away from zero and
\[
Y_{A,B}(t)-Y_{C,B}(t)\to+\infty
\quad
(C\neq A),
\]
each exponential term vanishes.
Thus
\[
p_{\mathrm{fwd}}(A;t)\to 1.
\]

It remains to check the reverse route to \(B\).
For RoPE,
\[
\frac{d}{dt}\mathbf{k}_B(t)
=
-\eta
\frac{\partial \mathcal{L}_{\mathrm{fwd}}}{\partial S_{\mathrm{fwd}}}
R(-\Delta_1)\mathbf{q}_{\M},
\]
with
\[
-\eta
\frac{\partial \mathcal{L}_{\mathrm{fwd}}}{\partial S_{\mathrm{fwd}}}
\ge 0.
\]
Therefore, for some nondecreasing \(\lambda(t)\ge 0\),
\[
\mathbf{k}_B(t)
=
\mathbf{k}_B(0)
+
\lambda(t)R(-\Delta_1)\mathbf{q}_{\M}.
\]
Then
\[
\begin{aligned}
S_{\mathrm{rev}}(t)
&=
\mathbf{q}_{\M}^{\top}R(-\Delta_2)\mathbf{k}_B(t)\\
&=
\mathbf{q}_{\M}^{\top}R(-\Delta_2)\mathbf{k}_B(0)
+
\lambda(t)
\mathbf{q}_{\M}^{\top}R(-\Delta_2)R(-\Delta_1)\mathbf{q}_{\M}\\
&=
\mathbf{q}_{\M}^{\top}R(-\Delta_2)\mathbf{k}_B(0)
+
\lambda(t)
\mathbf{q}_{\M}^{\top}R(-(\Delta_1+\Delta_2))\mathbf{q}_{\M}.
\end{aligned}
\]
For any RoPE rotation \(R(r)\),
\[
\mathbf{q}_{\M}^{\top}R(-r)\mathbf{q}_{\M}
=
\left(
\mathbf{q}_{\M}^{\top}R(-r)\mathbf{q}_{\M}
\right)^\top
=
\mathbf{q}_{\M}^{\top}R(-r)^\top\mathbf{q}_{\M}
=
\mathbf{q}_{\M}^{\top}R(r)\mathbf{q}_{\M}.
\]
Thus
\[
\mathbf{q}_{\M}^{\top}R(-(\Delta_1+\Delta_2))\mathbf{q}_{\M}
=
\mathbf{q}_{\M}^{\top}R(\Delta_1+\Delta_2)\mathbf{q}_{\M}.
\]
By the RoPE geometric alignment condition,
\[
\mathbf{q}_{\M}^{\top}R(\Delta_1+\Delta_2)\mathbf{q}_{\M}>0.
\]
Since \(\lambda(t)\ge 0\), this gives
\[
S_{\mathrm{rev}}(t)
\ge
\mathbf{q}_{\M}^{\top}R(-\Delta_2)\mathbf{k}_B(0).
\]
Therefore
\[
\alpha_B^{\mathrm{rev}}(t)
=
\sigma(S_{\mathrm{rev}}(t)/\sqrt{D})
\ge
\sigma\left(
\mathbf{q}_{\M}^{\top}R(-\Delta_2)\mathbf{k}_B(0)/\sqrt{D}
\right)
>0.
\]

For ALiBi,
\[
S_{\mathrm{fwd}}(t)
=
\mathbf{q}_{\M}(t)^\top\mathbf{k}_B(t)+b\Delta_1,
\]
and
\[
S_{\mathrm{rev}}(t)
=
\mathbf{q}_{\M}(t)^\top\mathbf{k}_B(t)-b\Delta_2.
\]
Hence
\[
S_{\mathrm{rev}}(t)
=
S_{\mathrm{fwd}}(t)-b(\Delta_1+\Delta_2).
\]
Since \(S_{\mathrm{fwd}}(t)\) is bounded below and \(b(\Delta_1+\Delta_2)\) is finite, \(S_{\mathrm{rev}}(t)\) is bounded below.
Thus
\[
\alpha_B^{\mathrm{rev}}(t)
=
\sigma(S_{\mathrm{rev}}(t)/\sqrt{D})
\]
is bounded away from zero.

Finally,
\[
\begin{aligned}
p_{\mathrm{rev}}(A;t)
&=
\frac{
\exp\left(
\alpha_B^{\mathrm{rev}}(t)Y_{A,B}(t)
\right)
}{
\sum_{C'\in\mathcal{V}}
\exp\left(
\alpha_B^{\mathrm{rev}}(t)Y_{C',B}(t)
\right)
}\\
&=
\frac{1}{
1+
\sum_{C\neq A}
\exp\left(
-\alpha_B^{\mathrm{rev}}(t)
\left(
Y_{A,B}(t)-Y_{C,B}(t)
\right)
\right)
}.
\end{aligned}
\]
Since \(\alpha_B^{\mathrm{rev}}(t)\) is bounded away from zero and
\[
Y_{A,B}(t)-Y_{C,B}(t)\to+\infty
\quad
(C\neq A),
\]
each exponential term vanishes.
Therefore,
\[
p_{\mathrm{rev}}(A;t)\to 1.
\]
This proves both propositions.
\qed

\paragraph{Interpretation.}
These propositions isolate probability-level transfer in a single evidence token setting.
Forward training is not guaranteed to raise the forward and reverse probabilities at the same rate.
However, the result is still useful: once training makes \(Y_{:,B}\) increasingly favor \(A\), the reverse probability also converges as long as the reverse attention to \(B\) stays bounded away from zero.
Thus the result gives sufficient storage and attention conditions under which forward learning becomes reusable in the reverse prompt.

ALiBi and RoPE differ in how naturally they keep this reverse attention active.
For ALiBi, the forward and reverse scores differ only by a finite additive bias, so active forward attention implies active reverse attention.
For RoPE, this depends on the geometric factor \(\mathbf{q}_{\M}^{\top}R(\Delta_1+\Delta_2)\mathbf{q}_{\M}\), which has positive expectation under isotropy in the trace-positive regimes studied in App.~\ref{app:rope_trace_bound}, but need not be positive for every query vector.
Thus the RoPE statement is weaker: it captures the favorable routing geometry that typically supports transfer, rather than an unconditional guarantee.

\paragraph{Role of additional context tokens.}
The clean convergence result relies on the single evidence token setting.
If additional context tokens are present, they can also receive attention, contribute to the logits, and interact with the routing dynamics.
These terms may dilute the direct transfer path through \(Y_{:,B}\), so the reverse probability need not saturate as cleanly as in the minimal proof.
The value of the proposition is therefore mechanistic: it identifies a concrete probability-level route from forward learning to reverse prediction, while showing that the strength of this route depends on positional attention geometry.

\subsection{Connection to Prior One-Layer ARM Analysis}
\label{app:arm_comparison}

\citet{zhu2024towards} analyze a one-layer causal Transformer on three-token entity--relation--entity sequences with zero-initialized effective weights and batch-size-one SGD under their stated data and learning-rate assumptions.
In our input--target index convention, forward learning strengthens \(Y_{B,A}\) while leaving the reverse-relevant entry \(Y_{A,B}\) at its initial value.
Their Theorem~3 separates forward target probabilities approaching one from held-out reverse target probabilities at most \(1/|\mathcal{V}|\).
Below we establish the corresponding causal storage asymmetry and derive a two-token gradient-flow analogue in our effective coordinates, rather than restating their full model and optimization setup.

\paragraph{Setup.}
Consider the one-layer model of Sec.~\ref{sec:setup} and App.~\ref{app:full_grad_update} with the same effective parameters
\[
Y=W_UW_VW_E,
\quad
Q=W_QW_E,
\quad
K=W_KW_E,
\]
but replace full attention and masked reconstruction by causal attention and the standard left-to-right next-token loss.
Suppose the forward target is token~\(B\), and its causal context contains token~\(A\) but no occurrence of \(B\).
Let \(\beta_A>0\) denote the total attention mass assigned to all occurrences of \(A\) in this context.
The forward logits are
\[
\mathbf{z}_{\mathrm{fwd}}
=
\sum_{k\text{ in context}}\alpha_kY_{:,x^k},
\quad
\mathbf{p}_{\mathrm{fwd}}
=
\operatorname{softmax}(\mathbf{z}_{\mathrm{fwd}}).
\]

\begin{proposition}[Causal Storage Asymmetry]
\label{prop:causal_storage_asymmetry}
For the forward loss
\[
\mathcal{L}_{\mathrm{fwd}}^{\mathrm{ARM}}
=
-\log p_{\mathrm{fwd}}(B),
\]
the effective storage gradients satisfy
\[
\frac{\partial\mathcal{L}_{\mathrm{fwd}}^{\mathrm{ARM}}}{\partial Y_{B,A}}
=
-\beta_A(1-p_{\mathrm{fwd}}(B))<0,
\quad
\frac{\partial\mathcal{L}_{\mathrm{fwd}}^{\mathrm{ARM}}}{\partial Y_{C,A}}
=
\beta_A p_{\mathrm{fwd}}(C)>0
\quad(C\neq B),
\]
while
\[
\frac{\partial\mathcal{L}_{\mathrm{fwd}}^{\mathrm{ARM}}}{\partial Y_{C,B}}=0
\quad\text{for all }C.
\]
Thus forward training strengthens the \(A\)-to-\(B\) storage margin but does not update the reverse-relevant column \(Y_{:,B}\).
\end{proposition}

\begin{proof}
Since
\[
\mathbf{z}_{\mathrm{fwd}}
=
\sum_{k\text{ in context}}\alpha_kY_{:,x^k},
\]
we have
\[
\frac{\partial\mathcal{L}_{\mathrm{fwd}}^{\mathrm{ARM}}}{\partial Y_{C,A}}
=
\beta_A\bigl(p_{\mathrm{fwd}}(C)-\mathbf{1}\{C=B\}\bigr).
\]
The forward loss does not depend on \(Y_{:,B}\), since \(B\) does not appear in the causal context, which gives the remaining statement.
\end{proof}

This is the causal counterpart of Proposition~\ref{prop:shared_y_update}.
The difference is the direction of supervision: left-to-right training on a context containing \(A\) but not \(B\) strengthens \(Y_{B,A}\), whereas masked training can use \(B\) as visible evidence for target~\(A\) and therefore update the reverse-relevant entry \(Y_{A,B}\).

\paragraph{Two-token probability-level comparison.}
We now specialize to a two-token sequence \(AB\) with distinct tokens \(A\) and \(B\).
The representation predicting \(B\) has the single-token causal prefix \(A\), so its attention weight is identically one:
\[
\mathbf{z}_{\mathrm{fwd}}=Y_{:,A},
\quad
\mathbf{p}_{\mathrm{fwd}}(t)=\operatorname{softmax}(Y_{:,A}(t)).
\]
The held-out reverse sequence \(BA\) analogously gives
\[
\mathbf{z}_{\mathrm{rev}}=Y_{:,B},
\quad
\mathbf{p}_{\mathrm{rev}}(t)=\operatorname{softmax}(Y_{:,B}(t)).
\]
Thus \(Q\) and \(K\) play no role in this single-prefix reduction.
This two-token analogue isolates the directional storage asymmetry studied by \citet{zhu2024towards}, without additional context-dependent routing; adding a relation token can change the detailed dynamics.

\begin{theorem}[Probability-Level ARM Reversal]
\label{thm:arm_probability_reversal}
Assume gradient flow on \(\mathcal{L}_{\mathrm{fwd}}^{\mathrm{ARM}}\) with respect to the independent effective storage parameter \(Y\), with rate \(\eta>0\), and
\[
Y_{:,A}(0)=Y_{:,B}(0)=0.
\]
Then
\[
\lim_{t\to\infty}p_{\mathrm{fwd}}(B;t)=1,
\quad
p_{\mathrm{rev}}(A;t)=\frac{1}{|\mathcal{V}|}
\quad\text{for all }t\ge 0.
\]
\end{theorem}

\begin{proof}
For \(C\neq B\), define
\[
m_C(t)=Y_{B,A}(t)-Y_{C,A}(t).
\]
Proposition~\ref{prop:causal_storage_asymmetry}, with \(\beta_A=1\), gives
\[
\frac{d}{dt}m_C(t)
=
\eta\left(1-p_{\mathrm{fwd}}(B;t)+p_{\mathrm{fwd}}(C;t)\right)>0.
\]
Since \(m_C(0)=0\), all margins remain nonnegative.
If some \(m_C(t)\) were bounded above by \(M<\infty\), then
\[
p_{\mathrm{fwd}}(C;t)
=
\frac{e^{-m_C(t)}}{1+\sum_{C'\neq B}e^{-m_{C'}(t)}}
\ge
\frac{e^{-M}}{|\mathcal{V}|},
\]
and hence \(dm_C/dt\ge\eta e^{-M}/|\mathcal{V}|>0\), a contradiction.
Therefore
\[
Y_{B,A}(t)-Y_{C,A}(t)\to\infty
\quad(C\neq B),
\]
which implies \(p_{\mathrm{fwd}}(B;t)\to 1\).
Meanwhile, the forward loss never updates \(Y_{:,B}\), so \(Y_{:,B}(t)=0\) for all \(t\).
Hence
\[
p_{\mathrm{rev}}(A;t)
=
\frac{e^0}{\sum_{C\in\mathcal{V}}e^0}
=
\frac{1}{|\mathcal{V}|}.
\]
\end{proof}

The theorem is the probability-level counterpart of Theorem~3 of \citet{zhu2024towards}: the observed direction can be learned perfectly while the unseen reverse remains at random guess.
The contrast with the MDM result in App.~\ref{app:probability_transfer} is direct.
Under masked training, the same observed pair can produce a forward example with visible \(B\) and target~\(A\), directly strengthening \(Y_{A,B}\); under the additional assumptions of App.~\ref{app:probability_transfer}, the corresponding reverse probability converges to one.
Under left-to-right training, the observed sequence \(AB\) instead strengthens \(Y_{B,A}\), while \(Y_{A,B}\) receives no update.

\section{Experimental Details and Additional Results}
\label{app:exp_details}

\paragraph{Compute resources.}
All experiments were conducted on a single NVIDIA L40S GPU with 48GB of GPU memory.
The one-layer RADD experiments, GPT-2-style causal Transformer baselines, and positional-encoding interventions each used a single L40S GPU and took less than one hour per run.
All large-scale LoRA fine-tuning experiments, including both MDMs and ARM baselines, also used a single L40S 48GB GPU per run.
Each MDM fine-tuning run took approximately 7--8 hours, with Dream runs taking about 7.5 hours, while ARM fine-tuning runs were of comparable or shorter duration.
Across all reported experiments, including repeated random seeds, the total compute was approximately 500 GPU-hours.
Additional pilot and debugging runs used the same hardware and did not require substantially more compute than the reported experiments.

\subsection{Details for One-Layer Validation}
\label{app:one_layer_exp_details}

\paragraph{Data construction.}
For each sequence length \(L\), we construct synthetic sequences containing one lowercase--uppercase pair and \(L-2\) placeholder tokens.
The lowercase token is the target token, and the uppercase token is the evidence token.
Training includes only forward-order sequences, where the lowercase target precedes the uppercase evidence token.
All reverse-order sequences are excluded from training.
At evaluation time, we query both forward and reverse configurations from the same pair.
This construction isolates whether evidence learned from the forward order can be reused when the input and target roles are positionally reversed.

\begin{table}[H]

\centering
\caption{Hyperparameters for the one-layer experiments.}
\vspace{0.5em}
\label{tab:one_layer_hyperparams}
\small
\setlength{\tabcolsep}{4pt}

\begin{tabular}{lc}
\toprule
\textbf{Hyperparameter} & \textbf{Value} \\
\midrule
Batch size & \(256\) \\
Learning rate & \(3\times 10^{-4}\) \\
Training steps & \(3{,}000\) \\
Gradient clipping & \(1.0\) \\
Weight decay & \(0.0\) \\
Dropout & \(0.02\) \\
Warmup steps & \(1{,}000\) \\
Hidden dimension & \(256\) \\
Attention heads & \(1\) \\
EMA decay for RADD & \(0.9999\) \\
\bottomrule
\end{tabular}

\end{table}

\paragraph{Models and hyperparameters.}
We train a one-layer RADD model as the MDM and a one-layer GPT-2-style causal Transformer as the ARM baseline.
Unless otherwise stated, both models use hidden dimension \(256\), one attention head, batch size \(256\), learning rate \(3\times 10^{-4}\), dropout \(0.02\), gradient clipping \(1.0\), and \(3{,}000\) training steps.
The RADD model additionally uses EMA with decay \(0.9999\).
All results are averaged over three random seeds, excluding one $L = 20$ run that failed to fit the forward direction (0\% forward accuracy).

\paragraph{Measurements.}
We measure three quantities corresponding to the theory.
First, we compute the Pearson correlation between forward and reverse attention scores assigned to the shared evidence token.
Second, we compute gradient cosine similarity over all trainable parameters between paired forward and reverse losses.
Third, we track the probability assigned to the correct target token under forward and reverse queries.
For the positional encoding intervention, we keep the data, model size, and training protocol fixed while changing only the positional encoding among learned absolute embeddings, RoPE, and ALiBi.

\clearpage

\begin{figure}[H]
\centering
\setlength{\tabcolsep}{4pt}

\begin{tabular}{@{}cc@{}}
\includegraphics[width=0.48\linewidth]{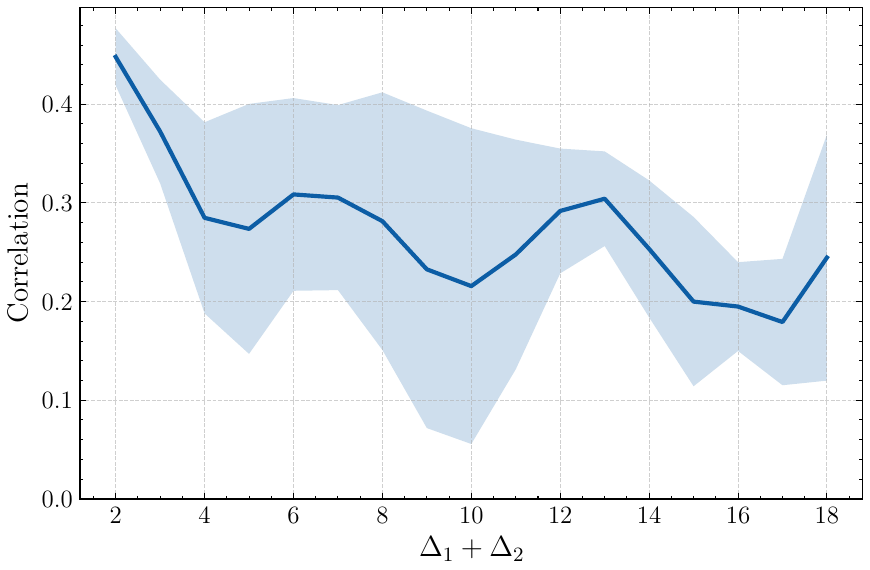} &
\includegraphics[width=0.48\linewidth]{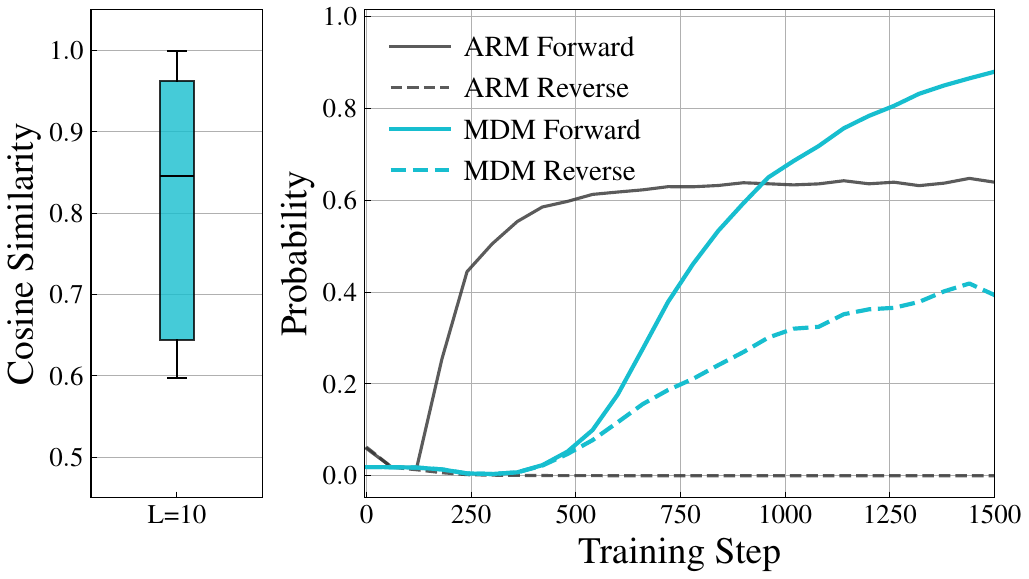} \\
\includegraphics[width=0.48\linewidth]{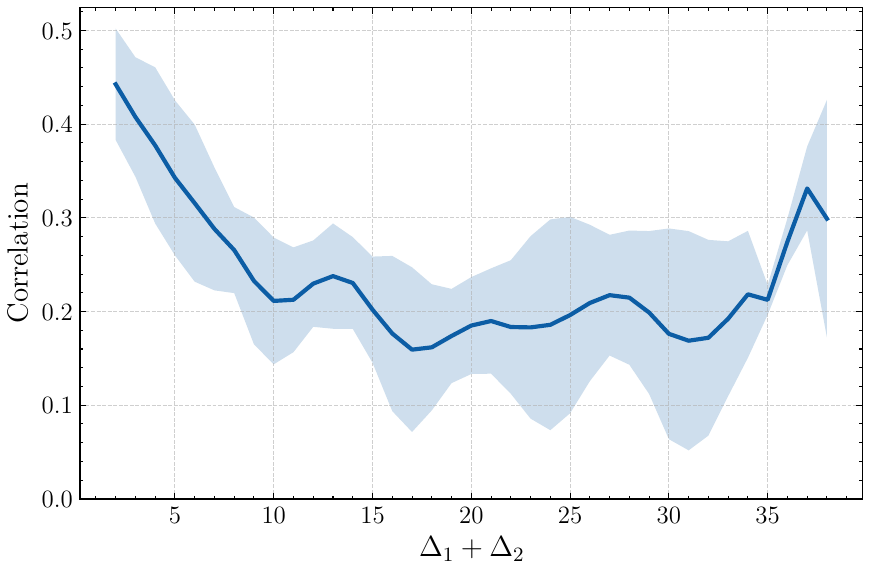} &
\includegraphics[width=0.48\linewidth]{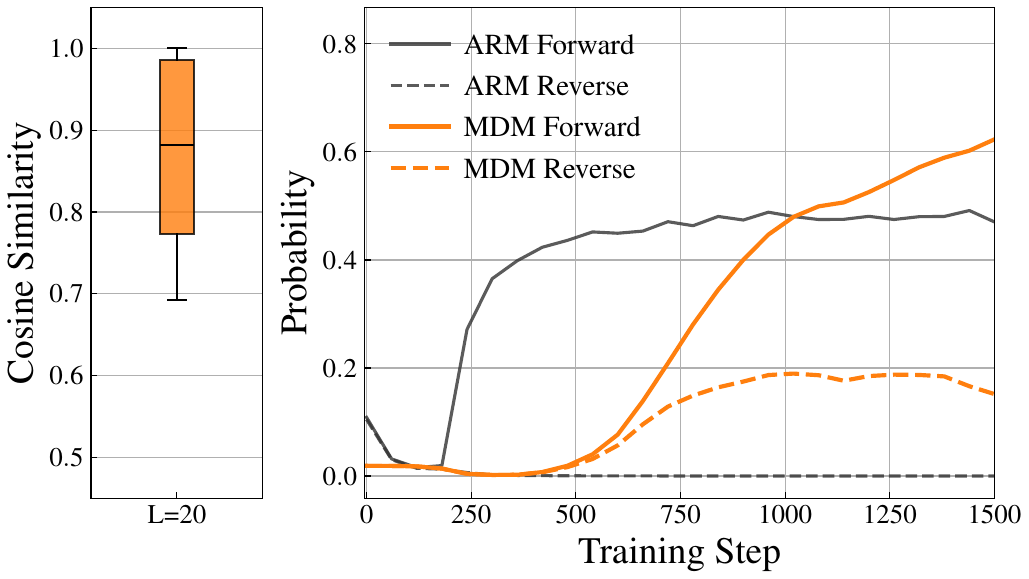} \\
\includegraphics[width=0.48\linewidth]{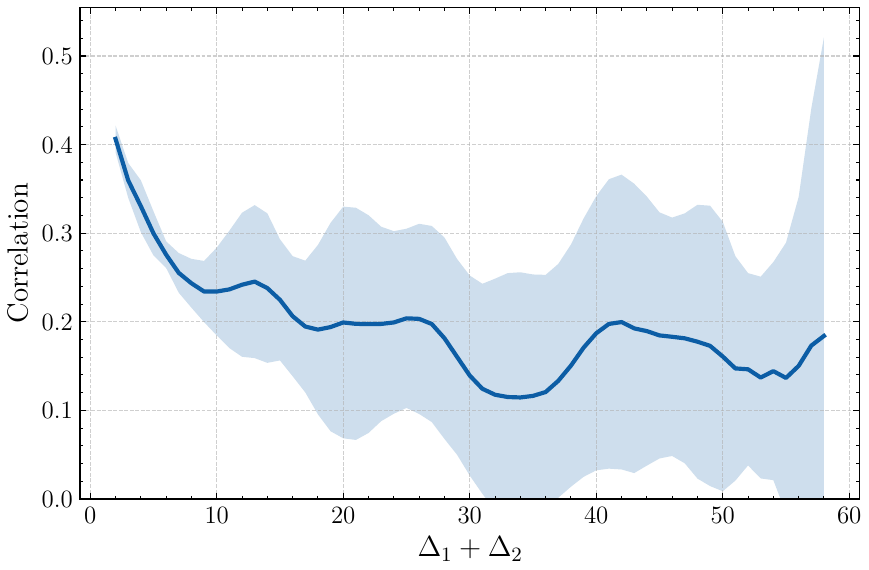} &
\includegraphics[width=0.48\linewidth]{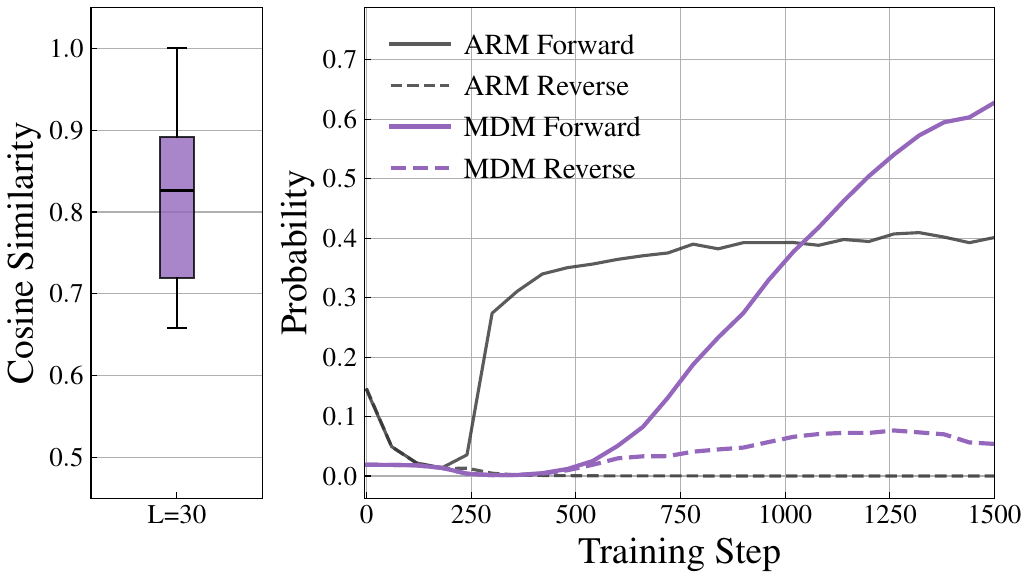} \\
\includegraphics[width=0.48\linewidth]{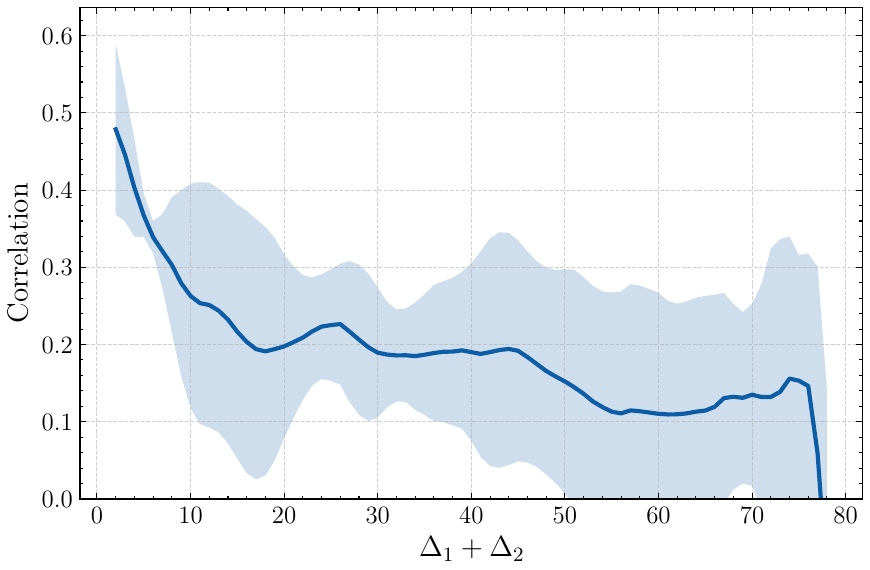} &
\includegraphics[width=0.48\linewidth]{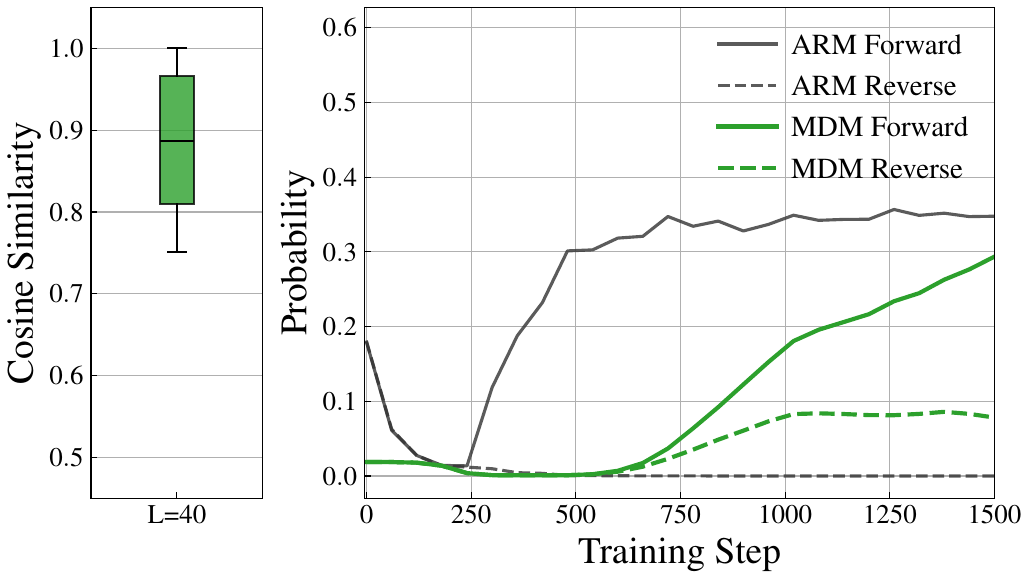}
\end{tabular}

\caption{
Additional one-layer validation across sequence lengths \(L=10,20,30,40\).
\textbf{Left column:} correlation of forward and reverse attention scores as a function of total relative distance \(\Delta_1+\Delta_2\).
\textbf{Right column:} gradient alignment and probability transfer during training.
}
\label{fig:appendix_attention_alignment_lengths}
\end{figure}

\begin{table}[H]
\centering
\caption{
Large-scale reversal evaluation on Parent--Child, Person--Description, and T-REx.
Results are exact-match accuracy (\%) averaged across three random seeds.
Fwd. denotes the trained forward direction, and Rev. denotes the held-out reverse direction.
MDMs retain substantially stronger reverse accuracy, whereas ARMs, including the reasoning-oriented Qwen-3.5 model, largely fail under order reversal.
}
\vspace{0.5em}
\label{tab:large_scale_reversal}
\small
\setlength{\tabcolsep}{3.5pt}

\resizebox{\textwidth}{!}{
\begin{tabular}{@{}l|cccc|cccccc@{}}
\toprule
\multicolumn{1}{@{}l|}{} 
& \multicolumn{4}{c|}{MDM} 
& \multicolumn{6}{c@{}}{ARM} \\
\cmidrule(lr){2-5} \cmidrule(lr){6-11}
& \multicolumn{2}{c}{LLaDA 8B} 
& \multicolumn{2}{c|}{Dream 7B} 
& \multicolumn{2}{c}{LLaMA-3.1 8B} 
& \multicolumn{2}{c}{Qwen-2.5 7B} 
& \multicolumn{2}{c@{}}{Qwen-3.5 4B} \\
Train Dataset 
& Fwd. & Rev. 
& Fwd. & Rev. 
& Fwd. & Rev. 
& Fwd. & Rev. 
& Fwd. & Rev. \\
\midrule
Parent \(\to\) Child (P2C)        
& 76.7 & 48.3
& 62.0 & 50.0
& 89.9 & 15.9 
& 89.9 & 0.5 
& 99.0 & 3.0 \\
Child \(\to\) Parent (C2P)        
& 87.7 & 43.7
& 83.0 & 54.0
& 95.9 & 6.9  
& 89.0 & 1.4 
& 100.0 & 1.0 \\
\midrule
Person \(\to\) Description (P2D)  
& 72.7 & 99.5
& 54.3 & 99.8 
& 72.7 & 3.5  
& 70.7 & 2.2 
& 81.3 & 0.0 \\
Description \(\to\) Person (D2P)  
& 99.7 & 41.3
& 100.0 & 27.2 
& 83.0 & 1.8  
& 80.0 & 1.5 
& 100.0 & 0.0 \\
\midrule
T-REx                           
& 92.3 & 81.5
& 83.0 & 35.0 
& 87.3 & 2.8  
& 89.8 & 2.3 
& 100.0 & 0.0 \\
\bottomrule
\end{tabular}
}
\end{table}
\subsection{Details for Large-Scale Validation}
\label{app:large_scale_details}

\paragraph{Datasets and evaluation protocol.}
We evaluate on three relation-reversal benchmarks: Parent--Child, Person--Description, and T-REx.
Parent--Child contains child--parent pairs annotated with parent type, Person--Description contains entity--description pairs, and T-REx contains entity--relation--description triples.
For each dataset, models are fine-tuned only on the forward direction and evaluated on both the trained forward direction and the held-out reverse direction.
For example, a Parent-to-Child training instance uses a parent entity as input and a child entity as target, while the reverse evaluation asks for the parent entity from the child entity.
We report exact-match accuracy after minimal normalization, including lowercasing and whitespace stripping.
For Parent--Child, either parent is accepted as correct when both father and mother are valid.

\paragraph{Models.}
We compare two masked diffusion language models and three autoregressive language models.
For MDMs, we use LLaDA-8B and Dream-7B.
For ARMs, we use LLaMA-3.1-8B, Qwen-2.5-7B, and Qwen-3.5-4B.
The exact model identifiers are:
\[
\begin{aligned}
&\text{LLaDA-8B: } \texttt{GSAI-ML/LLaDA-8B-Instruct},\\
&\text{Dream-7B: } \texttt{Dream-org/Dream-v0-Instruct-7B},\\
&\text{LLaMA-3.1-8B: } \texttt{meta-llama/Meta-Llama-3.1-8B-Instruct},\\
&\text{Qwen-2.5-7B: } \texttt{Qwen/Qwen2.5-7B-Instruct},\\
&\text{Qwen-3.5-4B: } \texttt{Qwen/Qwen3.5-4B}.
\end{aligned}
\]
Dream-7B is added to test whether the MDM reversal advantage persists beyond LLaDA.
Qwen-3.5-4B is added to test whether a reasoning-oriented ARM overcomes the same reversal failure.

\paragraph{Fine-tuning details.}
All models are fine-tuned with LoRA adapters.
Unless otherwise stated, we use rank \(r=32\), LoRA scaling \(\alpha=64\), AdamW optimizer, weight decay \(0.1\), batch size \(8\), and train for \(150\) epochs.
Each experiment is repeated with three random seeds, \(\{1,42,1234\}\).
Evaluation uses greedy decoding with temperature \(T=0\) and maximum generation length \(32\).
Learning rates are selected per model and dataset:
\[
\begin{aligned}
&\text{LLaDA-8B: } 5\times 10^{-5}\ \text{for Parent--Child and Person--Description, } 2\times 10^{-4}\ \text{for T-REx},\\
&\text{Dream-7B: }  5\times 10^{-5}\ \text{for Parent--Child and Person--Description, } 2\times 10^{-5}\ \text{for T-REx},\\
&\text{LLaMA-3.1-8B: } 5\times 10^{-5}\ \text{for Parent--Child and Person--Description, } 9\times 10^{-5}\ \text{for T-REx},\\
&\text{Qwen-2.5-7B: } 1\times 10^{-4}\ \text{for Parent--Child and Person--Description, } 5\times 10^{-5}\ \text{for T-REx},\\
&\text{Qwen-3.5-4B: }  5\times 10^{-5}\ \text{for Parent--Child and Person--Description, } 5\times 10^{-5}\ \text{for T-REx}.
\end{aligned}
\]

\paragraph{Forward and reverse accuracy.}
Tab.~\ref{tab:large_scale_reversal} reports large-scale exact-match accuracy averaged over three random seeds.
Across all datasets, ARMs achieve high forward accuracy but almost collapse in the reverse direction.
This pattern holds not only for LLaMA-3.1 and Qwen-2.5, but also for Qwen-3.5, suggesting that reasoning-oriented autoregressive models do not automatically escape the directional limitation induced by left-to-right training.
In contrast, LLaDA retains substantially stronger reverse accuracy across all datasets.
Dream also shows clear reverse-transfer behavior compared with ARMs, although its performance is less uniform than LLaDA.
One plausible reason is that Dream is initialized from an autoregressive model, so part of the left-to-right inductive bias may remain after conversion and fine-tuning.
Overall, the results support the main empirical claim: under the same forward-only training protocol, MDMs show substantially stronger reverse inference than ARMs.

\begin{figure}[H]
\centering
\begin{subfigure}{0.49\linewidth}
\centering
\includegraphics[width=\linewidth]{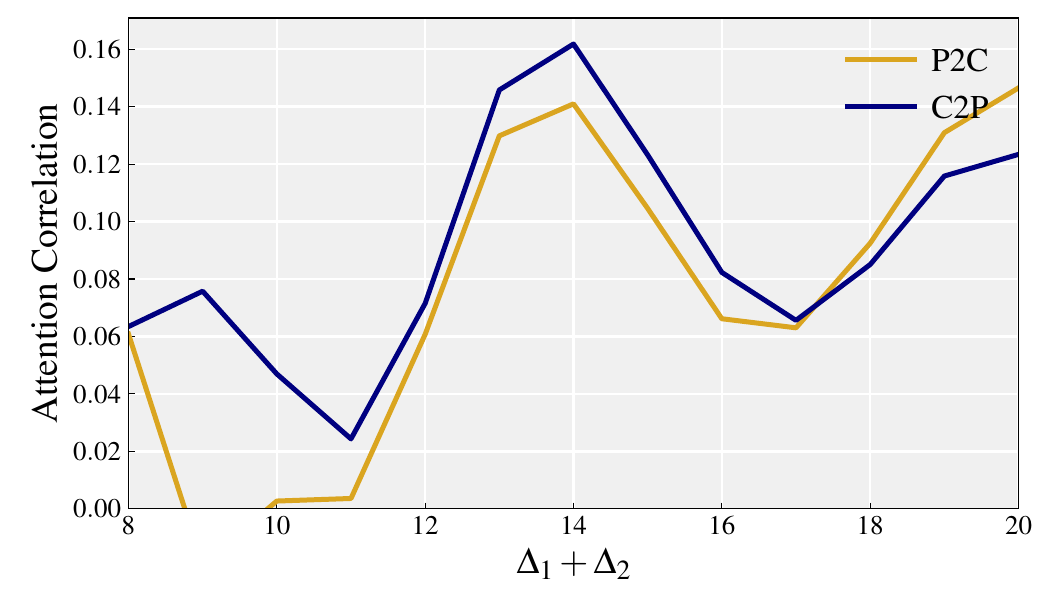}
\caption{Dream attention-score correlation.}
\label{fig:dream_attn}
\end{subfigure}
\hfill
\begin{subfigure}{0.49\linewidth}
\centering
\includegraphics[width=\linewidth]{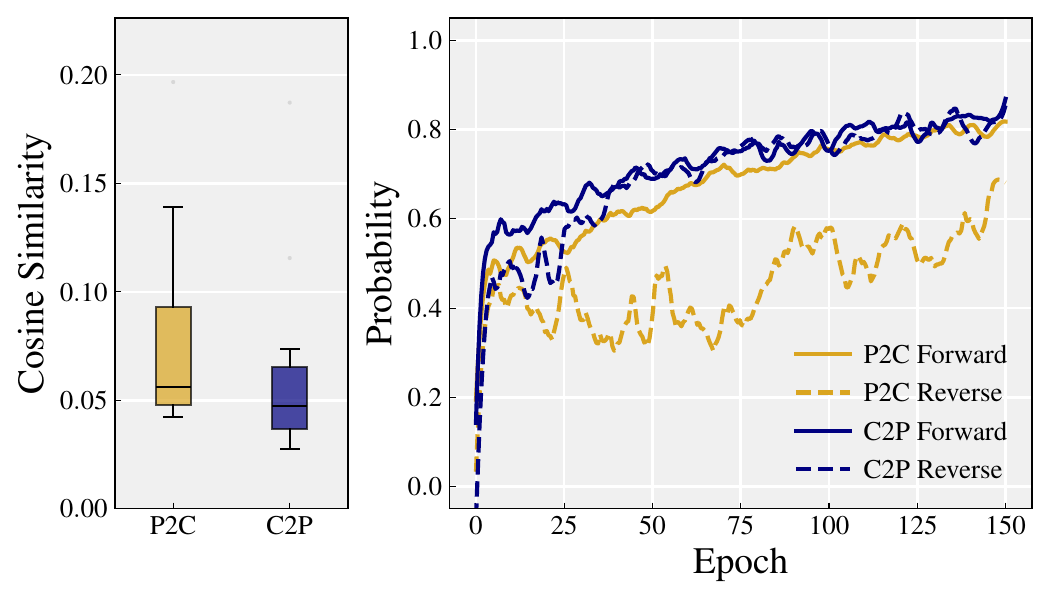}
\caption{Dream gradient and probability transfer.}
\label{fig:dream_large_validation}
\end{subfigure}
\caption{
Large-scale validation of storage-routing signatures in Dream.
\textbf{Left:} Pearson correlation between forward and reverse attention scores for the evidence entity across relative-distance placements.
\textbf{Right:} Forward--reverse gradient cosine similarity over all trainable parameters and correct target-entity probabilities during forward-only fine-tuning.
}
\label{fig:dream_large_mechanism}
\end{figure}

\paragraph{Mechanistic probes in large models.}
For LLaDA and Dream, we additionally measure the three signatures predicted by the theory.
First, we compute the Pearson correlation between forward and reverse attention scores assigned to the shared evidence entity.
Second, we compute the cosine similarity between forward and reverse gradients over trainable parameters.
Third, we track the probability assigned to the correct reverse entity during forward-only fine-tuning.
These probes test whether the storage-routing signatures observed in the one-layer model persist in practical multi-layer MDMs.

In LLaDA, the main text shows that forward and reverse attention scores remain positively correlated, forward and reverse gradients remain positively aligned, and the reverse-entity probability increases during forward-only fine-tuning.
Fig.~\ref{fig:dream_large_mechanism} reports the corresponding measurements for Dream.
Dream shows the same qualitative pattern: forward/reverse attention scores are positively correlated, gradient alignment remains positive, and the reverse probability increases during training.
The signatures are somewhat weaker and less uniform than in LLaDA; residual bias from autoregressive initialization is one possible explanation, not isolated by these experiments.
Nevertheless, the Dream results support the same conclusion as the LLaDA experiments: the storage-routing signatures are not artifacts of the one-layer model and persist in large masked diffusion LLMs.

\section{Discussion and Limitations}
\label{app:discussion}

\subsection{The MDM Framing of Our Analysis}
\label{app:mdm_framing}

We clarify why the paper is framed around masked diffusion models, even though our analysis identifies a more specific set of ingredients behind reversal curse mitigation.
Our claim is not that the word ``diffusion'' itself explains the effect.
Rather, the relevant mechanism is the interaction of three ingredients: \textit{random-masking supervision, full attention, and position-invariant value vectors.}

This framing is natural because modern masked diffusion language models instantiate these ingredients in a scalable generative setting.
Early discrete diffusion models for text were formulated through a forward noising process and a learned reverse denoising process~\citep{austin2021structured,lou2024discrete}.
Recent masked diffusion language models, however, are implemented in a simpler time-free form~\citep{sahoo2024simple,ouyour,nie2025large,ye2025dream7bdiffusionlarge}.
They take partially or fully masked sequences as input, predict distributions over the original tokens at masked positions, and are trained with a weighted masked-token cross-entropy loss.
In this operational sense, they are generative masked language models: they retain the masked-token prediction structure of BERT-style models~\citep{devlin2019bert}, but use it generatively by iteratively filling masked positions.
While BERT-style MLMs may share this mechanism, they are not standalone generative LLMs; modern MDMs are the practical setting where masked reconstruction, full attention, and position-invariant value vectors are jointly realized.

Thus, the MDM framing is not an appeal to diffusion terminology.
It identifies the practical model class in which the relevant ingredients appear together, while our analysis isolates which of those ingredients drive reversal transfer.

\subsection{Clarification on Absolute and Relative Positional Encodings}
\label{app:pe_clarification}
We clarify that the absolute/relative distinction is formally about how position is represented, not about whether positional information enters the value vector.
Absolute encodings use position-specific signals, whereas relative encodings represent pairwise offsets or distances.
For our storage-routing mechanism, however, the relevant issue is whether the value-output path remains position-free.
Widely used absolute encodings~\citep{vaswani2017attention,gehring2017convolutional} are added before the query, key, and value projections, making the value vector position-dependent.
Modern relative encodings~\citep{dai2019transformer,raffel2020exploring,he2021deberta} such as RoPE~\citep{su2024roformer} and ALiBi~\citep{press2022alibi} instead inject position through the query-key attention score or attention bias, without adding positional terms directly to value vectors.
Thus, although ``relative'' does not mean ``position-invariant value vectors'' by definition, the two distinctions largely coincide for the modern positional encodings.

\subsection{The Reversal Curse Formulation}
\label{app:reversal_curse_scope}
The reversal curse refers to a model's failure to answer an order-reversed query about a relation learned in the forward direction.
More generally, suppose training presents a fact \(A\,R\,B\), and the task defines \(B\,R^{-1}\,A\) as its valid inverse.
Success under reversal means that, after learning the forward fact, the model can correctly predict \(A\) from \(B\) under the inverse relation \(R^{-1}\).
Following \citet{berglundreversal}, we adopt ``\(A\) is \(B\)'' and ``\(B\) is \(A\)'' as shorthand for forward and reverse statements.
However, this shorthand can be misleading: the reversal curse is not limited to relations expressed by the verb ``is,'' and ``\(B\) is \(A\)'' is not necessarily a valid inversion of ``\(A\) is \(B\).''
Our analysis instead considers task-defined forward--reverse pairs \(A\,R\,B\) and \(B\,R^{-1}\,A\), where \(R^{-1}\) denotes the appropriate inverse relation.
When a reversed query is not semantically supported by the learned forward fact, simply exchanging \(A\) and \(B\) does not define a valid reversal task and may instead probe inappropriate over-reversal.

\subsection{Parametric Memory versus In-Context Evidence}
\label{app:parametric_memory}
The reversal curse should be distinguished from in-context learning.
If the relevant premise or inverse examples are supplied at test time, a model may answer the reverse query by using information newly provided in the prompt.
This is useful, but it is a different problem from the one studied here.
We ask whether a relation learned during training, without being restated in the prompt, becomes accessible through the model's parameters under an order-reversed query.
This parametric-memory setting is important because a model cannot always rely on having the relevant fact or inverse examples in context.
The same distinction applies to chain-of-thought prompting and reasoning-oriented inference: such methods can help manipulate evidence that is already available, 
but they do not by themselves guarantee that forward-learned evidence is retrievable from parameters under a reverse query.

\subsection{Scope of the Claims}
\label{app:theory_scope}

\paragraph{What the theory guarantees.}
We clarify the level of guarantee provided by our theory.
Secs.~\ref{sec:shared_y_update} and \ref{sec:attention_correlation} identify the storage-routing mechanism by which forward and reverse configurations share evidence and maintain a correlated route to that evidence.
These results explain why reverse transfer can occur, but they do not predict the final reverse accuracy.
Sec.~\ref{sec:gradient_alignment} gives a local first-order alignment result: along the shared storage column, a forward gradient step decreases the corresponding reverse loss to first order.
This identifies the sign of the transfer effect in the isolated component, but does not characterize the full training trajectory, the exact magnitude of the effect, or the final probability assigned to the reverse answer.

The analysis separates three levels.
Sec.~\ref{sec:attention_correlation} characterizes the raw score assigned to the designated evidence token.
Softmax competition converts this score into an attention weight, and the attended value-side evidence from all positions then determines the target-versus-competitor logits and final retrieval.

Proposition~\ref{prop:reverse_retrieval_margin} makes this hierarchy exact.
The reverse margin contains a designated-evidence term \(\alpha_{i,j}^{\mathrm{rev}}(Y_{A,B}-Y_{C,B})\) and an aggregate term from the remaining context.
Softmax competition determines \(\alpha_{i,j}^{\mathrm{rev}}\), while attended context can contribute stored evidence favoring either \(A\) or \(C\).
Tab.~\ref{tab:reverse_attention_weights} provides the corresponding softmax-level diagnostic for the synthetic one-layer checkpoints.
App.~\ref{app:probability_transfer} establishes convergence in the single-informative-token setting, whereas Proposition~\ref{prop:reverse_retrieval_margin} identifies the additional terms that govern retrieval with competing context.

\paragraph{Positional evidence reuse and semantic inversion.}
Our theory isolates a mechanism supporting valid reversal: entity-level evidence learned from the forward statement can remain stored and routable when the target and evidence positions are reversed.
Concretely, forward masked training can create \(B\)-to-\(A\) evidence that remains usable when a valid reverse query asks for \(A\) from \(B\).
This does not establish that the model derives the meaning of \(R^{-1}\) from \(R\), or decides whether the reverse relation is logically valid.
The synthetic experiments remove semantic cues to isolate this evidence-reuse mechanism, while the Parent--Child experiments test its predicted signatures on a task with valid inverse relations.

The same caution applies to positional encodings.
Our ALiBi results show that stronger preservation of the content score can yield stronger reverse transfer in our reversal task.
They do not imply that ALiBi is universally preferable to RoPE or other positional encodings.
A positional encoding that produces high reverse transfer may also induce different tradeoffs on tasks requiring fine positional discrimination, long-context extrapolation, or order-sensitive reasoning.
Thus, the positional encoding intervention should be read as evidence that routing geometry affects parametric reversibility, not as a general ranking of positional encodings.

\subsection{The One-Layer Theory versus Deep Models}
\label{app:one_layer_vs_deep}

The one-layer analysis follows a standard approach in theoretical studies of Transformer training dynamics~\citep{tian2023scan,zhu2024towards,huang2025generalization,li2023transformers,li2024one,huangtransformers}.
Its value here is that position-independent storage and position-dependent routing become analytically separable, yielding falsifiable predictions for each.
The main-text guarantees apply to this one-layer model and its local updates.

Deep MDMs additionally contain residual paths, MLP blocks, layer composition, head specialization, and repeated denoising steps, which may modulate the isolated signal.
The LLaDA and Dream experiments test the aggregate score, gradient, and probability signatures predicted by the one-layer analysis and find qualitatively consistent trends, but do not verify the one-layer guarantees under full deep-model dynamics.
A theory of how these signals compose across layers and denoising steps remains open.

\end{document}